%% file: main.tex
\documentclass{article} %

\usepackage{caption}

\usepackage[accepted]{icml2025}

\input{math_commands.tex}

\usepackage{microtype}
\usepackage{framed}
\usepackage{url}
\usepackage[utf8]{inputenc} %
\usepackage[T1]{fontenc}    %
\usepackage{url}            %
\usepackage{booktabs}       %
\usepackage{amsfonts}       %
\usepackage{nicefrac}       %
\usepackage{microtype}      %
\usepackage{xcolor}         %
\usepackage{amsmath,amssymb,amsfonts,amsthm,mathrsfs}
\usepackage[page,title,titletoc,header]{appendix}
\usepackage{enumerate}
\usepackage{color}
\usepackage{colortbl}
\usepackage{hyperref}
\usepackage{natbib}
\usepackage{graphicx}
\usepackage{indentfirst}
\usepackage[mathscr]{eucal}
\usepackage{multirow}
\usepackage[ruled,linesnumbered]{algorithm2e}
\usepackage{algpseudocode}

\usepackage{nicefrac}
\usepackage{subfigure}
\usepackage{enumitem}
\usepackage{wrapfig}

\usepackage{CJKutf8}
\usepackage{tabularray}
\usepackage{bigstrut}
\usepackage{bbm}
\usepackage{wrapfig}
\usepackage{float}
\usepackage{fontawesome}
\usepackage[capitalize]{cleveref}
\usepackage{listings}
\usepackage{tikz}
\usepackage[normalem]{ulem}

\newcommand{\emoji}[2][1.2em]{\raisebox{-0.2\height}{\includegraphics[height=#1]{#2}}}
\definecolor{fbApp}{HTML}{ffe4e3}
\definecolor{mydarkblue}{rgb}{0,0.3,0.9}

\newcommand{\rowc}{\rowcolor{fbApp}}

\newcommand*\circled[1]{\tikz[baseline=(char.base)]{\node[shape=circle,fill,inner sep=0.5pt] (char) {\textcolor{white}{\small \bf #1}};}}
\hypersetup{colorlinks=true,citecolor=mydarkblue}
 
\icmltitlerunning{\ours: Visual Masked Autoencoders Are Free-Lunch Zero-Shot Time Series Forecasters}

\newcommand{\method}[1]{\textsc{#1}\xspace}
\newcommand{\mae}{\texttt{MAE}\xspace}
\newcommand{\moirai}[1]{\textsc{Moirai}\textsubscript{#1}\xspace}
\newcommand{\ours}{\method{VisionTS}}
\newcommand{\eg}{{\it e.g.}}

\newcommand{\ie}{{\it i.e.}}

\newcommand{\hyperparameterwidth}{0.17\textwidth}

\begin{document}

\twocolumn[
\icmltitle{\ours: Visual Masked Autoencoders Are Free-Lunch\\ Zero-Shot Time Series Forecasters}

\begin{icmlauthorlist}
\icmlauthor{Mouxiang Chen}{zju}
\icmlauthor{Lefei Shen}{zju}
\icmlauthor{Zhuo Li}{state_street}
\icmlauthor{Xiaoyun Joy Wang}{state_street}
\icmlauthor{Jianling Sun}{zju}
\icmlauthor{Chenghao Liu}{salesforce}
\end{icmlauthorlist}

\icmlaffiliation{zju}{Zhejiang University}
\icmlaffiliation{state_street}{State Street Technology (Zhejiang) Ltd}
\icmlaffiliation{salesforce}{Salesforce Research Asia}

\icmlcorrespondingauthor{Chenghao Liu}{chenghao.liu@salesforce.com}
\icmlcorrespondingauthor{Zhuo Li}{lizhuo@zju.edu.cn}

\icmlkeywords{Time Series Forecasting, Foundation Models, Transfer Learning}

\vskip 0.3in

{%

\begin{center}
    \centering
    \captionsetup{type=figure}
    \includegraphics[width=\textwidth]{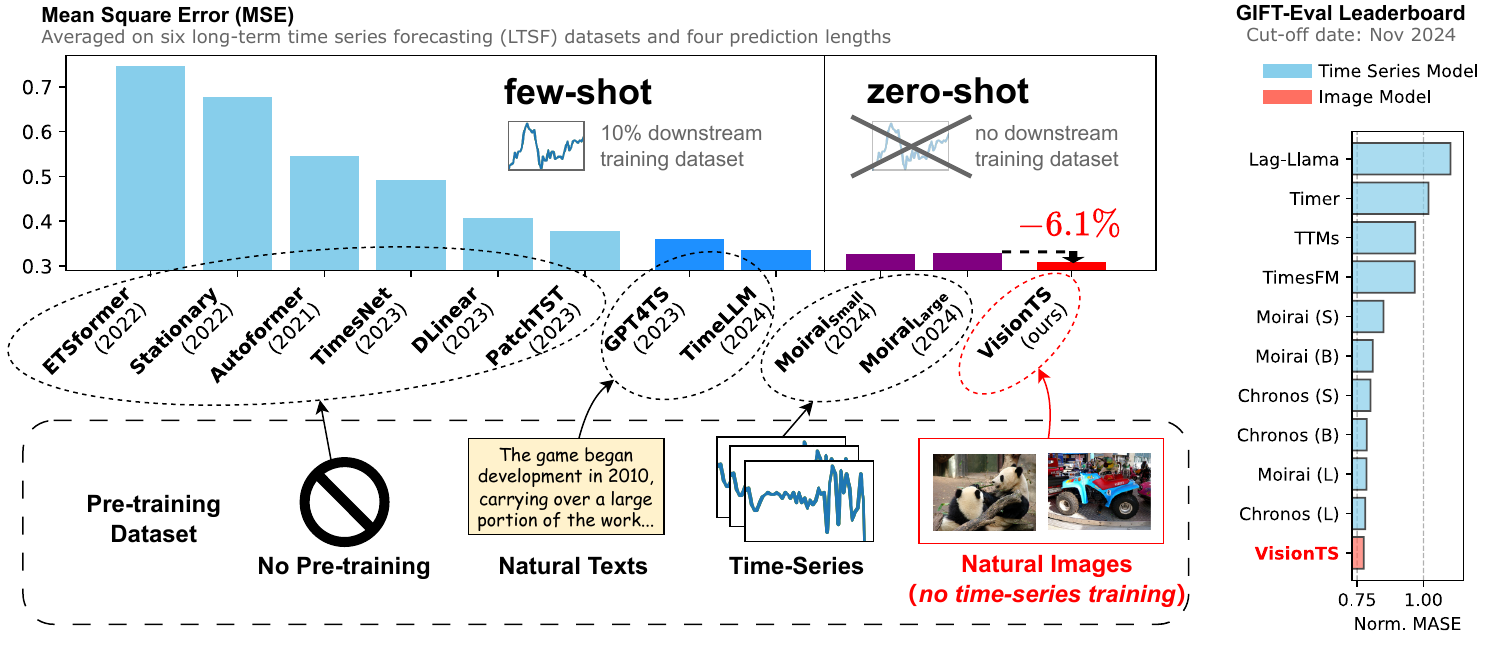}
    \captionof{figure}{Long-term forecasting (left) and GIFT-Eval (right) performance comparison. Our \ours, \textit{without any training on time series data}, outperforms the pure time series foundation models in the zero-shot setting.}
    \label{fig:intro}
\end{center}
\vspace{0.4cm}
}

]
\printAffiliationsAndNotice{}

\input{section/abstract}

\input{section/introduction}
\input{section/method}

\input{section/experiment}

\input{section/related-work}
\input{section/conclusion}

\section*{Impact Statement}

This paper presents work whose goal is to advance the field of time series forecasting. There are many potential societal consequences of our work, none of which we feel must be specifically highlighted here.

\bibliography{ref}
\bibliographystyle{icml2025}

\appendix

\input{section/appendix}

\end{document}

%% file: math_commands.tex
\usepackage{amsmath,amsfonts,bm}

\def\eqref#1{equation~\ref{#1}}

\def\1{\bm{1}}

\def\vx{{\bm{x}}}

\def\mI{{\bm{I}}}

\def\mX{{\bm{X}}}

\DeclareMathAlphabet{\mathsfit}{\encodingdefault}{\sfdefault}{m}{sl}
\SetMathAlphabet{\mathsfit}{bold}{\encodingdefault}{\sfdefault}{bx}{n}

%% file: section/abstract.tex
\begin{abstract}

Foundation models have emerged as a promising approach in time series forecasting (TSF). Existing approaches either repurpose large language models (LLMs) or build large-scale time series datasets to develop TSF foundation models for universal forecasting. However, these methods face challenges due to the severe cross-domain gap or in-domain heterogeneity. This paper explores a new road to building a TSF foundation model from rich, high-quality natural images. Our key insight is that a visual masked autoencoder, pre-trained on the ImageNet dataset, can naturally be a numeric series forecaster. By reformulating TSF as an image reconstruction task, we bridge the gap between image pre-training and TSF downstream tasks. Surprisingly, without further adaptation in the time series domain, the proposed \ours could achieve better zero-shot forecast performance than existing TSF foundation models. With fine-tuning for one epoch, \ours could further improve the forecasting and achieve state-of-the-art performance in most cases. Extensive experiments reveal \textit{intrinsic similarities} between images and real-world time series, suggesting that visual models may offer a ``free lunch'' for TSF and highlight the potential for future cross-modality research. Our code is publicly available at \url{https://github.com/Keytoyze/VisionTS}.

\end{abstract}

%% file: section/introduction.tex
\section{Introduction}

Foundation models \citep{FoundationModel} have revolutionized natural language processing (NLP) and computer vision (CV) in recent years \citep{GPT,mae}. By pretraining on large-scale data, they have shown remarkable few-shot and even zero-shot performance across various downstream tasks. This has motivated an emergent paradigm shift in time series forecasting (TSF), moving from a traditional one-model-per-dataset framework to \textit{universal forecasting} with a single pre-trained model \citep{Moirai,Moment}. A TSF foundation model can greatly reduce the need for downstream data and demonstrate strong forecasting performance on diverse domains, such as energy consumption planning, weather forecasting, and traffic flow.

We have recently witnessed two roads to building a TSF foundation model for universal forecasting. The \textit{first} tries to repurpose large language models (LLMs) that have been pre-trained on text data for TSF tasks (\ie, \textbf{text-based}) \citep{GPT4TS,timellm}, based on the observation that LLMs and TSF models share a similar left-to-right forecasting paradigm. However, due to the significant gap between these two modalities, the effectiveness of such transferability between language and time series has recently been questioned by \citet{AreLLMUseful}.

The \textit{second} road focuses on constructing large-scale time-series datasets collected from diverse domains to train a TSF foundation model from scratch (\ie, time series-based or \textbf{TS-based}) \citep{Moirai,TimesFM}. Nevertheless, unlike images or language with unified formats, time series data is highly heterogeneous in length, frequency, number of variates, domains, and semantics, limiting the transferability between pre-training and downstream domains. Until recently, constructing a high-quality dataset remains challenging and is still in the early exploration stage.

\begin{figure}[t]
  \centering
  \includegraphics[width=0.4\textwidth]{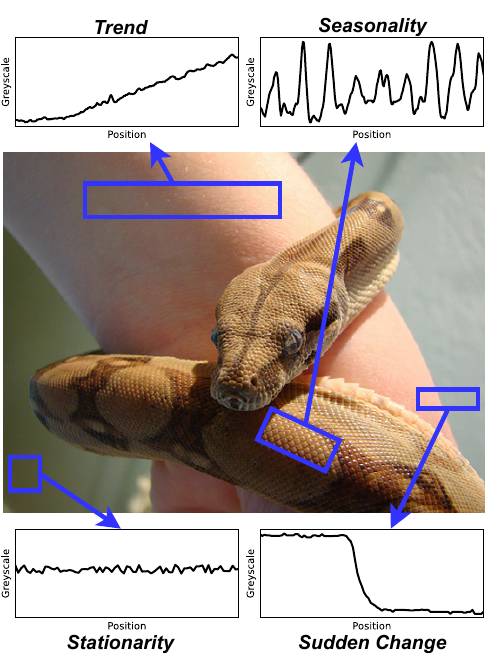}
  \label{fig:imagenet-sample}
  \caption{An image of the ImageNet dataset \citep{ImageNet}, in which the pixel arrays can display many well-known features of real-world time series, such as trend, seasonality, and stationarity \citep{qiu2024tfb}. By self-supervised pre-training on ImageNet, it is reasonable that a visual model could understand these features and exhibit a level of time series forecasting ability.}
  \vspace{-0.3cm}
\end{figure}

In this paper, we investigate a \textit{third} road that is less explored yet promising: building TSF foundation models with pre-trained \textit{visual} models. Our key idea is that pixel variations in a natural image can be interpreted as temporal sequences, which share many intrinsic similarities with time series: \circled{1} \textbf{Similar modalities}: Unlike discrete texts, both images and time series are continuous; \circled{2} \textbf{Similar origin}: Both time series and images are observations of real-world physical systems, whereas languages are products of human cognitive processes; \circled{3} \textbf{Similar information density}: Languages are human-generated signals with high semantic density, while images and time series are natural signals with heavy redundancy \citep{mae}; and \circled{4} \textbf{Similar features}: As shown in \cref{fig:imagenet-sample}, images often display many features of real-world time series, which are rarely found in language data. Based on these findings, images could be a promising modality for transferring to TSF. We are motivated to answer the question: \textit{Can a visual model pre-trained on images be a free-lunch foundation model for time series forecasting?}

We focus on visual masked autoencoder (\mae)\footnote{We use fonts to distinguish \mae (Masked Autoencoder) and MAE (Mean Absolute Error) in this paper.}, a popular CV foundation model \citep{mae} by self-supervised pre-training on ImageNet \citep{ImageNet}. As an image reconstruction and completion model, \mae can naturally be a \textit{numeric series forecaster.} Inspired by the well-known prompt technique in NLP \citep{PromptTuning}, we propose a simple method to reformulate TSF as a patch-level image reconstruction task to bridge the gap between pre-training and downstream tasks. Specifically, we transform 1D time-series data into 2D matrices via segmentation. Then, we render the matrices into images and align the forecasting window with masked image patches. This allows us to make a zero-shot forecast without further adaptation.

We evaluate our proposed \ours on large-scale benchmarks, including 8 long-term TSF \citep{Informer}, 29 Monash \citep{Monash}, and 23 GIFT-Eval \citep{GIFT-Eval} datasets, spanning diverse domains, frequencies, and multivariates. To the best of our knowledge, \textbf{the scale of our evaluation benchmark is the largest among existing TSF foundation models}. As demonstrated in \cref{fig:intro}, \textit{without} further adaptation on time series, a vanilla \mae can surprisingly achieve a comparable performance or even outperform the strong zero-shot TSF foundation models. By fine-tuning \mae in each downstream dataset for a single epoch, \ours can lead to SOTA performance in most long-term TSF benchmarks. 

To further understand and explain the transferability, we use an \mae encoder to visualize both modalities, showing a level of similarity between time series and natural image representations. Additionally, \textbf{we observe considerable heterogeneity within time-series data across domains, and images can serve as a \textit{bridge} to connect these isolated time-series representations}. This could further explain why \ours performs better than some cross-domain TSF models. Our findings suggest that time series and natural images may be two sides of a coin, and visual models can be a \textit{free lunch} for time series forecasting. We hope our findings inspire future cross-modal research on CV and TSF.

Our contributions are summarized as follows:

\begin{itemize}[leftmargin=*]

\item We explore a road to building a TSF foundation model from natural images, conceptually different from the existing text-based and TS-based pre-training methods.

\item We introduce \ours, a novel TSF foundation model based on a visual \mae. To bridge the gap between the two modalities, we reformulate the TSF task into an image reconstruction task.

\item Comprehensive evaluations of \ours on large-scale benchmarks across multiple domains demonstrate its significant forecasting performance, surpassing few-shot text-based TSF foundation models and achieving comparable or superior results to zero-shot TS-based models.

\end{itemize}

%% file: section/method.tex
\section{Preliminaries}

\paragraph{Time Series Forecasting (TSF)}

For a multivariate time series with $M$ variables, let $\vx_{t} \in \mathbb{R}^M$ represent the value at $t$-th time step. Given a historical sequence (\ie, look-back window) $\mX_{t-L:t} = [\vx_{t-L}, \cdots, \vx_{t-1}] \in \mathbb{R}^{L\times M}$ with context length $L$, the TSF task is to predict future values (\ie, forecast horizon) with prediction length $H$: $\hat{\mX}_{t:t+H} = [\vx_{t}, \cdots, \vx_{t+H-1}] \in \mathbb{R}^{H\times M}$.

\paragraph{Patch-Level Image Reconstruction} 
To obtain high-quality visual representation for downstream CV tasks, \citet{mae} proposed masked autoencoder (\mae) to pre-train a Vision Transformer (ViT) \citep{ViT} using a patch-level image reconstruction task on ImageNet. Specifically, for an image of size $W\times W$ (where $W$ represents both the width and height, as ImageNet images are square), the image is evenly divided into $N\times N$ patches, each with a width and height of $S=\nicefrac{W}{N}$. During pre-training, some random patches are masked, while the remaining \textit{visible patches} are fed into the ViT with their position encodings. \mae are trained to reconstruct the masked pixel values from these visible patches.

\section{Methodology}\label{sec:method}

As noted in the Introduction, time series and images share intrinsic \textit{similarities}, suggesting the transfer potential of pre-trained visual models (particularly \mae in this paper) for TSF. To reformulate TSF tasks into \mae's pre-training task, our high-level idea is straightforward: map the look-back/forecasting windows to visible/masked patches, respectively. This idea is supported by the prompt tuning \citep{PromptTuning} in NLP, where the predictions for \texttt{[mask]} token in pre-trained language models, \eg, BERT \citep{BERT}, are directly used for downstream tasks. By unifying the forms of the two tasks, we bridge the gap between the two modalities without further training.

\begin{figure*}[t]
  \centering
  \includegraphics[width=0.75\textwidth]{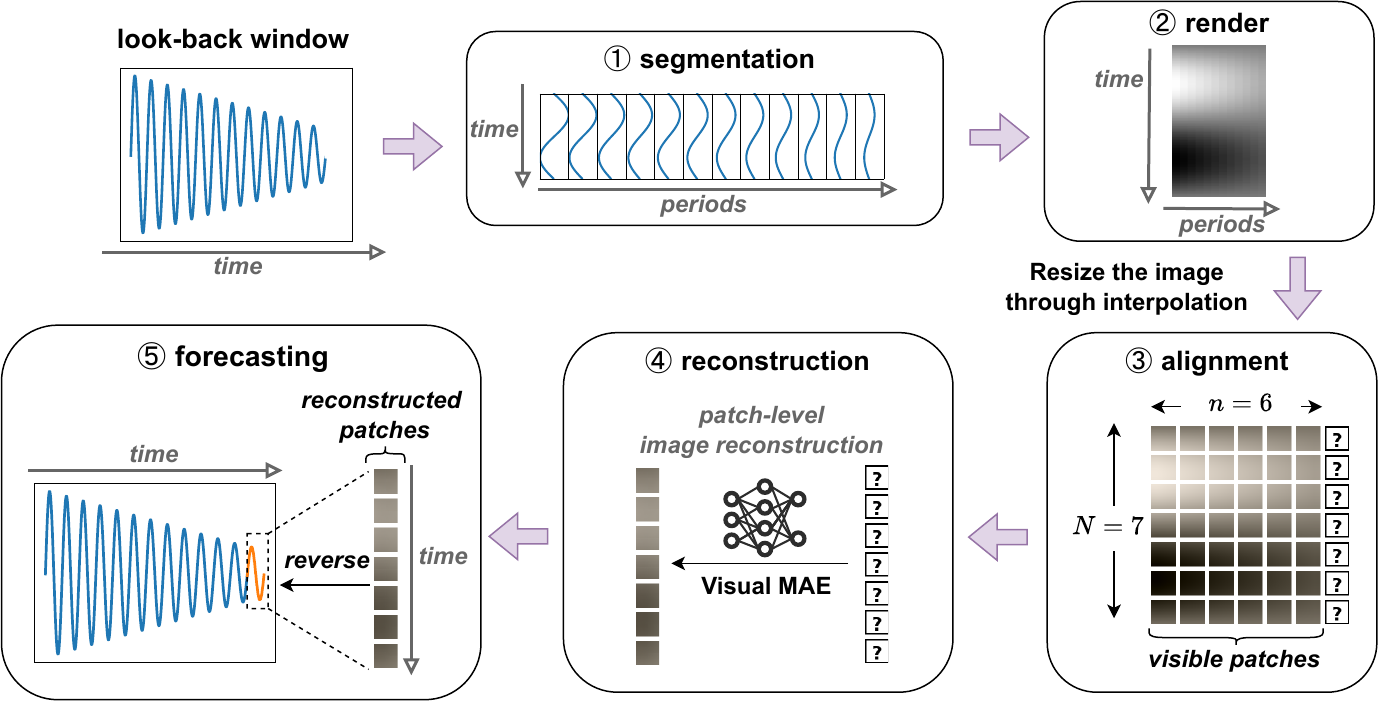}
  \caption{\ours architecture. The input is first segmented by period, rendered into a grayscale image, and then aligned with the visible patches on the left through resampling. \mae is used to predict the masked patches on the right, and the reconstructed image is then reversed to forecasting.}
  \label{fig:method}
\end{figure*}

However, implementing this idea poses a challenge: the dimension of time-series data (1D) is different from images (2D). Moreover, the size of images in the pre-training dataset is fixed at 224 $\times$ 224, while the lengths of time series data can vary dynamically. In the following, we describe the details of \ours to address this challenge. Our architecture is depicted in \cref{fig:method}.

\paragraph{Segmentation} Given a univariate input $X\in \mathbb R^{L}$, the first goal is to transform it into a 2D matrix. We propose to segment it into $\lfloor \nicefrac{L}{P} \rfloor$ subsequences of length $P$, where $P$ is the periodicity. Notably, when the time series lacks clear periodicity, we can set $P=1$ directly, which is also effective in our experiments (\cref{sec:app_zs_monash}). In practice, $P$ can be determined using statistical methods like Fast Fourier Transform \citep{TimesNet,CalibrationCDS} or domain knowledge like sampling frequency \citep{Monash,GluonTS}. In this paper, we select $P$ based on the sampling frequency, elaborated in \cref{sec:app_p}.

After that, these subsequences are then stacked into a 2D matrix, denoted by $\mI_{\text{raw}}\in \mathbb R^{P\times \lfloor \nicefrac{L}{P} \rfloor}$. This encoding strategy is proven to be efficient by recent work like TimesNet \citep{TimesNet} and SparseTSF \citep{SparseTSF}, as it allows for the simultaneous capture of both variations within the same period (\ie, intra-period) and across periods with the same phase (\ie, inter-period). Moreover, it ensures that each element in $\mI_{\text{raw}}$ and its neighbors align with the \textit{spatial locality} property of images \citep{CNN}, where nearby pixels tend to be similar due to the inherent cohesiveness of objects in the real world. Therefore, this further narrows the gap between time series and images. 

\paragraph{Normalization} \mae standardizes each image based on the mean and standard deviation computed on ImageNet. Therefore, we apply instance normalization to $\mI_{\text{raw}}$, which is also a standard practice in current TSF \citep{RevIn}. Notably, we observed that normalizing $\mI_{\text{raw}}$ to a standard deviation of $r$, where $r$ is a hyperparameter less than 1, yields superior performance. One explanation is that the magnitude of inputs/outputs during \mae pretraining is constrained by the limited range of color values. Therefore, reducing the magnitude of $\mI_{\text{raw}}$ prevents exceeding these limits. However, an excessively low $r$ can result in values that are difficult to distinguish. We found that a moderate value (0.4) of $r$ performs well across most scenarios (See \cref{sec:app_zs_param} for more details). Let $\mI_{\text{norm}}$ denote the normalized matrix, which is computed as follows:
\begin{align*}
    \mI_{\text{norm}}=r\cdot \frac{\mI_{\text{raw}} - \text{Mean}(\mI_{\text{raw}})}{\text{Standard-Deviation}(\mI_{\text{raw}})}.
\end{align*}

\paragraph{Rendering} Since each image has three channels, we simply render $\mI_{\text{norm}}$ as a grayscale image $\mI_{\text{grey}}\in \mathbb R^{P\times \lfloor \nicefrac{L}{P} \rfloor \times 3}$, where all three channels are identical to $\mI_{\text{norm}}$. This choice is purely result-driven: In our early experiments, we added a convolutional layer with three output channels to convert the grayscale image into a color image and then fine-tuned it to find the optimal color transformation, which, however, did not significantly improve the performance.

\paragraph{Alignment} Our goal is to predict the columns on the right of $\mI_{\text{grey}}$ to forecast the future sequence. A straightforward approach is to treat $\mI_{\text{grey}}$ as the visible left portion and the predicted columns as the masked right portion. However, since the image size during pre-training may not match the size of $\mI_{\text{grey}}$, we propose to resize $\mI_{\text{grey}}$ to align with the pre-training data. Formally, let the total number of 2D patches used in pre-training be $N\times N$ and the size of each patch be $S\times S$. We set the number of visible patches to $N\times n$ and the masked patches to $N\times (N-n)$, where $n = \lfloor N\cdot \nicefrac{L}{(L+H)} \rfloor$ is determined by the ratio of context length $L$ to prediction length $H$. We resample the image $\mI_{\text{grey}}$ to adjust the size from the original dimensions $(P, \lfloor \nicefrac{L}{P} \rfloor)$ to $(N\cdot S, n\cdot S)$, making it more compatible with \mae. We select \textit{bilinear interpolation} for the resampling process.

Moreover, we found that reducing the width of the visible portion can further improve performance.
One possible explanation is that \mae uses a large masked ratio during pre-training, with only 25\% of patches visible. Reducing the image width may align the masked ratio more closely with pre-training. Therefore, we propose multiplying $n$ by a hyperparameter $c\in [0,1]$. Similar to $r$, we found that setting $c=0.4$ performs well in our experiments (See \cref{sec:app_zs_param}). This can be formulated as $n=\left\lfloor c\cdot N\cdot\nicefrac{L}{(L+H)} \right\rfloor$.

\paragraph{Reconstruction and Forecasting} After obtaining the \mae-reconstructed image, we simply reverse the previous steps for forecasting. Specifically, we resize the entire image back to the original time series segmentations through the same bilinear interpolation, and average the three channels to obtain a single-channel image. After de-normalizing and flattening, the forecasting window can be extracted.

\input{table/tab_zero_shot_avg}

 \paragraph{Discussion on Multivariate Forecasting} In addition to the temporal interactions, multivariate time series data sometimes show interactions between variables. While pre-trained vision models effectively capture temporal interactions based on the intrinsic similarities between images and time series, they struggle to capture inter-variable interactions due to a limited number of image channels, especially without further training. Fortunately, recent work shows that channel independence --- forecasting each variable separately --- can be effective and is widely used in recent deep forecasting models \citep{PatchTST,ChannelIndependence,timellm,GPT4TS,SparseTSF}. Following these works, we adopt channel independence in our paper while leaving the exploration of capturing inter-variable interactions to future work.

%% file: table/tab_zero_shot_avg.tex
\begin{table*}[t]
  \centering
  \caption{Zero-shot or few-shot results on the long-term TSF benchmark. Results are averaged across prediction lengths \{96, 192, 336, 720\}, with full results in \cref{tab:zero_shot_full} (\cref{sec:app_zs_full}). \textbf{Bold}: the best result.}
\vskip 0.15in

    \resizebox{\linewidth}{!}{

\begin{tabular}{ccccccccccccccccc}
\toprule
\multicolumn{2}{c}{} &   \quad\quad   & \multicolumn{5}{c}{\textbf{\emoji{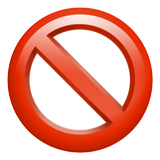} Zero-Shot}} &  \quad\quad    & \multicolumn{8}{c}{\textbf{\emoji{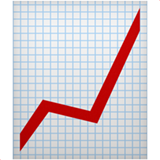} Few-Shot (10\% In-distribution Downstream Dataset)}} \\
\cmidrule{4-8}\cmidrule{10-17}\multicolumn{3}{c}{\textbf{Pretrain}} & \textit{\textbf{\emoji{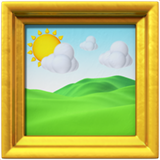} Images}} &      & \multicolumn{3}{c}{\textit{\textbf{\emoji{figure/em-chart_with_upwards_trend} Time series}}} &      & \multicolumn{2}{c}{\textit{\textbf{\emoji{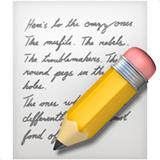} Text}}} &      & \multicolumn{5}{c}{\textit{\textbf{\emoji{figure/em-no_entry_sign} No Pretrain}}} \\
\cmidrule{4-4}\cmidrule{6-8}\cmidrule{10-11}\cmidrule{13-17}\multicolumn{3}{c}{\textbf{Method}} & \textbf{\ours} &      & \textbf{\moirai{S}} & \textbf{\moirai{B}} & \textbf{\moirai{L}} &      & \textbf{TimeLLM} & \textbf{GPT4TS} &      & \textbf{DLinear} & \textbf{PatchTST} & \textbf{TimesNet} & \textbf{Autoformer} & \textbf{Informer} \\
\cmidrule{1-4}\cmidrule{6-8}\cmidrule{10-11}\cmidrule{13-17}\multicolumn{2}{c}{\multirow{2}[2]{*}{ETTh1}} & MSE & \textbf{0.390} &      & 0.400  & 0.434  & 0.510  &      & 0.556  & 0.590  &      & 0.691  & 0.633  & 0.869  & 0.702  & 1.199  \\
\multicolumn{2}{c}{} & MAE & \textbf{0.414} &      & 0.424  & 0.439  & 0.469  &      & 0.522  & 0.525  &      & 0.600  & 0.542  & 0.628  & 0.596  & 0.809  \\
\midrule
\multicolumn{2}{c}{\multirow{2}[1]{*}{ETTh2}} & MSE & \textbf{0.333} &      & 0.341  & 0.346  & 0.354  &      & 0.370  & 0.397  &      & 0.605  & 0.415  & 0.479  & 0.488  & 3.872  \\
\multicolumn{2}{c}{} & MAE & \textbf{0.375} &      & 0.379  & 0.382  & 0.377  &      & 0.394  & 0.421  &      & 0.538  & 0.431  & 0.465  & 0.499  & 1.513  \\
\midrule
\multicolumn{2}{c}{\multirow{2}[1]{*}{ETTm1}} & MSE & \textbf{0.374} &      & 0.448  & 0.382  & 0.390  &      & 0.404  & 0.464  &      & 0.411  & 0.501  & 0.677  & 0.802  & 1.192  \\
\multicolumn{2}{c}{} & MAE & \textbf{0.372} &      & 0.410  & 0.388  & 0.389  &      & 0.427  & 0.441  &      & 0.429  & 0.466  & 0.537  & 0.628  & 0.821  \\
\midrule
\multicolumn{2}{c}{\multirow{2}[1]{*}{ETTm2}} & MSE & 0.282  &      & 0.300  & \textbf{0.272} & 0.276  &      & 0.277  & 0.293  &      & 0.316  & 0.296  & 0.320  & 1.342  & 3.370  \\
\multicolumn{2}{c}{} & MAE & 0.321  &      & 0.341  & 0.321  & \textbf{0.320} &      & 0.323  & 0.335  &      & 0.368  & 0.343  & 0.353  & 0.930  & 1.440  \\
\midrule
\multicolumn{2}{c}{\multirow{2}[1]{*}{Electricity}} & MSE & 0.207  &      & 0.233  & 0.188  & 0.188  &      & \textbf{0.175} & 0.176  &      & 0.180  & 0.180  & 0.323  & 0.431  & 1.195  \\
\multicolumn{2}{c}{} & MAE & 0.294  &      & 0.320  & 0.274  & 0.273  &      & 0.270  & \textbf{0.269} &      & 0.280  & 0.273  & 0.392  & 0.478  & 0.891  \\
\midrule
\multicolumn{2}{c}{\multirow{2}[2]{*}{Weather}} & MSE & 0.269  &      & 0.242  & 0.238  & 0.260  &      & \textbf{0.234} & 0.238  &      & 0.241  & 0.242  & 0.279  & 0.300  & 0.597  \\
\multicolumn{2}{c}{} & MAE & 0.292  &      & 0.267  & \textbf{0.261} & 0.275  &      & 0.273  & 0.275  &      & 0.283  & 0.279  & 0.301  & 0.342  & 0.495  \\
\midrule
\multicolumn{2}{c}{\multirow{2}[2]{*}{\textbf{Average}}}
& MSE & \textbf{0.309} &      & 0.327  & 0.310  & 0.329  &      & 0.336  & 0.360  &      & 0.407  & 0.378  & 0.491  & 0.678  & 1.904  \\
\multicolumn{2}{c}{} & MAE & 0.345  &      & 0.357  & \textbf{0.344} & 0.350  &      & 0.368  & 0.378  &      & 0.416  & 0.389  & 0.446  & 0.579  & 0.995  \\
\rowc
\multicolumn{3}{c}{\textbf{$\bf 1^{st}$ count}} & \textbf{7} &      & 0    & 3    & 1    &      & 2    & 1    &      & 0    & 0    & 0    & 0    & 0 \\
\bottomrule
\end{tabular}%

 }
  \label{tab:zero_shot_avg}%
\vskip -0.1in
\end{table*}%

%% file: section/experiment.tex
\section{Experiments}

\begin{figure}[t]
  \centering
  \includegraphics[width=0.45\textwidth]{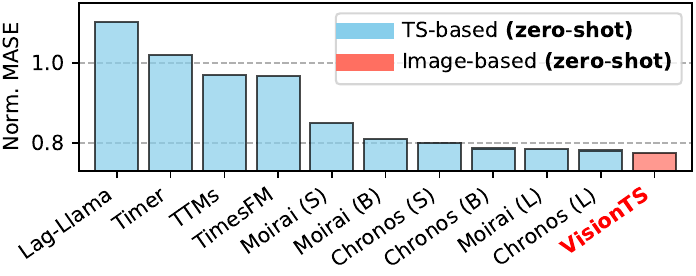}
  \caption{Performance on the GIFT-Eval Leaderboard (cut-off at \ours's release).}
  \label{fig:zs_gift_aggregated}
\end{figure}

\begin{figure}[t]
  \centering
  \includegraphics[width=0.45\textwidth]{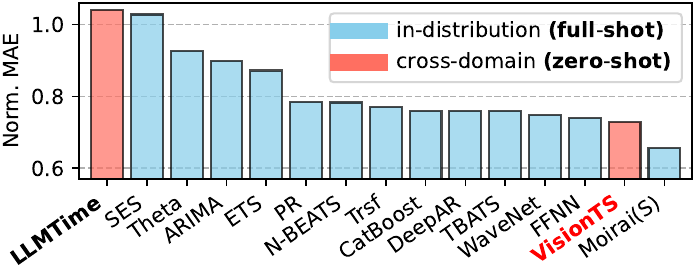}
  \caption{Aggregated results on the Monash TSF Benchmark, with full results in \cref{tab:zero_shot_monash} (\cref{sec:app_zs_monash}).}
  \label{fig:zs_monash_aggregated}
\end{figure}

We follow the standard evaluation protocol proposed by \citet{Moirai} to test our \ours on 35 widely-used TSF benchmarks, and additionally evaluate it on the GIFT-Eval \citep{GIFT-Eval} which is the \textit{largest} TSF benchmark for zero-shot foundation models. We use \mae (Base) as our backbone by default. Baseline and benchmark details are elaborated in \cref{sec:app_exp_detail}.

\subsection{Zero-Shot Time Series Forecasting}\label{sec:exp_zero}

\paragraph{Setups} We first evaluate \ours's \textbf{zero-shot} TSF performance without fine-tuning on time-series modalities. To prevent data leakage, we selected six widely-used datasets from the long-term TSF benchmark that are not included in \moirai{}'s pre-training set for evaluation. Since most baselines cannot perform zero-shot forecasting, we report their \textbf{few-shot} results by fine-tuning on the 10\% of the individual target datasets. We also evaluate the Monash benchmark and GIFT-Eval benchmark. Notably, the Monash benchmark is more challenging for \ours since they were used in \moirai{}'s pre-training but not for \ours. We set the hyperparameters to $r=c=0.4$. Following common practice \citep{GPT4TS,Moirai}, we conduct hyperparameter tuning on validation sets to determine the optimal context length $L$, detailed in \cref{sec:app_zs_tune}.

\input{table/tab_zero_shot_exps}

\paragraph{Results on Long-Term TSF Benchmark} \cref{tab:zero_shot_avg} shows that \ours surprisingly achieves the best forecasting performance in most cases (7 out of 14). Specifically, \ours demonstrates a relative average MSE reduction of approximately 6\% compared to \moirai{Small} and \moirai{Large}, and performs comparably to \moirai{Base}. When compared to the various few-shot baselines, \ours shows a relative average MSE reduction ranging from 8\% to 84\%. Given that all baselines except for \ours are trained on the time-series domain, this result is particularly encouraging. It suggests that \textbf{the transferability from images to time-series is stronger than from text to time-series, and even comparable to the in-domain transferability between time-series}. We also include a comparison with two TSF foundation models, TimesFM \citep{TimesFM} and LLMTime \citep{LLMTime}, in \cref{sec:app_zs_timesfm}, as well as traditional algorithms (ETS, ARIMA, and Seasonal Naïve) in \cref{sec:app_zs_traditional}. Results show that \ours still outperforms all of these baselines.

\paragraph{Results on GIFT-Eval Benchmarks} \cref{fig:zs_gift_aggregated} shows the comparison of \ours with six previously published TSF foundation models on the GIFT-Eval TSF Leaderboard\footnote{https://huggingface.co/spaces/Salesforce/GIFT-Eval}, where \ours surprisingly ranked first in terms of normalized MASE. It should be noted that although some concurrent works (after the release of \ours) in the current leaderboard outperform \ours, there may be data leakage issues for these works. In contrast, visual MAE was trained on ImageNet, long before the release of GIFT-Eval leaderboard, which can ensure no leakage.

\paragraph{Results on Monash Benchmark} 
\cref{fig:zs_monash_aggregated} shows the results aggregated from 29 Monash datasets, showing that \ours in the zero-shot setting surpasses all models \textit{individually} trained on each dataset and significantly outperforms the other cross-domain baseline (\ie, LLMTime). It achieves second place among all baselines, just behind \moirai{} that pre-trained on \textit{all} the training datasets. 
This promising result highlights VisionTS's strong zero-shot forecasting ability and effective cross-modality transferability.

\subsection{Further Analysis of \ours}

\paragraph{Backbone Analysis} In \cref{tab:zs_short_size} (full results in \cref{sec:app_zs_size}), we observe that the overall performance of three \mae variants (112M, 330M, and 657M) outperforms \moirai{Small} and \moirai{Large}. Particularly, larger models show a slight decrease in performance. This may be due to \textbf{larger visual models overfitting image-specific features, reducing their transferability}. A similar phenomenon was reported in \moirai{}, where larger models were found to degrade performance. We leave the exploration of scaling laws in image-based TSF foundation models for the future. Additionally, to explore the potential with other vision models, we also test LaMa \citep{LaMa}, a visual inpainting model. Results in \cref{sec:app_zs_size} demonstrate that \ours with LaMa performs similarly to \moirai{} in the zero-shot setting. This suggests that the performance is driven by the inherent similarity between images and time series, not solely by the \mae model.

\paragraph{Computational Cost}

We evaluate the computation cost of different baselines on an NVIDIA A800 GPU. Results are averaged on 90 runs. 
\cref{tab:zs_short_computation} shows the results between various TSF foundation models, showing that \ours are comparable to \moirai{Base} and GPT4TS and faster than TimesFM, which is an auto-regressive model. While computation time increases with context length for all the other Transformer-based baselines, \ours remains nearly constant. This is because \ours encodes input sequences into an image with constant size, ensuring $O(1)$ efficiency. In contrast, Transformer-based methods operate at $O(L^2)$ relative to context length $L$.

\input{figure/visual_embedding}
\input{table/tab_full_shot_short}

\paragraph{Hyperparameter Analysis} 

\cref{sec:app_zs_param} illustrates the impact of three hyperparameters. For context length $L$, as shown in \cref{fig:main_hyper_l}, performance typically improves with increasing $L$, particularly on high-frequency datasets like Weather (10-minute frequency) and ETTm1/ETTm2 (15-minute frequency). This aligns with other TSF foundation models like \moirai{}. As for the normalization constant $r$ and alignment constant $c$, when both of them are around 0.4, performance is generally well across most benchmarks.

\paragraph{Modality Analysis: Where does the zero-shot forecastability come from?} We further examine the gap between time series and images to explain the transferability of zero-shot forecasting. We sampled 1,000 images from ImageNet-1k and 300 samples from each time series dataset. We fed them into the \mae, maintaining a consistent image mask across all data. \cref{fig:visual_emb} visualizes the \mae encoder outputs of these data, which are flattened and reduced to 2-dimension by t-SNE. Notably, some time series, such as ETTm1 and Electricity, fall within the ImageNet distribution. \textbf{It suggests a relatively small gap between images and some time series (\eg, Electricity and ETTm1), which could explain the good transferability}. Additionally, while ImageNet displays a concentrated distribution, time series are generally more scattered. For instance, ETTm1 clusters in the upper right, whereas Monash is found in the lower left, with a significant gap. \textbf{This indicates strong heterogeneity within time series data and suggests that images may serve as a bridge to connect isolated time series modality}. 

\paragraph{Ablation Study}

We conduct experiments to validate our choices in the Alignment step, detailed in \cref{sec:app_zs_image}. First, we test three different interpolation strategies, which shows that \textbf{Bilinear interpolation performs best}. Second, we apply horizontal and vertical flips on the image to examine whether the assumed left-to-right, top-to-bottom order is efficient. Results show that these changes do not significantly affect performance, suggesting that \textbf{image reconstruction is isotropic and not influenced by certain orientation}.

\paragraph{Qualitative Analysis: When does \ours perform well, and when does it not?} In \cref{sec:app_visual}, we visualize the zero-shot forecasting of \ours alongside the input and reconstruction images, highlighting both \textit{successful} cases (where \ours outperforms \moirai{}) and \textit{failures} (where \moirai{} prevails). When the input exhibits strong regularity (\cref{fig:etth1_case1}), \ours effectively forecasts both the periodicity (via segmentation) and trends (via \mae's capabilities). In contrast, \moirai{}, akin to seasonal naïve methods, struggles to capture inter-period trends. For less-structured input (\cref{fig:etth2_case1,fig:etth2_case2,fig:etth1_case_bad}), \moirai{} adopts a conservative approach with lower volatility to minimize errors, while \ours takes a more aggressive stance. This strategy occasionally yields more accurate trend predictions (\cref{fig:etth2_case1,fig:etth2_case2}) but may also result in greater MAE (\cref{fig:etth1_case_bad}).

\subsection{Full-Shot Long-Term Time Series Forecasting}\label{sec:exp_full}

\paragraph{Setups} We evaluate the full-shot capability of each baseline trained on individual long-term TSF benchmarks. In addition to the six datasets used for zero-shot forecasting, we also include the popular Traffic and Illness datasets. As self-attention and feed-forward layers contain rich knowledge that can be transferred to TSF, we choose to \textbf{fine-tune only the layer normalization (LN) layers while freezing the other parameters}, which is also adopted by \citet{GPT4TS}. Training details are elaborated in \cref{sec:app_training_details}.

\paragraph{Main Results} \cref{tab:full_shot_short} summarizes the full-shot results, with the full results and standard deviations detailed in \cref{sec:app_fs_std}. It shows that \ours outperforms other baselines in most cases (46 out of 80), surpassing the non-pretrained PatchTST and the language-pretrained GPT4TS. Remarkably, except for Illness with the least data, \ours demands \textbf{only a single epoch of fine-tuning}. This suggests that even minimal fine-tuning enables VisionTS to adapt to time series effectively. Compared with \cref{tab:zero_shot_avg}, fine-tuning provides limited benefits for ETTh1 and ETTh2 but significantly improves other datasets. We attribute this to the smaller data scale of ETTh1 and ETTh2.

\paragraph{Ablation Study} \citet{AreLLMUseful} proposed several ablation variants for text-based foundation models, including \textbf{w/o LLM} (removing the LLM), \textbf{LLM2Attn}/\textbf{LLM2Trsf} (replacing the LLM with a single self-attention/Transformer layer), and \textbf{RandLLM} (randomly initializing the LLM). They found no significant performance differences and concluded that textual knowledge is unnecessary for TSF. We conducted similar ablations to assess the role of the vision model (VM), including \textbf{w/o VM}, \textbf{VM2Attn}, \textbf{VM2Trsf}, and \textbf{RandVM}. \cref{sec:app_fs_ablation} shows that these variants lead to worse performance, indicating that visual knowledge is beneficial for TSF.

\paragraph{Analysis: Fine-tuning strategies} As stated before, we fine-tune only the layer normalization (LN). We also tested fine-tuning the bias, MLP, or attention layers, in addition to full fine-tuning and freezing. All hyperparameters were kept constant. Note that freezing differs from the previous zero-shot experiment, where a longer context length was used. \cref{sec:app_fs_ablation} show that fine-tuning LN is the best. Modifying MLP or attention layers results in significant performance drops, suggesting that valuable knowledge resides in these components.

%% file: table/tab_zero_shot_exps.tex
\begin{table*}[t]
    \centering
    \begin{minipage}{0.35\textwidth}
        \centering
        \caption{Average MSE of different \mae variants, with full results in \cref{tab:zero_shot_size} (\cref{sec:app_zs_size}).}
        \vskip 0.15in
        \resizebox{\linewidth}{!}{
            \begin{tabular}{ccccccc}
            \toprule
            \multirow{2}[2]{*}{} & \quad\quad & \textbf{Base} & \quad\quad & \textbf{Large} & \quad\quad & \textbf{Huge} \\
                  & & 112M & & 330M & & 657M \\
            \midrule
            ETTh1 &  & 0.390  & & \textbf{0.378} & & 0.391  \\
            ETTh2 &  & \textbf{0.333} & & 0.340  & & 0.339  \\
            ETTm1 &  & \textbf{0.374} & & 0.379  & & 0.383  \\
            ETTm2 &  & \textbf{0.282} & & 0.286  & & 0.284  \\
            Electricity &  & 0.207 & & 0.209  & & \textbf{0.202}  \\
            Weather &  & \textbf{0.269} & & 0.272  & & 0.292  \\
            \midrule
            \rowc Avg.  & & \textbf{0.309} & & 0.311  & & 0.315  \\
            \bottomrule
            \end{tabular}
        }
        \label{tab:zs_short_size}
    \end{minipage}
    \hspace{0.7cm}
    \begin{minipage}{0.51\textwidth}
        \centering        \caption{Computational cost in terms of seconds for forecasting a batch of 32 time series data.}
        \vskip 0.15in
        \resizebox{\linewidth}{!}{

    \begin{tabular}{lccccccccc}
    \toprule
    \textbf{Context Length} & \multicolumn{4}{c}{\textbf{1k}} &      & \textbf{1k} & \textbf{2k} & \textbf{3k} & \textbf{4k}   \\
\cmidrule{2-5}\cmidrule{7-10}    \textbf{Prediction Length} & \textbf{1k} & \textbf{2k} & \textbf{3k} & \textbf{4k} &      & \multicolumn{4}{c}{\textbf{1k}}\\
    \midrule
    {PatchTST}  & 0.01 & 0.01 & 0.01 & 0.01 & & 0.01 & 0.02 & 0.03 & 0.04  \\
    {DeepAR}  & 0.26 & 0.32 & 0.37 & 0.43 & & 0.26 & 4.06 & 6.10 & 8.17   \\
    GPT4TS  & 0.01 & 0.01 & 0.01 & 0.02 & & 0.01 & 0.03 & 0.04 & 0.06  \\
    \moirai{Base} & 0.03  & 0.04  & 0.04  & 0.05 && 0.03  & 0.04  & 0.05  & 0.06       \\
    TimesFM & 0.08 & 0.14 & 0.20 & 0.27 && 0.07 & 0.13 & 0.20 & 0.25 \\
    LLMTime (8B) & \multicolumn{4}{c}{> 200} & & \multicolumn{4}{c}{> 200}\\
    \midrule
    \rowc \ours ($c=0.4$)  & 0.04  & 0.03  & 0.03  & 0.03 & & 0.04  & 0.04  & 0.05  & 0.05   \\
    \bottomrule
    \end{tabular}

      }

        \label{tab:zs_short_computation}
    \end{minipage}
\vskip -0.1in
\end{table*}

%% file: figure/visual_embedding.tex
\begin{figure*}[t]
    \centering
    \begin{minipage}[b]{0.43\linewidth}
        \centering
        \subfigure[Weather]{\includegraphics[width=0.48\linewidth]{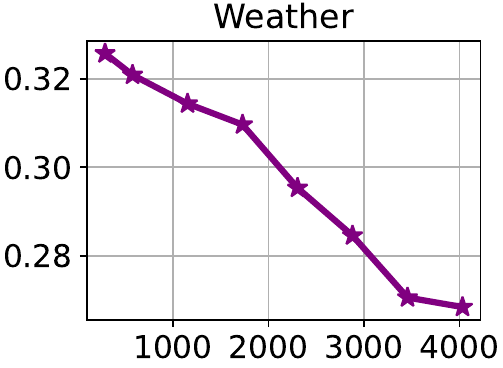}}\hfill
        \subfigure[ETTm1]{\includegraphics[width=0.48\linewidth]{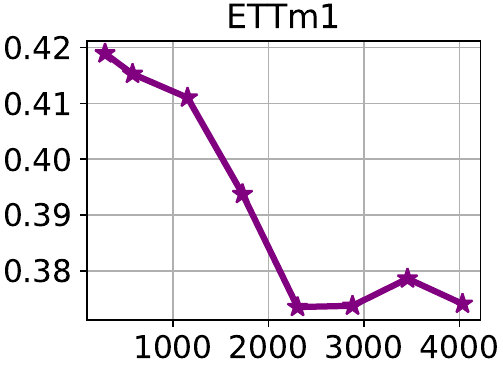}}\\
        \subfigure[ETTm2]{\includegraphics[width=0.48\linewidth]{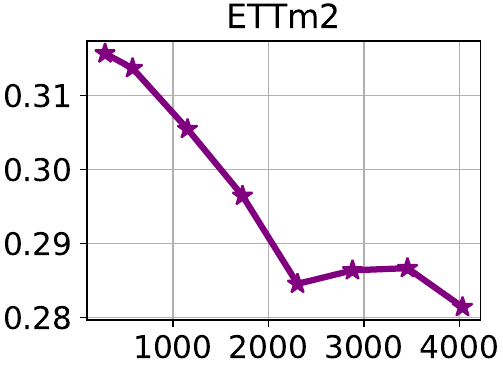}}\hfill
        \subfigure[Electricity]{\includegraphics[width=0.48\linewidth]{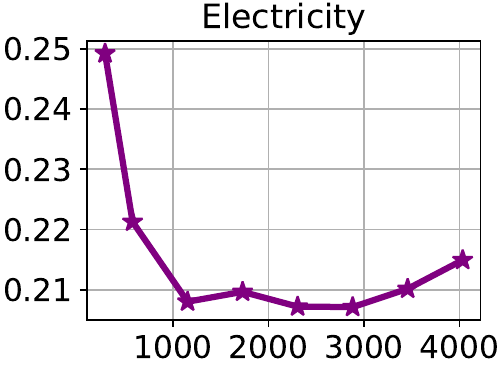}}\hfill
        \caption{MSE (Y-axis) performance of different context lengths $L$ (X-axis), averaged on four prediction lengths.}
        \label{fig:main_hyper_l}
    \end{minipage}
    \hfill
    \begin{minipage}[b]{0.53\linewidth}
        \centering
        \includegraphics[width=\linewidth]{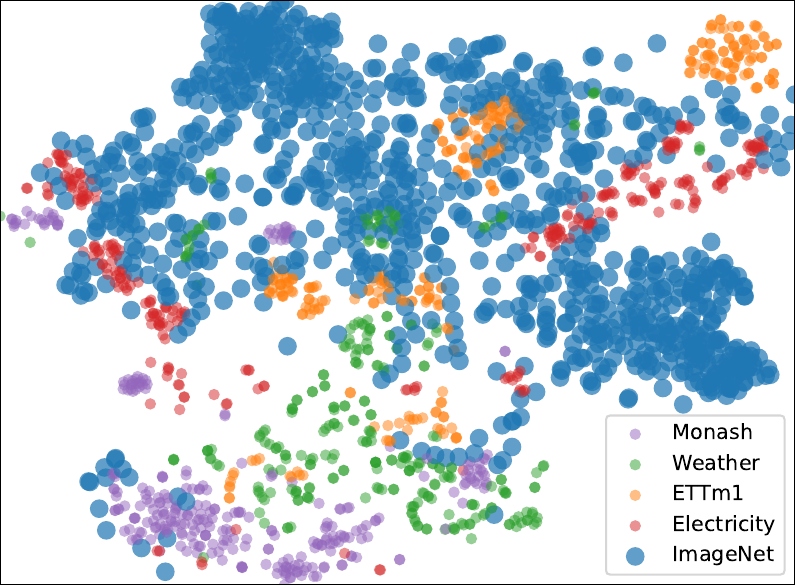}
        \caption{Modality visualization of the images (ImageNet) and time series (Monash, Weather, Electricity, and ETTm1) based on the \mae encoder.}
        \label{fig:visual_emb}
    \end{minipage}
\end{figure*}

%% file: table/tab_full_shot_short.tex
\begin{table*}[t]
  \centering
  \caption{Aggregated full-shot forecasting performance on eight long-term TSF benchmarks (ETTh1, ETTh2, ETTm1, ETTm2, Illness, Weather, Traffic, and Electricity). \ours is fine-tuned only a single epoch on each dataset except for Illness. Due to the space limit, we report the $1^{\text{st}}$ count for each baseline, with full results in \cref{tab:full_shot} (\cref{sec:app_fs_std}).}
\vskip 0.15in
  
    \resizebox{\linewidth}{!}{
    \begin{tabular}{ccccccccccccccc}
    \toprule
    \textbf{Pretrain} &      & \textit{\textbf{\emoji{figure/em-frame_with_picture} Images}} &      & \multicolumn{2}{c}{\textit{\textbf{\emoji{figure/em-memo} Text}}} &      & \multicolumn{8}{c}{\textit{\textbf{\emoji{figure/em-no_entry_sign} No Pretrain}}} \\
\cmidrule{3-3}\cmidrule{5-6}\cmidrule{8-15}    \textbf{Method} &      & \ours &      & Time-LLM & GPT4TS &      & Dlinear & PatchTST & TimesNet & \multicolumn{1}{l}{FEDformer} & \multicolumn{1}{l}{Autoformer} & \multicolumn{1}{l}{Stationary} & \multicolumn{1}{l}{ETSformer} & \multicolumn{1}{l}{Informer} \\
    \rowc \boldmath{}\textbf{$\bf 1^{st}$ count}\unboldmath{} &      & \textbf{46}   &      & 4    & 12   &      & 0    & 19   & 0    & 0    & 0    & 0    & 0    & 0 \\
    \bottomrule
    \end{tabular}%
  \label{tab:full_shot_short}%
  }
\end{table*}%

%% file: section/related-work.tex
\section{Related Work}

Depending on the pre-training data, TSF foundation models can be categorized into Text-based and TS-based models. We first review related works and then introduce recent research on image-based time series analysis.

\paragraph{Text-based TSF Foundation Models} Large Language Models (LLMs) pre-trained on large amounts of text data are being applied to TSF tasks. For example, \citet{GPT4TS} fine-tuned a pre-trained GPT \citep{GPT-2} on each time-series downstream task, such as forecasting, classification, imputation, and anomaly detection. Based on Llama \citep{llama}, \citet{timellm} froze the pre-trained LLM and reprogrammed the time series to align with the language modality. \citet{aLLM4TS} adopted a two-stage approach by continually pre-training GPT \citep{GPT-2} on the time-series domain. Nevertheless, the TSF performance of LLMs has recently been questioned by \citet{AreLLMUseful}, which designed several ablation studies to show that textual knowledge is unnecessary for forecasting. In this paper, we attribute it to the large modality gap. Some recent approaches focus on directly transforming the time series into natural texts for LLMs, allowing for zero-shot forecasting. For example, PromptCast \citep{PromptCast} used pre-defined templates to describe numerical time series data, while LLMTime \citep{LLMTime} directly separated time steps using commas and separates digits using spaces to construct the text input. However, due to the efficiency issue of the autoregressive decoding strategy and the expensive inference cost of large language models, their practical use is limited.

\paragraph{Time Series-Based TSF Foundation Models}

Self-supervised pre-training a TSF model on the same dataset used for downstream TSF tasks is a well-explored topic \citep{Pretraining-Survey,Self-Supervised-Survey}, such as denoising autoencoders \citep{Denoising-Autoencoders} or contrastive learning \citep{CoST,TS2Vec}. They follow a similar paradigm to the masked autoencoder (\mae) in computer vision, which is a well-studied topic in other machine learning fields, such as BERT \citep{BERT}, CBraMod \citep{CBraMod}, and HuBERT \citep{HuBert}. However, these methods rarely examine the cross-dataset generalization capabilities. Recently, research has shifted towards training universal foundation models, by collecting large-scale time series datasets from diverse domains \citep{Chronos,Moment,Timer,TimesFM,TimeSiam,GTT} or generating numerous synthetic time series data \citep{SyntheticTS,ViTime}. As a representative method, \citet{Moirai} collected 27 billion observations across nine domains and trained TSF foundation models of various scales, achieving strong zero-shot performance. However, given the severe heterogeneity, constructing high-quality large datasets poses significant challenges for building these foundation models.

\paragraph{Image-Based Time-Series Analysis} Previous research has investigated encoding time series data into images and used convolutional neural networks (CNNs) trained from scratch for classification \citep{wang2015imaging,wang2015spatially,hatami2018classification} or forecasting \citep{li2020forecasting,sood2021visual,semenoglou2023image}. Recent researchers explored using pre-trained models for these imaging time series. \citet{li2024time} used a pre-trained vision transformer (ViT) for classification. \citet{wimmer2023leveraging} and \citet{zhang2023insight} employed vision-language multimodal pre-trained models to extract predictive features and generate text descriptions. \citet{ViTime} generated synthetic time series data to pre-train a vision model for the TSF task. However, these studies did not deeply examine the transferability from natural images to TSF. Despite early efforts by \citet{GPT4TS} to fine-tune a BEiT \citep{BEiT} trained on images for time series forecasting, it still falls short of the leading text-based and TS-based TSF foundation models. To the best of our knowledge, we are the first to show that an image-based foundation model, without further time-series adaptation, can match or even surpass other types of TSF foundation models.

%% file: section/conclusion.tex
\section{Conclusion}

In this paper, we explore a novel approach to building a time series forecasting (TSF) foundation model using natural images, offering a new perspective distinct from the traditional text-based and TS-based methods. By leveraging the intrinsic similarities between images and time series, we introduced \ours, an \mae-based TSF foundation model that reformulates the TSF task as an image reconstruction problem. Our extensive evaluations demonstrate that \ours achieves outstanding forecasting performance in zero-shot and full-shot settings, being a free lunch for a TSF foundation model. We hope our findings could open new avenues for further cross-modality research.

\paragraph{Limitations and Future Directions.} (1) As a preliminary study, we employed \mae and LaMa. Utilizing more advanced models like diffusion models \citep{Latent-Diffusion,DiT} presents a promising research direction. (2) Due to limitations in the visual model, \ours cannot capture multivariate interactions and perform distribution forecasting. Future modifications to the model structure may empower it with more time series capabilities.

\section*{Acknowledgments}
Research work mentioned in this paper is supported by State Street Zhejiang University Technology Center. We would also like to thank reviewers for their valuable comments.

%% file: section/appendix.tex
\newpage

\onecolumn

\numberwithin{equation}{section}

\section*{Appendix}

\section{Details of Experiments}

\subsection{Benchmark and baselines}\label{sec:app_exp_detail}

\paragraph{Long-Term TSF Benchmark} We evaluate our model on 8 widely used long-term TSF datasets \citep{Informer, Autoformer}, including ETTh1, ETTh2, ETTm1, ETTm2, Electricity, Traffic, Illness, and Weather. Performance is assessed using Mean Squared Error (MSE) and Mean Absolute Error (MAE), with lower values indicating better forecasting accuracy.

\paragraph{Monash Benchmark} Following \citet{Moirai}, we tested 29 Monash datasets \citep{Monash} using GluonTS \citep{GluonTS}, including M1 Monthly, M3 Monthly, M3 Other, M4 Monthly, M4 Weekly, M4 Daily, M4 Hourly, Tourism Quarterly, Tourism Monthly, CIF 2016, Australian Electricity Demand, Bitcoin, Pedestrian Counts, Vehicle Trips, KDD Cup, Weather, NN5 Daily, NN5 Weekly, Carparts, FRED-MD, Traffic Hourly, Traffic Weekly, Rideshare, Hospital, COVID Deaths, Temperature Rain, Sunspot, Saugeen River Flow, and US Births. Performance is assessed using MAE.

\paragraph{GIFT-Eval Benchmark} \citet{GIFT-Eval} introduces the General Time Series Forecasting Model Evaluation, GIFT-Eval, encompasses 23 datasets over 144,000 time series and 177 million data points, spanning seven domains, 10 frequencies, multivariate inputs, and prediction lengths ranging from short to long-term forecasts. We use a constant context length 2,000 for \ours and we report the point forecast performance using MAPE.

\paragraph{Baselines} We select representative baselines for comparison, including \textbf{TS-based} and \textbf{Text-based} foundation models, and \textbf{other popular TSF baselines} covering both Transformer-based, MLP-based and CNN-based architectures. The baseline models selected for comparison are briefly described below:

\begin{enumerate}[leftmargin=*]
\item \textbf{\moirai{}} \citep{Moirai} is a TSF foundation model trained on the Large-scale Open Time Series Archive (LOTSA), with over 27B observations across nine domains. It has three variants: \textbf{small}, \textbf{base}, and \textbf{large}.
\item \textbf{TimesFM} \citep{TimesFM} is a decoder-style TSF foundation model, using a large time-series corpus comprising both real-world and synthetic datasets.
\item \textbf{Time-LLM} \citep{timellm} is a text-based TSF foundation model built on Llama, which reprograms time series data to align with the language modality, keeping the LLM frozen.
\item \textbf{GPT4TS} \citep{GPT4TS} (OneFitsAll) is another text-based model based on GPT, fine-tuned for forecasting tasks.
\item \textbf{LLMTime} \citep{LLMTime} encodes time series data to a text sequence, supporting zero-shot forecasting.
\item \textbf{DLinear} \citep{LTSF-Linear} proposes a linear forecasting model, enhanced by seasonal-trend decomposition or normalization.
\item \textbf{PatchTST} \citep{PatchTST} uses Transformer encoders with patching and channel independence techniques for improved predictions.
\item \textbf{TimesNet} \citep{TimesNet} applies convolution kernels along the time dimension, using temporal decomposition and periodical segmentation to capture temporal patterns.
\item \textbf{FEDformer} \citep{FEDformer} employs a sparse frequency domain representation, using frequency-enhanced blocks for cross-time dependency.
\item \textbf{Autoformer} \citep{Autoformer} uses series decomposition blocks and Auto-Correlation to capture cross-time dependency.
\item \textbf{Stationary} \citep{Non_stationary_Transformer} introduces stationarization and de-stationary attention mechanisms.
\item \textbf{ETSFormer} \citep{ETSformer} leverages exponential smoothing principles, including exponential smoothing and frequency attention mechanisms.
\item \textbf{Informer} \citep{Informer} proposes ProbSparse self-attention and distillation operations.
\end{enumerate}

For the long-term TSF benchmark, we include TS-based foundation model results from their original papers, Text-based model results from \citet{AreLLMUseful}, and other baseline results from \citet{GPT4TS}. For the Monash and PF benchmark, we include results from \citet{Moirai}.

\input{table/tab_possible_p}

\input{table/tab_final_p}

\input{table/tab_p}

\paragraph{Environment} All experiments are conducted using \textit{Time-Series-Library} (\url{https://github.com/thuml/Time-Series-Library}) and GluonTS library \citep{GluonTS} on an NVIDIA A800 GPU.

\subsection{Periodicity selection}\label{sec:app_p}

We first determine a range of period lengths based on the sampling frequency of the data, shown in \cref{tab:possible_p}. This frequency-based strategy is also employed by \citet{GluonTS} while we extend the search range for tuning. We select the optimal $P$ from this range on the validation set. The final $P$ used in our experiments are summarized in \cref{tab:final_p}.

To demonstrate the influence of $P$ and the effectiveness of our periodicity selection strategy, we set $P=1$ and compare the results with the above strategy. \cref{tab:p_eq_1} shows that such strategy (denoted as \ours) significantly outperforms the naive strategy that sets $P=1$.

\section{Zero-Shot Forecasting}

\subsection{Hyperparameters}\label{sec:app_zs_tune}

\input{table/tab_zero_shot_param}

We conduct hyperparameter tuning on validation sets to determine the optimal context length $L$. Final used hyperparameters are summarized in \cref{tab:zero_shot_tune}.

\subsection{Full forecasting results of the long-term TSF benchmark}\label{sec:app_zs_full}

\input{table/tab_zero_shot_full}

\cref{tab:zero_shot_full} shows the full results of zero-shot/few-shot long-term forecasting performance. \ours achieves the best results in most cases (32 out of 62), outperforming \moirai{Base} (10 out of 62) and \moirai{Large} (8 out of 62). 

\subsection{Comparison of TimesFM and LLMTime}\label{sec:app_zs_timesfm}

\input{table/tab_zero_shot_timesfm}

\input{table/tab_zero_shot_seasonal}

Due to the step-by-step output of the decoder architecture, the efficiency of TimesFM \citep{TimesFM} and LLMTime \citep{LLMTime} are relatively slower. Thus, \citet{TimesFM} only reported results for the last test window of the original split. We compared \ours with their results under the same setting, as shown in \cref{tab:zero_shot_timesfm}. \ours outperforms TimesFM and LLMTime in terms of MAE, indicating that image-based TSF models are on par with or even better than TS-based and text-based models.

\subsection{Comparison of traditional methods}\label{sec:app_zs_traditional}

In addition to deep learning models, we also compare traditional methods, including ARIMA, ETS, and two methods that require periodicity as our \ours: Seasonal Naïve (repeating the last period) and Seasonal Avg (similar to Seasonal Naïve but repeating the average of all periods in the look-back window). Due to the high computational cost of ARIMA and ETS, we only compare them on the small-scale benchmarks, \ie, four ETT datasets. \cref{tab:zero_shot_season} shows that \ours also achieves the best performance.

\subsection{Comparison of concurrent works}

\begin{table*}[h!]
  \centering
  \caption{Comparison of Time-MoE and TTM in the \textbf{zero-shot} setting. We report the base and large model results for Time-MoE, as the ultra model weights are not yet released. For TTM, we used the official HuggingFace model for replication. The following table summarizes the performance of various zero-shot foundation models.}
  
  \small

    \begin{tabular}{ccccccc}
    \toprule
    \multicolumn{3}{c}{} & \textbf{\ours} & \textbf{Time-MoE (base)} & \textbf{Time-MoE (large)} & \textbf{TTM (v1)} \\
    \midrule
    \multicolumn{2}{c}{\multirow{2}[2]{*}{ETTh1}} & MSE  & \textbf{0.390} & 0.400  & 0.394  & 0.398  \\
    \multicolumn{2}{c}{} & MAE  & \textbf{0.414} & 0.424  & 0.419  & 0.421  \\
    \midrule
    \multicolumn{2}{c}{\multirow{2}[1]{*}{ETTh2}} & MSE  & \textbf{0.333} & 0.366  & 0.405  & 0.348  \\
    \multicolumn{2}{c}{} & MAE  & \textbf{0.375} & 0.404  & 0.415  & 0.393  \\
    \midrule
    \multicolumn{2}{c}{\multirow{2}[1]{*}{ETTm1}} & MSE  & \textbf{0.374} & 0.394  & 0.376  & 0.520  \\
    \multicolumn{2}{c}{} & MAE  & \textbf{0.372} & 0.415  & 0.405  & 0.479  \\
    \midrule
    \multicolumn{2}{c}{\multirow{2}[1]{*}{ETTm2}} & MSE  & \textbf{0.282} & 0.317  & 0.316  & 0.312  \\
    \multicolumn{2}{c}{} & MAE  & \textbf{0.321} & 0.365  & 0.361  & 0.348  \\
    \midrule
    \multicolumn{2}{c}{\multirow{2}[1]{*}{Electricity}} & MSE  & 0.207  & (data leakage) & (data leakage) & \textbf{0.201} \\
    \multicolumn{2}{c}{} & MAE  & 0.294  & (data leakage) & (data leakage) & \textbf{0.293} \\
    \midrule
    \multicolumn{2}{c}{\multirow{2}[2]{*}{Weather}} & MSE  & 0.269  & 0.265  & 0.270  & \textbf{0.234} \\
    \multicolumn{2}{c}{} & MAE  & 0.292  & 0.297  & 0.300  & \textbf{0.266} \\
    \midrule
    \multicolumn{2}{c}{\multirow{2}[2]{*}{\textbf{Average}}} & MSE  & \textbf{0.309} & -    & -    & 0.335  \\
    \multicolumn{2}{c}{} & MAE  & \textbf{0.345} & -    & -    & 0.367  \\
    \midrule
    \rowc \multicolumn{3}{c}{\textbf{1$^{\text{st}}$ count}} & \textbf{10} & 0    & 0    & 4 \\
    \bottomrule
    \end{tabular}%
  
  \label{tab:concurrent_zeroshot}%
\end{table*}%

\begin{table*}[h!]
  \centering
  \caption{Comparison of CALF in both zero-shot and full-shot settings.}
  
  \small

    \begin{tabular}{cccccc}
    \toprule
    \multicolumn{3}{c}{} & \textbf{\ours (zero-shot)} & \textbf{\ours (full-shot)} & CALF \\
    \midrule
    \multicolumn{2}{c}{\multirow{2}[1]{*}{ETTh1}} & MSE  & \textbf{0.390} & 0.395  & 0.432  \\
    \multicolumn{2}{c}{} & MAE  & 0.414  & \textbf{0.409} & 0.428  \\
    \multicolumn{2}{c}{\multirow{2}[0]{*}{ETTh2}} & MSE  & \textbf{0.333} & 0.336  & 0.349  \\
    \multicolumn{2}{c}{} & MAE  & \textbf{0.375} & 0.382  & 0.382  \\
    \multicolumn{2}{c}{\multirow{2}[0]{*}{ETTm1}} & MSE  & 0.374  & \textbf{0.338} & 0.395  \\
    \multicolumn{2}{c}{} & MAE  & 0.372  & \textbf{0.367} & 0.390  \\
    \multicolumn{2}{c}{\multirow{2}[0]{*}{ETTm2}} & MSE  & 0.282  & \textbf{0.261} & 0.281  \\
    \multicolumn{2}{c}{} & MAE  & 0.321  & \textbf{0.319} & 0.321  \\
    \multicolumn{2}{c}{\multirow{2}[0]{*}{Electricity}} & MSE  & 0.207  & \textbf{0.156} & 0.175  \\
    \multicolumn{2}{c}{} & MAE  & 0.294  & \textbf{0.249} & 0.265  \\
    \multicolumn{2}{c}{\multirow{2}[1]{*}{Weather}} & MSE  & 0.269  & \textbf{0.227} & 0.250  \\
    \multicolumn{2}{c}{} & MAE  & 0.292  & \textbf{0.262} & 0.274  \\
    \midrule
    \multicolumn{2}{c}{\multirow{2}[2]{*}{\textbf{Average}}} & MSE  & 0.309  & \textbf{0.286} & 0.314  \\
    \multicolumn{2}{c}{} & MAE  & 0.345  & \textbf{0.331} & 0.343  \\
    \bottomrule
    \end{tabular}%
  
  \label{tab:concurrent_calf}%
\end{table*}%

We compare our work with other concurrent TSF methods. \cref{tab:concurrent_zeroshot} presents the comparison from Time-MoE \citep{Time-MoE} and TTM \citep{TTM}, and \cref{tab:concurrent_calf} shows the comparison with CALF \citep{CALF}, which is the existing SOTA LLMs-based time series forecasting work. These findings highlight the promising potential of vision models in TSF scenarios.

\subsection{Full forecasting results of the Monash TSF benchmark}\label{sec:app_zs_monash}

\input{table/tab_zero_shot_monash}

\paragraph{Setup} \cref{tab:final_p} lists the sampling frequency and the selected period $P$ for each dataset. Datasets with $P=1$ indicate no significant periodicity, where we use a context length of $L=300$. For other datasets with $P>1$, we select a longer context length of $L=1000$. All datasets were tested with the hyperparameters $r=c=0.4$ as we had done for the long-term TSF benchmark.

\paragraph{Results} \cref{tab:zero_shot_monash} presents \ours’s MAE test results, with the normalized MAE calculated by dividing each dataset's MAE by the naive forecast’s MAE and aggregated using the geometric mean across datasets. We include the result of each baseline from \citet{Moirai}. Particularly, we find that \ours outperforms \moirai{} on some datasets with $P=1$ (\eg, FRED-MD and NN5 Weekly), showing that \ours can still work effectively without significant periodicity.

\input{table/tab_lama}

\subsection{Impact of backbones}\label{sec:app_zs_size}

\input{table/tab_zero_shot_size}

\cref{tab:zero_shot_size} compares zero-shot forecasting performance of three \mae variants (112M, 330M, and 657M), showing that the three variants are similar, but larger models show a slight decrease. Particularly, the smallest model excels in ETTh2, ETTm1, ETTm2, and Weather, while the largest model excels in Electricity. Additionally, \cref{tab:lama} compares \ours with another visual backbone, LaMa.

\subsection{Impact of the different image encoding strategies}\label{sec:app_zs_image}

\input{table/tab_image_ablation}

\cref{tab:zero_shot_ablation} summarizes the impact of interpolation strategies and image orientations in the Alignment step. It shows that the smoother Bilinear and Bicubic interpolation perform similarly, both significantly better than the rougher Nearest Neighbor. This suggests that smooth resizing effectively handles time series interpolation. Moreover, image orientation has little impact on performance.

\subsection{Hyperparameter analysis}\label{sec:app_zs_param}

\cref{fig:ratio,fig:scalex,fig:sl} show the influence of three hyperparameters, $r$, $c$, and $L$. We report the MSE averaged on four prediction lengths \{96, 192, 336, 720\}. 

\clearpage

\input{figure/ratio/exp_ratio}

\input{figure/scalex/exp_scalex}

\input{figure/sl/exp_sl}

\clearpage

\section{Full-Shot Forecasting}

\subsection{Training details}\label{sec:app_training_details}

\input{table/tab_full_shot_param}

Based on the principle of channel independence \citep{PatchTST,ChannelIndependence}, we treat the variables of each time series as individual data samples. We use an Adam optimizer with a learning rate 0.0001 and a batch size 256 to fine-tune \mae. All experiments are repeated three times. The training epoch is one for all the datasets except Illness, for which we train \mae for 100 epochs with an early stop due to the limited training dataset scale. We conduct tuning on validation sets for the three hyperparameters, $r$, $c$, and $L$. The final hyperparameters used are summarized in \cref{tab:full_shot_tune}.

\subsection{Full results and standard deviations}\label{sec:app_fs_std}

\input{table/tab_full_shot_std}

\input{table/tab_full_shot}

\cref{tab:full_shot} shows the full results of the full-shot experiments. We also report the standard deviations of our full-shot experiments computed on three runs in \cref{tab:full_shot_std}, including the results of Time-LLM and GPT4TS from \citet{AreLLMUseful} for reference.

\subsection{Ablation study and fine-tuning strategy comparison}\label{sec:app_fs_ablation}

\input{table/tab_ablation}

We compare the following ablation variants to verify the role of the visual model (VM), similar to \citet{AreLLMUseful}. 

\begin{itemize}[leftmargin=*]
\item \textbf{w/o VM} removes all the transformer blocks in encoders and decoders.
\item \textbf{VM2Attn} replaces both the encoder and decoder with a self-attention layer, matching \mae structure but with random initialization.
\item \textbf{VM2Trsf} is similar to \textbf{VM2Attn} but replaces them with a Transformer block (\ie, a self-attention layer plus an MLP layer).
\item \textbf{Rand-VM} keeps the same architecture as the vanilla \mae, but all the weights are randomly initialized.
\end{itemize}

We also compare fine-tuning different components in \mae as follows:

\begin{itemize}[leftmargin=*]
\item \textbf{All} fine-tunes all the trainable weights in \mae.
\item \textbf{LN} fine-tunes only the layer normalization, which is the default setting used in our experiments.
\item \textbf{Bias} fine-tunes only the bias term of all the linear layers, proposed by \citet{BitFit}.
\item \textbf{MLP} and \textbf{Attn} fine-tune only the feed-forward layer and the self-attention layer, respectively.
\item \textbf{Freeze} does not fine-tune any weight.  Note that it differs from the previous zero-shot experiment, where a longer context length was used (see \cref{tab:zero_shot_tune} and \cref{tab:full_shot_tune}).
\end{itemize}

The results are shown in \cref{tab:full_shot_ablation}, suggesting that visual knowledge is crucial for \ours and fine-tuning the layer normalization is the best.

\section{Visualization}\label{sec:app_visual}

We visualized the predictions of \ours in the zero-shot setting, including its input and reconstructed images. We also visualized the predictions of \moirai{Large} and Seasonal Naïve, with their MAE metrics for comparison. \cref{fig:etth1_case1,fig:etth2_case1,fig:etth2_case2} show examples where \ours performed well, with \cref{fig:etth1_case1} depicting a more regular pattern, while \cref{fig:etth2_case1,fig:etth2_case2} display less obvious patterns. \cref{fig:etth1_case_bad} illustrates a case where \ours underperformed, as it aggressively predicted the trend despite the lack of clear patterns in the input sequence, whereas \moirai{Large} made more conservative predictions.

\input{figure/etth1_case1/etth1_case1}
\input{figure/etth2_case1/etth2_case1}
\input{figure/etth2_case2/etth2_case2}
\input{figure/etth1_case_bad/etth1_case_bad}

%% file: table/tab_possible_p.tex
\begin{table}[t]
  \centering
  \small
  \caption{Periodicity ($P$) search range for the sampling frequency. $x$ denotes the number of sampling frequencies. For example, for data with a sampling frequency of 2 minutes (2T), we have $x=2$, and the possible search range of $P$ is $\{\nicefrac{1440}{x}, \nicefrac{10080}{x}, 1\}=\{720, 5040, 1\}$.}
\vskip 0.15in
    \begin{tabular}{lll}
    \toprule
    \textbf{Sampling Frequency} & \textbf{Possible Seasonalities} & \textbf{Possible P} \\
    \midrule
    Second (S) & 1 hour & $\{\nicefrac{3600}{x}, 1\}$ \\
    Minute (T) & 1 day or 1 week & $\{\nicefrac{1440}{x}, \nicefrac{10080}{x}, 1\}$ \\
    Hour (H) & 1 day or 1 week & $\{\nicefrac{24}{x}, \nicefrac{168}{x}, 1\}$ \\
    Day (D) & 1 week, 1 month, or 1 year & $\{\nicefrac{7}{x}, \nicefrac{30}{x}, \nicefrac{365}{x}, 1\}$ \\
    Week (W) & 1 year or 1 month & $\{\nicefrac{52}{x}, \nicefrac{4}{x}, 1\}$ \\
    Month (M) & 1 year, 6 months, or 3 months & $\{\nicefrac{12}{x}, \nicefrac{6}{x}, \nicefrac{3}{x}, 1\}$ \\
    Business Day (B) & 1 week & $\{\nicefrac{5}{x}, 1\}$ \\
    Quarter (Q) & 1 year or 6 months & $\{\nicefrac{4}{x}, \nicefrac{2}{x}, 1\}$ \\
    Others & -    & $\{1\}$ \\
    \bottomrule
    \end{tabular}%
  \label{tab:possible_p}%
\end{table}%

%% file: table/tab_final_p.tex
\begin{table*}[t]
  \centering
  \caption{Final $P$ used for each dataset in our experiment.}
\vskip 0.15in
  
    \resizebox{0.8\linewidth}{!}{

    \begin{tabular}{cccllrr}
    \toprule
         & \textbf{Frequency} & \boldmath{}\textbf{$P$}\unboldmath{} & \textbf{Datasets} &      &      &  \\
    \midrule
    \multirow{4}[2]{*}{Long-Term TSF} & H    & 24   & ETTh1 & ETTh2 & \multicolumn{1}{l}{Electricity} & \multicolumn{1}{l}{Traffic} \\
         & W    & 52   & Illness &      &      &  \\
         & 15T  & 96   & ETTm1 & ETTm2 &      &  \\
         & 10T  & 144  & Weather &      &      &  \\
    \midrule
    \multirow{15}[2]{*}{Monash} & D    & 1    & M4 Daily & COVID Deaths &      &  \\
         & W    & 1    & NN5 Weekly &      &      &  \\
         & M    & 1    & FRED-MD &      &      &  \\
         & Q    & 1    & M3 Other  &      &      &  \\
         & M    & 3    & M3 Monthly & M4 Monthly & \multicolumn{1}{l}{CIF 2016 (6) } &  \\
         & W    & 4    & M4 Weekly & Traffic Weekly &      &  \\
         & Q    & 4    & Tourism Quarterly &      &      &  \\
         & M    & 6    & CIF 2016 (12)  & Car Parts &      &  \\
         & D    & 7    & Bitcoin & Vehicle Trips & \multicolumn{1}{l}{Weather} & \multicolumn{1}{l}{NN5 Daily} \\
         & D    & 7    & US Births & Saugeen Day & \multicolumn{1}{l}{Temperature Rain} &  \\
         & M    & 12   & Tourism Monthly & Hospital & \multicolumn{1}{l}{M1 Monthly} &  \\
         & H    & 24   & M4 Hourly & KDD cup & \multicolumn{1}{l}{Pedestrian Counts} &  \\
         & H    & 24   & Traffic Hourly & Rideshare &      &  \\
         & D    & 30   & Sunspot &      &      &  \\
         & 0.5H & 336  & Aus. Elec. Demand  &      &      &  \\
    \bottomrule
    \end{tabular}%
    
    }
  \label{tab:final_p}%
\end{table*}%

%% file: table/tab_p.tex
\begin{table}[t!]
  \centering
  \caption{Comparison of setting $P=1$ for \ours.}
\vskip 0.15in
  \small
    \begin{tabular}{cccrcc}
    \toprule
         & \multicolumn{2}{c}{\textbf{\ours}} &      & \multicolumn{2}{c}{\boldmath{}\textbf{$P=1$}\unboldmath{}} \\
\cmidrule{2-3}\cmidrule{5-6}         & MSE  & MAE  &      & MSE  & MAE \\
\cmidrule{1-3}\cmidrule{5-6}    ETTh1 & 0.390  & 0.414  &      & 0.840  & 0.628  \\
    ETTh2 & 0.333  & 0.375  &      & 0.424  & 0.445  \\
    ETTm1 & 0.374  & 0.372  &      & 0.660  & 0.533  \\
    ETTm2 & 0.282  & 0.321  &      & 0.312  & 0.363  \\
    \rowc \textbf{Average} & 0.344  & 0.370  &      & 0.559  & 0.492  \\
    \bottomrule
    \end{tabular}%
  \label{tab:p_eq_1}%
\end{table}%

%% file: table/tab_zero_shot_param.tex
\begin{table*}[ht]
  \centering
  \caption{Hyperparameters for \ours used in our zero-shot forecasting (Long-term TSF).}
\vskip 0.15in
  \small

    \begin{tabular}{ccccccc}
    \toprule
         & \textbf{ETTh1} & \textbf{ETTh2} & \textbf{ETTm1} & \textbf{ETTm2} & \textbf{Weather} & \textbf{Electricity} \\
    \midrule
    Normalization constant $r$ & 0.4  & 0.4  & 0.4  & 0.4  & 0.4  & 0.4  \\
    Alignment constant $c$ & 0.4  & 0.4  & 0.4  & 0.4  & 0.4  & 0.4  \\
    Context length $L$ & 2880 & 1728 & 2304 & 4032 & 4032 & 2880 \\
    \bottomrule
    \end{tabular}%
  
  \label{tab:zero_shot_tune}%
\end{table*}%

%% file: table/tab_zero_shot_full.tex
\begin{table}[ht]
  \centering
  \caption{Full results of \cref{tab:zero_shot_avg}: Zero-shot or few-shot results on the long-term TSF benchmark. \textbf{Bold}: the best result.}
\vskip 0.15in
    \resizebox{\linewidth}{!}{

\begin{tabular}{ccccccccccccccccccccccccccccccccccc}
\toprule
\multicolumn{2}{c}{} &   \quad\quad    & \multicolumn{11}{c}{\textbf{\emoji{figure/em-no_entry_sign} Zero-Shot}}                                    &   \quad\quad   & \multicolumn{20}{c}{\textbf{\emoji{figure/em-chart_with_upwards_trend} Few-Shot (10\% Downstream Dataset)}} \\
\cmidrule{4-14}\cmidrule{16-35}\multicolumn{2}{c}{\textbf{Pretrain}} &      & \multicolumn{2}{c}{\textit{\textbf{\emoji{figure/em-frame_with_picture} Images}}} &      & \multicolumn{8}{c}{{\textbf{\textit{\emoji{figure/em-chart_with_upwards_trend} Time-series}}}}      &      & \multicolumn{5}{c}{{\textbf{\textit{\emoji{figure/em-memo} Text}}}} &      & \multicolumn{14}{c}{\textbf{\textit{\emoji{figure/em-no_entry_sign} No Pretrain}}} \\
\cmidrule{4-5}\cmidrule{7-14}\cmidrule{16-20}\cmidrule{22-35}\multicolumn{2}{c}{\textbf{Method}} &      & \multicolumn{2}{c}{\textbf{\method{VisionTime}}} &      & \multicolumn{2}{c}{\textbf{\moirai{S}}} &      & \multicolumn{2}{c}{\textbf{\moirai{B}}} &      & \multicolumn{2}{c}{\textbf{\moirai{L}}} &      & \multicolumn{2}{c}{\textbf{TimeLLM}} &      & \multicolumn{2}{c}{\textbf{GPT4TS}} &      & \multicolumn{2}{c}{\textbf{DLinear}} &      & \multicolumn{2}{c}{\textbf{PatchTST}} &      & \multicolumn{2}{c}{\textbf{TimesNet}} &      & \multicolumn{2}{c}{\textbf{Autoformer}} &      & \multicolumn{2}{c}{\textbf{Informer}} \\
\cmidrule{4-5}\cmidrule{7-8}\cmidrule{10-11}\cmidrule{13-14}\cmidrule{16-17}\cmidrule{19-20}\cmidrule{22-23}\cmidrule{25-26}\cmidrule{28-29}\cmidrule{31-32}\cmidrule{34-35}
\multicolumn{2}{c}{\textbf{Metric}} &      & MSE & MAE &      & MSE & MAE &      & MSE & MAE &      & MSE & MAE &      & MSE & MAE &      & MSE & MAE &      & MSE & MAE &      & MSE & MAE &      & MSE & MAE &      & MSE & MAE &      & MSE & MAE \\
\midrule
\multirow{5}[2]{*}{\rotatebox{90}{$ETTh1$}} & \multicolumn{1}{r}{96} &      & \textbf{0.353} & \textbf{0.383} &      & 0.375  & 0.402  &      & 0.384  & 0.402  &      & 0.380  & 0.398  &      & 0.448  & 0.460  &      & 0.458  & 0.456  &      & 0.492  & 0.495  &      & 0.516  & 0.485  &      & 0.861  & 0.628  &      & 0.613  & 0.552  &      & 1.179  & 0.792  \\
     & \multicolumn{1}{r}{192} &      & \textbf{0.392} & \textbf{0.410} &      & 0.399  & 0.419  &      & 0.425  & 0.429  &      & 0.440  & 0.434  &      & 0.484  & 0.483  &      & 0.570  & 0.516  &      & 0.565  & 0.538  &      & 0.598  & 0.524  &      & 0.797  & 0.593  &      & 0.722  & 0.598  &      & 1.199  & 0.806  \\
     & \multicolumn{1}{r}{336} &      & \textbf{0.407} & \textbf{0.423} &      & 0.412  & 0.429  &      & 0.456  & 0.450  &      & 0.514  & 0.474  &      & 0.589  & 0.540  &      & 0.608  & 0.535  &      & 0.721  & 0.622  &      & 0.657  & 0.550  &      & 0.941  & 0.648  &      & 0.750  & 0.619  &      & 1.202  & 0.811  \\
     & \multicolumn{1}{r}{720} &      & \textbf{0.406} & \textbf{0.441} &      & 0.413  & 0.444  &      & 0.470  & 0.473  &      & 0.705  & 0.568  &      & 0.700  & 0.604  &      & 0.725  & 0.591  &      & 0.986  & 0.743  &      & 0.762  & 0.610  &      & 0.877  & 0.641  &      & 0.721  & 0.616  &      & 1.217  & 0.825  \\
     & \multicolumn{1}{r}{avg} &      & \textbf{0.390} & \textbf{0.414} &      & 0.400  & 0.424  &      & 0.434  & 0.439  &      & 0.510  & 0.469  &      & 0.556  & 0.522  &      & 0.590  & 0.525  &      & 0.691  & 0.600  &      & 0.633  & 0.542  &      & 0.869  & 0.628  &      & 0.702  & 0.596  &      & 1.199  & 0.809  \\
\midrule
\multirow{5}[2]{*}{\rotatebox{90}{$ETTh2$}} & \multicolumn{1}{r}{96} &      & \textbf{0.271} & 0.328  &      & 0.281  & 0.334  &      & 0.277  & 0.327  &      & 0.287  & \textbf{0.325} &      & 0.275  & 0.326  &      & 0.331  & 0.374  &      & 0.357  & 0.411  &      & 0.353  & 0.389  &      & 0.378  & 0.409  &      & 0.413  & 0.451  &      & 3.837  & 1.508  \\
     & \multicolumn{1}{r}{192} &      & \textbf{0.328} & \textbf{0.367} &      & 0.340  & 0.373  &      & 0.340  & 0.374  &      & 0.347  & \textbf{0.367} &      & 0.374  & 0.373  &      & 0.402  & 0.411  &      & 0.569  & 0.519  &      & 0.403  & 0.414  &      & 0.490  & 0.467  &      & 0.474  & 0.477  &      & 3.856  & 1.513  \\
     & \multicolumn{1}{r}{336} &      & \textbf{0.345} & \textbf{0.381} &      & 0.362  & 0.393  &      & 0.371  & 0.401  &      & 0.377  & 0.393  &      & 0.406  & 0.429  &      & 0.406  & 0.433  &      & 0.671  & 0.572  &      & 0.426  & 0.441  &      & 0.537  & 0.494  &      & 0.547  & 0.543  &      & 3.952  & 1.526  \\
     & \multicolumn{1}{r}{720} &      & \textbf{0.388} & 0.422  &      & 0.380  & 0.416  &      & 0.394  & 0.426  &      & 0.404  & \textbf{0.421} &      & 0.427  & 0.449  &      & 0.449  & 0.464  &      & 0.824  & 0.648  &      & 0.477  & 0.480  &      & 0.510  & 0.491  &      & 0.516  & 0.523  &      & 3.842  & 1.503  \\
     & \multicolumn{1}{r}{avg} &      & \textbf{0.333} & \textbf{0.375} &      & 0.341  & 0.379  &      & 0.346  & 0.382  &      & 0.354  & 0.377  &      & 0.370  & 0.394  &      & 0.397  & 0.421  &      & 0.605  & 0.538  &      & 0.415  & 0.431  &      & 0.479  & 0.465  &      & 0.488  & 0.499  &      & 3.872  & 1.513  \\
\midrule
\multirow{5}[2]{*}{\rotatebox{90}{$ETTm1$}} & \multicolumn{1}{r}{96} &      & \textbf{0.341} & \textbf{0.347} &      & 0.404  & 0.383  &      & 0.335  & 0.360  &      & 0.353  & 0.363  &      & 0.346  & 0.388  &      & 0.390  & 0.404  &      & 0.352  & 0.392  &      & 0.410  & 0.419  &      & 0.583  & 0.501  &      & 0.774  & 0.614  &      & 1.162  & 0.785  \\
     & \multicolumn{1}{r}{192} &      & \textbf{0.360} & \textbf{0.360} &      & 0.435  & 0.402  &      & 0.366  & 0.379  &      & 0.376  & 0.380  &      & 0.373  & 0.416  &      & 0.429  & 0.423  &      & 0.382  & 0.412  &      & 0.437  & 0.434  &      & 0.630  & 0.528  &      & 0.754  & 0.592  &      & 1.172  & 0.793  \\
     & \multicolumn{1}{r}{336} &      & \textbf{0.377} & \textbf{0.374} &      & 0.462  & 0.416  &      & 0.391  & 0.394  &      & 0.399  & 0.395  &      & 0.413  & 0.426  &      & 0.469  & 0.439  &      & 0.419  & 0.434  &      & 0.476  & 0.454  &      & 0.725  & 0.568  &      & 0.869  & 0.677  &      & 1.227  & 0.908  \\
     & \multicolumn{1}{r}{720} &      & \textbf{0.416} & \textbf{0.405} &      & 0.490  & 0.437  &      & 0.434  & 0.419  &      & 0.432  & 0.417  &      & 0.485  & 0.476  &      & 0.569  & 0.498  &      & 0.490  & 0.477  &      & 0.681  & 0.556  &      & 0.769  & 0.549  &      & 0.810  & 0.630  &      & 1.207  & 0.797  \\
     & \multicolumn{1}{r}{avg} &      & \textbf{0.374} & \textbf{0.372} &      & 0.448  & 0.410  &      & 0.382  & 0.388  &      & 0.390  & 0.389  &      & 0.404  & 0.427  &      & 0.464  & 0.441  &      & 0.411  & 0.429  &      & 0.501  & 0.466  &      & 0.677  & 0.537  &      & 0.802  & 0.628  &      & 1.192  & 0.821  \\
\midrule
\multirow{5}[2]{*}{\rotatebox{90}{$ETTm2$}} & \multicolumn{1}{r}{96} &      & 0.228  & 0.282  &      & 0.205  & 0.282  &      & 0.195  & 0.269  &      & \textbf{0.189} & \textbf{0.260} &      & 0.177  & 0.261  &      & 0.188  & 0.269  &      & 0.213  & 0.303  &      & 0.191  & 0.274  &      & 0.212  & 0.285  &      & 0.352  & 0.454  &      & 3.203  & 1.407  \\
     & \multicolumn{1}{r}{192} &      & 0.262  & 0.305  &      & 0.261  & 0.318  &      & 0.247  & 0.303  &      & \textbf{0.247} & \textbf{0.300} &      & 0.241  & 0.314  &      & 0.251  & 0.309  &      & 0.278  & 0.345  &      & 0.252  & 0.317  &      & 0.270  & 0.323  &      & 0.694  & 0.691  &      & 3.112  & 1.387  \\
     & \multicolumn{1}{r}{336} &      & 0.293  & \textbf{0.328} &      & 0.319  & 0.355  &      & \textbf{0.291} & 0.333  &      & 0.295  & 0.334  &      & 0.274  & 0.327  &      & 0.307  & 0.346  &      & 0.338  & 0.385  &      & 0.306  & 0.353  &      & 0.323  & 0.353  &      & 2.408  & 1.407  &      & 3.255  & 1.421  \\
     & \multicolumn{1}{r}{720} &      & \textbf{0.343} & \textbf{0.370} &      & 0.415  & 0.410  &      & 0.355  & 0.377  &      & 0.372  & 0.386  &      & 0.417  & 0.390  &      & 0.426  & 0.417  &      & 0.436  & 0.440  &      & 0.433  & 0.427  &      & 0.474  & 0.449  &      & 1.913  & 1.166  &      & 3.909  & 1.543  \\
     & \multicolumn{1}{r}{avg} &      & 0.282  & 0.321  &      & 0.300  & 0.341  &      & \textbf{0.272} & 0.321  &      & 0.276  & \textbf{0.320} &      & 0.277  & 0.323  &      & 0.293  & 0.335  &      & 0.316  & 0.368  &      & 0.296  & 0.343  &      & 0.320  & 0.353  &      & 1.342  & 0.930  &      & 3.370  & 1.440  \\
\midrule
\multirow{5}[2]{*}{\rotatebox{90}{$Electricity$}} & \multicolumn{1}{r}{96} &      & 0.177  & 0.266  &      & 0.205  & 0.299  &      & 0.158  & 0.248  &      & 0.152  & 0.242  &      & \textbf{0.139} & 0.241  &      & \textbf{0.139} & \textbf{0.237} &      & 0.150  & 0.253  &      & 0.140  & 0.238  &      & 0.299  & 0.373  &      & 0.261  & 0.348  &      & 1.259  & 0.919  \\
     & \multicolumn{1}{r}{192} &      & 0.188  & 0.277  &      & 0.220  & 0.310  &      & 0.174  & 0.263  &      & 0.171  & 0.259  &      & \textbf{0.151} & \textbf{0.248} &      & 0.156  & 0.252  &      & 0.164  & 0.264  &      & 0.160  & 0.255  &      & 0.305  & 0.379  &      & 0.338  & 0.406  &      & 1.160  & 0.873  \\
     & \multicolumn{1}{r}{336} &      & 0.207  & 0.296  &      & 0.236  & 0.323  &      & 0.191  & 0.278  &      & 0.192  & 0.278  &      & \textbf{0.169} & \textbf{0.270} &      & 0.175  & \textbf{0.270} &      & 0.181  & 0.282  &      & 0.180  & 0.276  &      & 0.319  & 0.391  &      & 0.410  & 0.474  &      & 1.157  & 0.872  \\
     & \multicolumn{1}{r}{720} &      & 0.256  & 0.337  &      & 0.270  & 0.347  &      & 0.229  & \textbf{0.307} &      & 0.236  & 0.313  &      & 0.240  & 0.322  &      & \textbf{0.233} & 0.317  &      & 0.223  & 0.321  &      & 0.241  & 0.323  &      & 0.369  & 0.426  &      & 0.715  & 0.685  &      & 1.203  & 0.898  \\
     & \multicolumn{1}{r}{avg} &      & 0.207  & 0.294  &      & 0.233  & 0.320  &      & 0.188  & 0.274  &      & 0.188  & 0.273  &      & \textbf{0.175} & 0.270  &      & 0.176  & \textbf{0.269} &      & 0.180  & 0.280  &      & 0.180  & 0.273  &      & 0.323  & 0.392  &      & 0.431  & 0.478  &      & 1.195  & 0.891  \\
\midrule
\multirow{5}[2]{*}{\rotatebox{90}{$Weather$}} & \multicolumn{1}{r}{96} &      & 0.220  & 0.257  &      & 0.173  & 0.212  &      & 0.167  & \textbf{0.203} &      & 0.177  & 0.208  &      & \textbf{0.161} & 0.210  &      & 0.163  & 0.215  &      & 0.171  & 0.224  &      & 0.165  & 0.215  &      & 0.184  & 0.230  &      & 0.221  & 0.297  &      & 0.374  & 0.401  \\
     & \multicolumn{1}{r}{192} &      & 0.244  & 0.275  &      & 0.216  & 0.250  &      & 0.209  & \textbf{0.241} &      & 0.219  & 0.249  &      & \textbf{0.204} & 0.248  &      & 0.210  & 0.254  &      & 0.215  & 0.263  &      & 0.210  & 0.257  &      & 0.245  & 0.283  &      & 0.270  & 0.322  &      & 0.552  & 0.478  \\
     & \multicolumn{1}{r}{336} &      & 0.280  & 0.299  &      & 0.260  & 0.282  &      & \textbf{0.256} & \textbf{0.276} &      & 0.277  & 0.292  &      & 0.261  & 0.302  &      & \textbf{0.256} & 0.292  &      & 0.258  & 0.299  &      & 0.259  & 0.297  &      & 0.305  & 0.321  &      & 0.320  & 0.351  &      & 0.724  & 0.541  \\
     & \multicolumn{1}{r}{720} &      & 0.330  & 0.337  &      & 0.320  & 0.322  &      & 0.321  & \textbf{0.323} &      & 0.365  & 0.350  &      & \textbf{0.309} & 0.332  &      & 0.321  & 0.339  &      & 0.320  & 0.346  &      & 0.332  & 0.346  &      & 0.381  & 0.371  &      & 0.390  & 0.396  &      & 0.739  & 0.558  \\
     & \multicolumn{1}{r}{avg} &      & 0.269  & 0.292  &      & 0.242  & 0.267  &      & 0.238  & \textbf{0.261} &      & 0.260  & 0.275  &      & \textbf{0.234} & 0.273  &      & 0.238  & 0.275  &      & 0.241  & 0.283  &      & 0.242  & 0.279  &      & 0.279  & 0.301  &      & 0.300  & 0.342  &      & 0.597  & 0.495  \\
\midrule
\rowc
\multicolumn{2}{c}{\textbf{Average}} &      & \textbf{0.309} & 0.345  &      & 0.327  & 0.357  &      & 0.310  & \textbf{0.344} &      & 0.329  & 0.350  &      & 0.336  & 0.368  &      & 0.360  & 0.378  &      & 0.407  & 0.416  &      & 0.378  & 0.389  &      & 0.491  & 0.446  &      & 0.678  & 0.579  &      & 1.904  & 0.995  \\
\rowc
\multicolumn{2}{c}{\textbf{$\bf 1^{st}$ count}} &      & \multicolumn{2}{c}{\textbf{32}} &      & \multicolumn{2}{c}{0} &      & \multicolumn{2}{c}{10} &      & \multicolumn{2}{c}{8} &      & \multicolumn{2}{c}{10} &      & \multicolumn{2}{c}{6} &      & \multicolumn{2}{c}{0} &      & \multicolumn{2}{c}{0} &      & \multicolumn{2}{c}{0} &      & \multicolumn{2}{c}{0} &      & \multicolumn{2}{c}{0} \\
\bottomrule
\end{tabular}%

}
  \label{tab:zero_shot_full}%
\end{table}%

%% file: table/tab_zero_shot_timesfm.tex
\begin{table}[t]
    \centering
    \begin{minipage}[t]{0.34\textwidth}

  \centering
  \caption{MAE results of TimesFM and LLMTime for zero-shot forecasting, on the last test window of the original test split.}
\vskip 0.15in
\resizebox{1.0\linewidth}{!}{
\begin{tabular}{ccccc}
    \toprule
    \multicolumn{2}{c}{\textbf{Method}} & \textbf{\ours} & \textbf{TimesFM} & \textbf{LLMTime} \\
    \midrule
    \multirow{2}[1]{*}{ETTh1} & 96   & \textbf{0.35} & 0.45  & 0.42  \\
         & 192  & \textbf{0.45} & 0.53  & 0.50  \\
    \midrule
    \multirow{2}[0]{*}{ETTh2} & 96   & \textbf{0.24} & 0.35  & 0.33  \\
         & 192  & \textbf{0.60} & 0.62  & 0.70  \\
    \midrule
    \multirow{2}[0]{*}{ETTm1} & 96   & \textbf{0.12} & 0.19  & 0.37  \\
         & 192  & \textbf{0.23} & 0.26  & 0.71  \\
    \midrule
    \multirow{2}[0]{*}{ETTm2} & 96   & \textbf{0.19} & 0.24  & 0.29  \\
         & 192  & \textbf{0.24} & 0.27  & 0.31  \\
    \midrule
    \rowc \multicolumn{2}{c}{\textbf{Average}} & \textbf{0.30} & 0.36  & 0.45  \\
    \bottomrule
    \end{tabular}%
}
  \label{tab:zero_shot_timesfm}
    \end{minipage}
    \hfill
    \begin{minipage}[t]{0.64\textwidth}
        \centering
  \caption{Comparison of traditional forecasting baselines in the zero-shot setting.}
\vskip 0.15in
\resizebox{1.0\linewidth}{!}{

    \begin{tabular}{ccccccccccccccccc}
    \toprule
    \multicolumn{2}{c}{\textbf{Method}} &      & \multicolumn{2}{c}{\textbf{\ours}} &      & \multicolumn{2}{c}{\textbf{ETS}} &      & \multicolumn{2}{c}{\textbf{ARIMA}} &      & \multicolumn{2}{c}{\textbf{Seasonal Naïve}} &      & \multicolumn{2}{c}{\textbf{Seasonal Avg}} \\
\cmidrule{4-5}\cmidrule{7-8}\cmidrule{10-11}\cmidrule{13-14}\cmidrule{16-17}    \multicolumn{2}{c}{\textbf{Metric}} &      & MSE  & MAE  &      & MSE  & MAE  &      & MSE  & MAE  &      & MSE  & MAE  &      & MSE  & MAE \\
    \midrule
    \multirow{5}[1]{*}{\rotatebox{90}{$ETTh1$}} & \multicolumn{1}{r}{96} &      & \textbf{0.353} & \textbf{0.383} &      & 1.289  & 0.710  &      & 0.900  & 0.719  &      & 0.512  & 0.433  &      & 0.589  & 0.585  \\
         & \multicolumn{1}{r}{192} &      & \textbf{0.392} & \textbf{0.410} &      & 1.319  & 0.730  &      & 0.906  & 0.724  &      & 0.581  & 0.469  &      & 0.598  & 0.590  \\
         & \multicolumn{1}{r}{336} &      & \textbf{0.407} & \textbf{0.423} &      & 1.324  & 0.742  &      & 0.908  & 0.731  &      & 0.650  & 0.501  &      & 0.610  & 0.597  \\
         & \multicolumn{1}{r}{720} &      & \textbf{0.406} & \textbf{0.441} &      & 1.329  & 0.751  &      & 0.932  & 0.753  &      & 0.655  & 0.514  &      & 0.656  & 0.624  \\
         & \multicolumn{1}{r}{avg} &      & \textbf{0.390} & \textbf{0.414} &      & 1.315  & 0.733  &      & 0.912  & 0.732  &      & 0.600  & 0.479  &      & 0.613  & 0.599  \\
    \midrule
    \multirow{5}[1]{*}{\rotatebox{90}{$ETTh2$}} & \multicolumn{1}{r}{96} &      & \textbf{0.271} & \textbf{0.328} &      & 0.399  & 0.408  &      & 0.488  & 0.508  &      & 0.391  & 0.380  &      & 0.457  & 0.494  \\
         & \multicolumn{1}{r}{192} &      & \textbf{0.328} & \textbf{0.367} &      & 0.500  & 0.459  &      & 0.497  & 0.514  &      & 0.482  & 0.429  &      & 0.466  & 0.500  \\
         & \multicolumn{1}{r}{336} &      & \textbf{0.345} & \textbf{0.381} &      & 0.562  & 0.498  &      & 0.507  & 0.522  &      & 0.532  & 0.466  &      & 0.476  & 0.509  \\
         & \multicolumn{1}{r}{720} &      & \textbf{0.388} & \textbf{0.422} &      & 0.558  & 0.506  &      & 0.572  & 0.557  &      & 0.525  & 0.474  &      & 0.542  & 0.548  \\
         & \multicolumn{1}{r}{avg} &      & \textbf{0.333} & \textbf{0.375} &      & 0.505  & 0.468  &      & 0.516  & 0.525  &      & 0.483  & 0.437  &      & 0.485  & 0.513  \\
    \midrule
    \multirow{5}[1]{*}{\rotatebox{90}{$ETTm1$}} & \multicolumn{1}{r}{96} &      & \textbf{0.341} & \textbf{0.347} &      & 1.204  & 0.659  &      & 0.702  & 0.568  &      & 0.423  & 0.387  &      & 0.369  & 0.399  \\
         & \multicolumn{1}{r}{192} &      & \textbf{0.360} & \textbf{0.360} &      & 1.251  & 0.685  &      & 0.704  & 0.570  &      & 0.463  & 0.406  &      & 0.374  & 0.402  \\
         & \multicolumn{1}{r}{336} &      & \textbf{0.377} & \textbf{0.374} &      & 1.276  & 0.702  &      & 0.709  & 0.574  &      & 0.496  & 0.426  &      & 0.382  & 0.407  \\
         & \multicolumn{1}{r}{720} &      & 0.416  & \textbf{0.405} &      & 1.311  & 0.724  &      & 0.713  & 0.580  &      & 0.574  & 0.464  &      & \textbf{0.394} & 0.416  \\
         & \multicolumn{1}{r}{avg} &      & \textbf{0.374} & \textbf{0.372} &      & 1.261  & 0.693  &      & 0.707  & 0.573  &      & 0.489  & 0.421  &      & 0.380  & 0.406  \\
    \midrule
    \multirow{5}[1]{*}{\rotatebox{90}{$ETTm2$}} & \multicolumn{1}{r}{96} &      & \textbf{0.228} & \textbf{0.282} &      & 0.257  & 0.324  &      & 0.397  & 0.434  &      & 0.263  & 0.301  &      & 0.365  & 0.411  \\
         & \multicolumn{1}{r}{192} &      & \textbf{0.262} & \textbf{0.305} &      & 0.331  & 0.366  &      & 0.402  & 0.436  &      & 0.321  & 0.337  &      & 0.369  & 0.414  \\
         & \multicolumn{1}{r}{336} &      & \textbf{0.293} & \textbf{0.328} &      & 0.402  & 0.406  &      & 0.407  & 0.439  &      & 0.376  & 0.370  &      & 0.375  & 0.418  \\
         & \multicolumn{1}{r}{720} &      & \textbf{0.343} & \textbf{0.370} &      & 0.512  & 0.462  &      & 0.413  & 0.443  &      & 0.471  & 0.422  &      & 0.380  & 0.423  \\
         & \multicolumn{1}{r}{avg} &      & \textbf{0.282} & \textbf{0.321} &      & 0.376  & 0.390  &      & 0.405  & 0.438  &      & 0.358  & 0.357  &      & 0.372  & 0.417  \\
    \midrule
    \rowc\multicolumn{2}{c}{\textbf{Average}} &      & \textbf{0.344} & \textbf{0.370} &      & 0.864  & 0.571  &      & 0.635  & 0.567  &      & 0.482  & 0.424  &      & 0.463  & 0.484  \\
    \rowc\multicolumn{2}{c}{\boldmath{}\textbf{$\bf 1^{st}$ count}\unboldmath{}} &      & \multicolumn{2}{c}{\textbf{41}} &      & \multicolumn{2}{c}{0} &      & \multicolumn{2}{c}{0} &      & \multicolumn{2}{c}{0} &      & \multicolumn{2}{c}{1} \\
    \bottomrule
    \end{tabular}%

}
  \label{tab:zero_shot_season}%
    \end{minipage}
\end{table}

%% file: table/tab_zero_shot_seasonal.tex
\begin{table}[t]
  \centering
  \caption{Comparison of traditional zero-shot forecasting baselines.}
\vskip 0.15in
\resizebox{0.7\linewidth}{!}{

    \begin{tabular}{ccccccccccccccccc}
    \toprule
    \multicolumn{2}{c}{\textbf{Method}} &      & \multicolumn{2}{c}{\textbf{\ours}} &      & \multicolumn{2}{c}{\textbf{ETS}} &      & \multicolumn{2}{c}{\textbf{ARIMA}} &      & \multicolumn{2}{c}{\textbf{Seasonal Naïve}} &      & \multicolumn{2}{c}{\textbf{Seasonal Avg}} \\
\cmidrule{4-5}\cmidrule{7-8}\cmidrule{10-11}\cmidrule{13-14}\cmidrule{16-17}    \multicolumn{2}{c}{\textbf{Metric}} &      & MSE  & MAE  &      & MSE  & MAE  &      & MSE  & MAE  &      & MSE  & MAE  &      & MSE  & MAE \\
    \midrule
    \multirow{5}[1]{*}{\rotatebox{90}{$ETTh1$}} & \multicolumn{1}{r}{96} &      & \textbf{0.353} & \textbf{0.383} &      & 1.289  & 0.710  &      & 0.900  & 0.719  &      & 0.512  & 0.433  &      & 0.589  & 0.585  \\
         & \multicolumn{1}{r}{192} &      & \textbf{0.392} & \textbf{0.410} &      & 1.319  & 0.730  &      & 0.906  & 0.724  &      & 0.581  & 0.469  &      & 0.598  & 0.590  \\
         & \multicolumn{1}{r}{336} &      & \textbf{0.407} & \textbf{0.423} &      & 1.324  & 0.742  &      & 0.908  & 0.731  &      & 0.650  & 0.501  &      & 0.610  & 0.597  \\
         & \multicolumn{1}{r}{720} &      & \textbf{0.406} & \textbf{0.441} &      & 1.329  & 0.751  &      & 0.932  & 0.753  &      & 0.655  & 0.514  &      & 0.656  & 0.624  \\
         & \multicolumn{1}{r}{avg} &      & \textbf{0.390} & \textbf{0.414} &      & 1.315  & 0.733  &      & 0.912  & 0.732  &      & 0.600  & 0.479  &      & 0.613  & 0.599  \\
    \midrule
    \multirow{5}[1]{*}{\rotatebox{90}{$ETTh2$}} & \multicolumn{1}{r}{96} &      & \textbf{0.271} & \textbf{0.328} &      & 0.399  & 0.408  &      & 0.488  & 0.508  &      & 0.391  & 0.380  &      & 0.457  & 0.494  \\
         & \multicolumn{1}{r}{192} &      & \textbf{0.328} & \textbf{0.367} &      & 0.500  & 0.459  &      & 0.497  & 0.514  &      & 0.482  & 0.429  &      & 0.466  & 0.500  \\
         & \multicolumn{1}{r}{336} &      & \textbf{0.345} & \textbf{0.381} &      & 0.562  & 0.498  &      & 0.507  & 0.522  &      & 0.532  & 0.466  &      & 0.476  & 0.509  \\
         & \multicolumn{1}{r}{720} &      & \textbf{0.388} & \textbf{0.422} &      & 0.558  & 0.506  &      & 0.572  & 0.557  &      & 0.525  & 0.474  &      & 0.542  & 0.548  \\
         & \multicolumn{1}{r}{avg} &      & \textbf{0.333} & \textbf{0.375} &      & 0.505  & 0.468  &      & 0.516  & 0.525  &      & 0.483  & 0.437  &      & 0.485  & 0.513  \\
    \midrule
    \multirow{5}[1]{*}{\rotatebox{90}{$ETTm1$}} & \multicolumn{1}{r}{96} &      & \textbf{0.341} & \textbf{0.347} &      & 1.204  & 0.659  &      & 0.702  & 0.568  &      & 0.423  & 0.387  &      & 0.369  & 0.399  \\
         & \multicolumn{1}{r}{192} &      & \textbf{0.360} & \textbf{0.360} &      & 1.251  & 0.685  &      & 0.704  & 0.570  &      & 0.463  & 0.406  &      & 0.374  & 0.402  \\
         & \multicolumn{1}{r}{336} &      & \textbf{0.377} & \textbf{0.374} &      & 1.276  & 0.702  &      & 0.709  & 0.574  &      & 0.496  & 0.426  &      & 0.382  & 0.407  \\
         & \multicolumn{1}{r}{720} &      & 0.416  & \textbf{0.405} &      & 1.311  & 0.724  &      & 0.713  & 0.580  &      & 0.574  & 0.464  &      & \textbf{0.394} & 0.416  \\
         & \multicolumn{1}{r}{avg} &      & \textbf{0.374} & \textbf{0.372} &      & 1.261  & 0.693  &      & 0.707  & 0.573  &      & 0.489  & 0.421  &      & 0.380  & 0.406  \\
    \midrule
    \multirow{5}[1]{*}{\rotatebox{90}{$ETTm2$}} & \multicolumn{1}{r}{96} &      & \textbf{0.228} & \textbf{0.282} &      & 0.257  & 0.324  &      & 0.397  & 0.434  &      & 0.263  & 0.301  &      & 0.365  & 0.411  \\
         & \multicolumn{1}{r}{192} &      & \textbf{0.262} & \textbf{0.305} &      & 0.331  & 0.366  &      & 0.402  & 0.436  &      & 0.321  & 0.337  &      & 0.369  & 0.414  \\
         & \multicolumn{1}{r}{336} &      & \textbf{0.293} & \textbf{0.328} &      & 0.402  & 0.406  &      & 0.407  & 0.439  &      & 0.376  & 0.370  &      & 0.375  & 0.418  \\
         & \multicolumn{1}{r}{720} &      & \textbf{0.343} & \textbf{0.370} &      & 0.512  & 0.462  &      & 0.413  & 0.443  &      & 0.471  & 0.422  &      & 0.380  & 0.423  \\
         & \multicolumn{1}{r}{avg} &      & \textbf{0.282} & \textbf{0.321} &      & 0.376  & 0.390  &      & 0.405  & 0.438  &      & 0.358  & 0.357  &      & 0.372  & 0.417  \\
    \midrule
    \rowc\multicolumn{2}{c}{\textbf{Average}} &      & \textbf{0.344} & \textbf{0.370} &      & 0.864  & 0.571  &      & 0.635  & 0.567  &      & 0.482  & 0.424  &      & 0.463  & 0.484  \\
    \rowc\multicolumn{2}{c}{\boldmath{}\textbf{$\bf 1^{st}$ count}\unboldmath{}} &      & \multicolumn{2}{c}{\textbf{41}} &      & \multicolumn{2}{c}{0} &      & \multicolumn{2}{c}{0} &      & \multicolumn{2}{c}{0} &      & \multicolumn{2}{c}{1} \\
    \bottomrule
    \end{tabular}%

}
  \label{tab:zero_shot_season}%
\end{table}%

%% file: table/tab_zero_shot_monash.tex
\begin{table}[t]
  \centering
  \caption{Full results of \cref{fig:zs_monash_aggregated}: Forecasting results (MAE) on the Monash TSF benchmark. We reported the reproduction results of LLMTime based on the GPT3.5 API from \citet{Moirai}.}
\vskip 0.15in
    \resizebox{\linewidth}{!}{

    \begin{tabular}{lcccccccccccccccc}
    \toprule
         & \textbf{\ours} & \textbf{LLMTime} & \textbf{\moirai{Small}} & \textbf{Naive} & \textbf{SES} & \textbf{Theta} & \textbf{TBATS} & \textbf{ETS} & \textbf{(DHR-)ARIMA} & \textbf{PR} & \textbf{CatBoost} & \textbf{FFNN} & \textbf{DeepAR} & \textbf{N-BEATS} & \textbf{WaveNet} & \textbf{Transformer} \\
    \midrule
    M1 Monthly & 1987.69 & 2562.84 & 2082.26 & 2707.75 & 2259.04 & 2166.18 & 2237.5 & 1905.28 & 2080.13 & 2088.25 & 2052.32 & 2162.58 & 1860.81 & \textbf{1820.37} & 2184.42 & 2723.88 \\
    M3 Monthly & 737.93 & 877.97 & 713.41 & 837.14 & 743.41 & \textbf{623.71} & 630.59 & 626.46 & 654.8 & 692.97 & 732  & 692.48 & 728.81 & 648.6 & 699.3 & 798.38 \\
    M3 Other & 315.85 & 300.3 & 263.54 & 278.43 & 277.83 & 215.35 & \textbf{189.42} & 194.98 & 193.02 & 234.43 & 318.13 & 240.17 & 247.56 & 221.85 & 245.29 & 239.24 \\
    M4 Monthly & 666.54 & 728.27 & 597.6 & 671.27 & 625.24 & \textbf{563.58} & 589.52 & 582.6 & 575.36 & 596.19 & 611.69 & 612.52 & 615.22 & 578.48 & 655.51 & 780.47 \\
    M4 Weekly & 404.23 & 518.44 & 339.76 & 347.99 & 336.82 & 333.32 & 296.15 & 335.66 & 321.61 & 293.21 & 364.65 & 338.37 & 351.78 & \textbf{277.73} & 359.46 & 378.89 \\
    M4 Daily & 215.63 & 266.52 & 189.1 & 180.83 & 178.27 & 178.86 & \textbf{176.6} & 193.26 & 179.67 & 181.92 & 231.36 & 177.91 & 299.79 & 190.44 & 189.47 & 201.08 \\
    M4 Hourly & 288.37 & 576.06 & 268.04 & 1218.06 & 1218.06 & 1220.97 & 386.27 & 3358.1 & 1310.85 & \textbf{257.39} & 285.35 & 385.49 & 886.02 & 425.75 & 393.63 & 320.54 \\
    Tourism Quarterly & 12931.88 & 16918.86 & 18352.44 & 15845.1 & 15014.19 & \textbf{7656.49} & 9972.42 & 8925.52 & 10475.47 & 9092.58 & 10267.97 & 8981.04 & 9511.37 & 8640.56 & 9137.12 & 9521.67 \\
    Tourism Monthly & 2560.19 & 5608.61 & 3569.85 & 5636.83 & 5302.1 & 2069.96 & 2940.08 & 2004.51 & 2536.77 & 2187.28 & 2537.04 & 2022.21 & \textbf{1871.69} & 2003.02 & 2095.13 & 2146.98 \\
    CIF 2016 & 570907.24  & 599313.8 & 655888.58 & 578596.5 & 581875.97 & 714818.6 & 855578.4 & 642421.4 & \textbf{469059} & 563205.57 & 603551.3 & 1495923 & 3200418 & 679034.8 & 5998225 & 4057973 \\
    Aus. Elec. Demand & 237.44 & 760.81 & 266.57 & 659.6 & 659.6 & 665.04 & 370.74 & 1282.99 & 1045.92 & 247.18 & 241.77 & 258.76 & 302.41 & \textbf{213.83} & 227.5 & 231.45 \\
    Bitcoin & 2.33E+18 & 1.74E+18 & 1.76E+18 & 7.78E+17 & 5.33E+18 & 5.33E+18 & 9.9E+17 & 1.1E+18 & 3.62E+18 & \textbf{6.66E+17} & 1.93E+18 & 1.45E+18 & 1.95E+18 & 1.06E+18 & 2.46E+18 & 2.61E+18 \\
    Pedestrian Counts & 52.01 & 97.77 & 54.88 & 170.88 & 170.87 & 170.94 & 222.38 & 216.5 & 635.16 & 44.18 & \textbf{43.41} & 46.41 & 44.78 & 66.84 & 46.46 & 47.29 \\
    Vehicle Trips & 22.08 & 31.48 & 24.46 & 31.42 & 29.98 & 30.76 & \textbf{21.21} & 30.95 & 30.07 & 27.24 & 22.61 & 22.93 & 22   & 28.16 & 24.15 & 28.01 \\
    KDD cup & 38.16 & 42.72 & 39.81 & 42.13 & 42.04 & 42.06 & 39.2 & 44.88 & 52.2 & 36.85 & \textbf{34.82} & 37.16 & 48.98 & 49.1 & 37.08 & 44.46 \\
    Weather & 2.06 & 2.17 & \textbf{1.96} & 2.36 & 2.24 & 2.51 & 2.3  & 2.35 & 2.45 & 8.17 & 2.51 & 2.09 & 2.02 & 2.34 & 2.29 & 2.03 \\
    NN5 Daily & \textbf{3.51} & 7.1  & 5.37 & 8.26 & 6.63 & 3.8  & 3.7  & 3.72 & 4.41 & 5.47 & 4.22 & 4.06 & 3.94 & 4.92 & 3.97 & 4.16 \\
    NN5 Weekly & 14.67 & 15.76 & 15.07 & 16.71 & 15.66 & 15.3 & 14.98 & 15.7 & 15.38 & 14.94 & 15.29 & 15.02 & 14.69 & \textbf{14.19} & 19.34 & 20.34 \\
    Carparts & 0.58 & 0.44 & 0.53 & 0.65 & 0.55 & 0.53 & 0.58 & 0.56 & 0.56 & 0.41 & 0.53 & \textbf{0.39} & \textbf{0.39} & 0.98 & 0.4  & \textbf{0.39} \\
    FRED-MD & \textbf{1893.67} & 2804.64 & 2568.48 & 2825.67 & 2798.22 & 3492.84 & 1989.97 & 2041.42 & 2957.11 & 8921.94 & 2475.68 & 2339.57 & 4264.36 & 2557.8 & 2508.4 & 4666.04 \\
    Traffic Hourly & \textbf{0.01} & 0.03 & 0.02 & 0.03 & 0.03 & 0.03 & 0.04 & 0.03 & 0.04 & 0.02 & 0.02 & \textbf{0.01} & \textbf{0.01} & 0.02 & 0.02 & \textbf{0.01} \\
    Traffic Weekly & 1.14 & 1.15 & 1.17 & 1.19 & 1.12 & 1.13 & 1.17 & 1.14 & 1.22 & 1.13 & 1.17 & 1.15 & 1.18 & \textbf{1.11} & 1.2  & 1.42 \\
    Rideshare & 5.92 & 6.28 & \textbf{1.35} & 6.29 & 6.29 & 7.62 & 6.45 & 6.29 & 3.37 & 6.3  & 6.07 & 6.59 & 6.28 & 5.55 & 2.75 & 6.29 \\
    Hospital & 19.36 & 25.68 & 23   & 24.07 & 21.76 & 18.54 & \textbf{17.43} & 17.97 & 19.6 & 19.24 & 19.17 & 22.86 & 18.25 & 20.18 & 19.35 & 36.19 \\
    COVID Deaths & 137.51 & 653.31 & 124.32 & 353.71 & 353.71 & 321.32 & 96.29 & \textbf{85.59} & 85.77 & 347.98 & 475.15 & 144.14 & 201.98 & 158.81 & 1049.48 & 408.66 \\
    Temperature Rain & 6.37 & 6.37 & 5.3  & 9.39 & 8.18 & 8.22 & 7.14 & 8.21 & 7.19 & 6.13 & 6.76 & 5.56 & 5.37 & 7.28 & 5.81 & \textbf{5.24} \\
    Sunspot & 2.81 & 5.07 & \textbf{0.11} & 3.93 & 4.93 & 4.93 & 2.57 & 4.93 & 2.57 & 3.83 & 2.27 & 7.97 & 0.77 & 14.47 & 0.17 & 0.13 \\
    Saugeen River Flow & 30.22 & 34.84 & 24.07 & 21.5 & 21.5 & 21.49 & 22.26 & 30.69 & 22.38 & 25.24 & \textbf{21.28} & 22.98 & 23.51 & 27.92 & 22.17 & 28.06 \\
    US Births & 519.94 & 1374.99 & 872.51 & 1152.67 & 1192.2 & 586.93 & \textbf{399} & 419.73 & 526.33 & 574.93 & 441.7 & 557.87 & 424.93 & 422  & 504.4 & 452.87 \\
    \midrule
    \rowc \textbf{Normalized MAE} & 0.729  & 1.041  & 0.657  & 1.000  & 1.028  & 0.927  & 0.758  & 0.872  & 0.898  & 0.785  & 0.760  & 0.741  & 0.759  & 0.783  & 0.749  & 0.770  \\
    \rowc \textbf{Rank} & 2    & 16   & 1    & 14   & 15   & 13   & 5    & 11   & 12   & 10   & 7    & 3    & 6    & 9    & 4    & 8 \\
    \bottomrule
    \end{tabular}%

}
  \label{tab:zero_shot_monash}%
\end{table}%

%% file: table/tab_lama.tex
\begin{table}[htbp]
  \centering
  \caption{Comparison of LaMa as the backbone. Results are averaged on four prediction lengths.}
\vskip 0.15in
  \small
    \begin{tabular}{cccccccccrcc}
    \toprule
         & \multicolumn{2}{c}{\textbf{\mae}} &      & \multicolumn{2}{c}{\textbf{LaMa}} &      & \multicolumn{2}{c}{\textbf{\moirai{Small}}} &      & \multicolumn{2}{c}{\textbf{\moirai{Large}}} \\
\cmidrule{2-3}\cmidrule{5-6}\cmidrule{8-9}\cmidrule{11-12}         & MSE  & MAE  &      & MSE  & MAE  &      & MSE  & MAE  &      & MSE  & MAE \\
\cmidrule{1-3}\cmidrule{5-6}\cmidrule{8-9}\cmidrule{11-12}    ETTh1 & 0.390  & 0.414  &      & 0.425  & 0.433  &      & 0.400  & 0.424  &      & 0.510  & 0.469  \\
    ETTh2 & 0.333  & 0.375  &      & 0.376  & 0.408  &      & 0.341  & 0.379  &      & 0.354  & 0.377  \\
    ETTm1 & 0.374  & 0.372  &      & 0.400  & 0.391  &      & 0.448  & 0.410  &      & 0.390  & 0.389  \\
    ETTm2 & 0.282  & 0.321  &      & 0.294  & 0.337  &      & 0.300  & 0.341  &      & 0.276  & 0.320  \\
    \rowc \textbf{Average} & 0.344  & 0.370  &      & 0.374  & 0.392  &      & 0.372  & 0.388  &      & 0.382  & 0.388  \\
    \bottomrule
    \end{tabular}%
  \label{tab:lama}%
\end{table}%

%% file: table/tab_zero_shot_size.tex
\begin{table}[t]
  \centering
  \caption{Full results of \cref{tab:zs_short_size}: zero-shot forecasting results of different \mae variants. \textbf{Bold}: best results among three variants. We also include the results from \moirai{} for reference.}
\vskip 0.15in
    \resizebox{0.8\linewidth}{!}{

    \begin{tabular}{cccccccccccccccccrcc}
    \toprule
    \multicolumn{2}{c}{\multirow{2}[2]{*}{\textbf{Method}}} &      & \multicolumn{2}{c}{\mae (Base)} &      & \multicolumn{2}{c}{\mae (Large)} &      & \multicolumn{2}{c}{\mae (Huge)} &      & \multicolumn{2}{c}{\moirai{} (Small)} &      & \multicolumn{2}{c}{\moirai{} (Base)} &      & \multicolumn{2}{c}{\moirai{} (Huge)} \\
    \multicolumn{2}{c}{} &      & \multicolumn{2}{c}{112M} &      & \multicolumn{2}{c}{330M} &      & \multicolumn{2}{c}{657M} &      & \multicolumn{2}{c}{14M} &      & \multicolumn{2}{c}{91M} &      & \multicolumn{2}{c}{311M} \\
\cmidrule{4-5}\cmidrule{7-8}\cmidrule{10-11}\cmidrule{13-14}\cmidrule{16-17}\cmidrule{19-20}    \multicolumn{2}{c}{\textbf{Metric}} &      & MSE  & MAE  &      & MSE  & MAE  &      & MSE  & MAE  &      & MSE  & MAE  &      & MSE  & MAE  &      & MSE  & MAE \\
    \midrule
    \multirow{5}[1]{*}{\rotatebox{90}{$ETTh1$}} & \multicolumn{1}{r}{96} &      & 0.353  & 0.383  &      & \textbf{0.346} & \textbf{0.382} &      & 0.362  & 0.384  &      & \textit{0.375} & \textit{0.402} &      & \textit{0.384} & \textit{0.402} &      & \textit{0.380} & \textit{0.398} \\
         & \multicolumn{1}{r}{192} &      & 0.392  & 0.410  &      & \textbf{0.379} & \textbf{0.406} &      & 0.407  & 0.414  &      & \textit{0.399} & \textit{0.419} &      & \textit{0.425} & \textit{0.429} &      & \textit{0.440} & \textit{0.434} \\
         & \multicolumn{1}{r}{336} &      & 0.407  & 0.423  &      & \textbf{0.391} & \textbf{0.416} &      & 0.399  & 0.419  &      & \textit{0.412} & \textit{0.429} &      & \textit{0.456} & \textit{0.450} &      & \textit{0.514} & \textit{0.474} \\
         & \multicolumn{1}{r}{720} &      & 0.406  & 0.441  &      & 0.397  & \textbf{0.433} &      & \textbf{0.395} & \textbf{0.433} &      & \textit{0.413} & \textit{0.444} &      & \textit{0.470} & \textit{0.473} &      & \textit{0.705} & \textit{0.568} \\
         & \multicolumn{1}{r}{avg} &      & 0.390  & 0.414  &      & \textbf{0.378} & \textbf{0.409} &      & 0.391  & 0.412  &      & \textit{0.400} & \textit{0.424} &      & \textit{0.434} & \textit{0.439} &      & \textit{0.510} & \textit{0.469} \\
    \midrule
    \multirow{5}[1]{*}{\rotatebox{90}{$ETTh2$}} & \multicolumn{1}{r}{96} &      & \textbf{0.271} & \textbf{0.328} &      & 0.286  & 0.334  &      & 0.285  & 0.333  &      & \textit{0.281} & \textit{0.334} &      & \textit{0.277} & \textit{0.327} &      & \textit{0.287} & \textit{0.325} \\
         & \multicolumn{1}{r}{192} &      & \textbf{0.328} & \textbf{0.367} &      & 0.346  & 0.375  &      & 0.337  & 0.369  &      & \textit{0.340} & \textit{0.373} &      & \textit{0.340} & \textit{0.374} &      & \textit{0.347} & \textit{0.367} \\
         & \multicolumn{1}{r}{336} &      & \textbf{0.345} & \textbf{0.381} &      & 0.356  & 0.387  &      & 0.357  & 0.388  &      & \textit{0.362} & \textit{0.393} &      & \textit{0.371} & \textit{0.401} &      & \textit{0.377} & \textit{0.393} \\
         & \multicolumn{1}{r}{720} &      & 0.388  & 0.422  &      & \textbf{0.371} & \textbf{0.409} &      & 0.379  & 0.412  &      & \textit{0.380} & \textit{0.416} &      & \textit{0.394} & \textit{0.426} &      & \textit{0.404} & \textit{0.421} \\
         & \multicolumn{1}{r}{avg} &      & \textbf{0.333} & \textbf{0.375} &      & 0.340  & 0.377  &      & 0.339  & \textbf{0.375} &      & \textit{0.341} & \textit{0.379} &      & \textit{0.346} & \textit{0.382} &      & \textit{0.354} & \textit{0.377} \\
    \midrule
    \multirow{5}[2]{*}{\rotatebox{90}{$ETTm1$}} & \multicolumn{1}{r}{96} &      & \textbf{0.341} & \textbf{0.347} &      & 0.344  & 0.349  &      & 0.352  & 0.351  &      & \textit{0.404} & \textit{0.383} &      & \textit{0.335} & \textit{0.360} &      & \textit{0.353} & \textit{0.363} \\
         & \multicolumn{1}{r}{192} &      & \textbf{0.360} & \textbf{0.360} &      & 0.365  & 0.363  &      & \textbf{0.360} & 0.367  &      & \textit{0.435} & \textit{0.402} &      & \textit{0.366} & \textit{0.379} &      & \textit{0.376} & \textit{0.380} \\
         & \multicolumn{1}{r}{336} &      & \textbf{0.377} & \textbf{0.374} &      & 0.381  & 0.376  &      & 0.381  & 0.383  &      & \textit{0.462} & \textit{0.416} &      & \textit{0.391} & \textit{0.394} &      & \textit{0.399} & \textit{0.395} \\
         & \multicolumn{1}{r}{720} &      & \textbf{0.416} & \textbf{0.405} &      & 0.429  & 0.411  &      & 0.440  & 0.412  &      & \textit{0.490} & \textit{0.437} &      & \textit{0.434} & \textit{0.419} &      & \textit{0.432} & \textit{0.417} \\
         & \multicolumn{1}{r}{avg} &      & \textbf{0.374} & \textbf{0.372} &      & 0.379  & 0.375  &      & 0.383  & 0.378  &      & \textit{0.448} & \textit{0.410} &      & \textit{0.382} & \textit{0.388} &      & \textit{0.390} & \textit{0.389} \\
    \midrule
    \multirow{5}[2]{*}{\rotatebox{90}{$ETTm2$}} & \multicolumn{1}{r}{96} &      & 0.228  & \textbf{0.282} &      & \textbf{0.225} & \textbf{0.282} &      & 0.229  & \textbf{0.282} &      & \textit{0.205} & \textit{0.282} &      & \textit{0.195} & \textit{0.269} &      & \textit{0.189} & \textit{0.260} \\
         & \multicolumn{1}{r}{192} &      & \textbf{0.262} & \textbf{0.305} &      & \textbf{0.262} & \textbf{0.305} &      & 0.265  & 0.306  &      & \textit{0.261} & \textit{0.318} &      & \textit{0.247} & \textit{0.303} &      & \textit{0.247} & \textit{0.300} \\
         & \multicolumn{1}{r}{336} &      & 0.293  & 0.328  &      & 0.299  & 0.331  &      & \textbf{0.286} & \textbf{0.324} &      & \textit{0.319} & \textit{0.355} &      & \textit{0.291} & \textit{0.333} &      & \textit{0.295} & \textit{0.334} \\
         & \multicolumn{1}{r}{720} &      & \textbf{0.343} & \textbf{0.370} &      & 0.358  & 0.377  &      & 0.355  & 0.374  &      & \textit{0.415} & \textit{0.410} &      & \textit{0.355} & \textit{0.377} &      & \textit{0.372} & \textit{0.386} \\
         & \multicolumn{1}{r}{avg} &      & \textbf{0.282} & \textbf{0.321} &      & 0.286  & 0.324  &      & 0.284  & 0.322  &      & \textit{0.300} & \textit{0.341} &      & \textit{0.272} & \textit{0.321} &      & \textit{0.276} & \textit{0.320} \\
    \midrule
    \multirow{5}[2]{*}{\rotatebox{90}{$Electricity$}} & \multicolumn{1}{r}{96} &      & 0.177  & 0.266  &      & 0.177  & 0.268  &      & \textbf{0.170} & \textbf{0.259} &      & \textit{0.205} & \textit{0.299} &      & \textit{0.158} & \textit{0.248} &      & \textit{0.152} & \textit{0.242} \\
         & \multicolumn{1}{r}{192} &      & 0.188  & 0.277  &      & 0.192  & 0.283  &      & \textbf{0.182} & \textbf{0.273} &      & \textit{0.220} & \textit{0.310} &      & \textit{0.174} & \textit{0.263} &      & \textit{0.171} & \textit{0.259} \\
         & \multicolumn{1}{r}{336} &      & \textbf{0.207} & 0.296  &      & 0.213  & 0.303  &      & \textbf{0.207} & \textbf{0.295} &      & \textit{0.236} & \textit{0.323} &      & \textit{0.191} & \textit{0.278} &      & \textit{0.192} & \textit{0.278} \\
         & \multicolumn{1}{r}{720} &      & 0.256  & 0.337  &      & 0.256  & 0.337  &      & \textbf{0.250} & \textbf{0.333} &      & \textit{0.270} & \textit{0.347} &      & \textit{0.229} & \textit{0.307} &      & \textit{0.236} & \textit{0.313} \\
         & \multicolumn{1}{r}{avg} &      & 0.207  & 0.294  &      & 0.209  & 0.298  &      & \textbf{0.202} & \textbf{0.290} &      & \textit{0.233} & \textit{0.320} &      & \textit{0.188} & \textit{0.274} &      & \textit{0.188} & \textit{0.273} \\
    \midrule
    \multirow{5}[2]{*}{\rotatebox{90}{$Weather$}} & \multicolumn{1}{r}{96} &      & \textbf{0.220} & \textbf{0.257} &      & 0.222  & \textbf{0.257} &      & 0.235  & 0.265  &      & \textit{0.173} & \textit{0.212} &      & \textit{0.167} & \textit{0.203} &      & \textit{0.177} & \textit{0.208} \\
         & \multicolumn{1}{r}{192} &      & \textbf{0.244} & \textbf{0.275} &      & 0.246  & \textbf{0.275} &      & 0.276  & 0.288  &      & \textit{0.216} & \textit{0.250} &      & \textit{0.209} & \textit{0.241} &      & \textit{0.219} & \textit{0.249} \\
         & \multicolumn{1}{r}{336} &      & \textbf{0.280} & \textbf{0.299} &      & 0.283  & 0.301  &      & 0.304  & 0.309  &      & \textit{0.260} & \textit{0.282} &      & \textit{0.256} & \textit{0.276} &      & \textit{0.277} & \textit{0.292} \\
         & \multicolumn{1}{r}{720} &      & \textbf{0.330} & \textbf{0.337} &      & 0.338  & 0.343  &      & 0.351  & 0.350  &      & \textit{0.320} & \textit{0.322} &      & \textit{0.321} & \textit{0.323} &      & \textit{0.365} & \textit{0.350} \\
         & \multicolumn{1}{r}{avg} &      & \textbf{0.269} & \textbf{0.292} &      & 0.272  & 0.294  &      & 0.292  & 0.303  &      & \textit{0.242} & \textit{0.267} &      & \textit{0.238} & \textit{0.261} &      & \textit{0.260} & \textit{0.275} \\
    \midrule
    \rowc \multicolumn{2}{c}{\textbf{Average}} &      & \textbf{0.309} & \textbf{0.345} &      & 0.311  & 0.346  &      & 0.315  & 0.347  &      & \textit{0.327} & \textit{0.357} &      & \textit{0.310} & \textit{0.344} &      & \textit{0.329} & \textit{0.350} \\
    \rowc \multicolumn{2}{c}{\boldmath{}\textbf{$\bf 1^{st}$ count}\unboldmath{}} &      & \multicolumn{2}{c}{\textbf{38}} &      & \multicolumn{2}{c}{17} &      & \multicolumn{2}{c}{17} &      & \multicolumn{2}{c}{-} &      & \multicolumn{2}{c}{-} &      & \multicolumn{2}{c}{-} \\
    \bottomrule
    \end{tabular}%

}
  \label{tab:zero_shot_size}%
\end{table}%

%% file: table/tab_image_ablation.tex
\begin{table}[t]
  \centering
  \caption{Impact of resampling filters and image orientations.}
\vskip 0.15in
    \resizebox{\linewidth}{!}{

\begin{tabular}{cccccccccccrccccccccccc}
\toprule
     &      &      & \multicolumn{8}{c}{\textit{Interpolation strategies in resampling}} &      &      &      &      & \multicolumn{8}{c}{\textit{Image orientation}} \\
\multicolumn{2}{c}{\textbf{Method}} &      & \multicolumn{2}{c}{\textbf{Bilinear}} &      & \multicolumn{2}{c}{\textbf{Bicubic}} &      & \multicolumn{2}{c}{\textbf{Nearest Neighbor}} &  \quad\quad    & \multicolumn{2}{c}{\textbf{Method}} &      & \multicolumn{2}{c}{\textbf{-}} &      & \multicolumn{2}{c}{\textbf{Horizontal flip}} &      & \multicolumn{2}{c}{\textbf{Vertical flip}} \\
\cmidrule{4-5}\cmidrule{7-8}\cmidrule{10-11}\cmidrule{16-17}\cmidrule{19-20}\cmidrule{22-23}\multicolumn{2}{c}{\textbf{Metric}} &      & MSE  & MAE  &      & MSE  & MAE  &      & MSE  & MAE  &      & \multicolumn{2}{c}{\textbf{Metric}} &      & MSE  & MAE  &      & MSE  & MAE  &      & MSE  & MAE \\
\cmidrule{1-11}\cmidrule{13-23}\multirow{5}[2]{*}{\rotatebox{90}{$ETTh1$}} & \multicolumn{1}{r}{96} &      & 0.353  & \textbf{0.383} &      & \textbf{0.351} & \textbf{0.383} &      & 0.426  & 0.424  &      & \multirow{5}[2]{*}{\rotatebox{90}{$ETTh1$}} & \multicolumn{1}{r}{96} &      & 0.353  & 0.383  &      & \textbf{0.348} & \textbf{0.379} &      & 0.355  & 0.385  \\
     & \multicolumn{1}{r}{192} &      & \textbf{0.392} & 0.410  &      & \textbf{0.392} & \textbf{0.409} &      & 0.450  & 0.443  &      &      & \multicolumn{1}{r}{192} &      & 0.392  & 0.410  &      & \textbf{0.386} & \textbf{0.404} &      & 0.394  & 0.411  \\
     & \multicolumn{1}{r}{336} &      & \textbf{0.407} & 0.423  &      & \textbf{0.407} & \textbf{0.422} &      & 0.451  & 0.450  &      &      & \multicolumn{1}{r}{336} &      & 0.407  & 0.423  &      & \textbf{0.401} & \textbf{0.416} &      & 0.408  & 0.423  \\
     & \multicolumn{1}{r}{720} &      & 0.406  & 0.441  &      & \textbf{0.405} & \textbf{0.440} &      & 0.454  & 0.470  &      &      & \multicolumn{1}{r}{720} &      & 0.406  & 0.441  &      & \textbf{0.399} & \textbf{0.430} &      & 0.406  & 0.442  \\
     & \multicolumn{1}{r}{avg} &      & 0.390  & \textbf{0.414} &      & \textbf{0.389} & \textbf{0.414} &      & 0.445  & 0.446  &      &      & \multicolumn{1}{r}{avg} &      & 0.390  & 0.414  &      & \textbf{0.384} & \textbf{0.407} &      & 0.391  & 0.415  \\
\cmidrule{1-11}\cmidrule{13-23}\multirow{5}[2]{*}{\rotatebox{90}{$ETTh2$}} & \multicolumn{1}{r}{96} &      & \textbf{0.271} & \textbf{0.328} &      & 0.274  & 0.329  &      & 0.298  & 0.349  &      & \multirow{5}[2]{*}{\rotatebox{90}{$ETTh2$}} & \multicolumn{1}{r}{96} &      & \textbf{0.271} & \textbf{0.328} &      & 0.274  & 0.329  &      & 0.274  & 0.330  \\
     & \multicolumn{1}{r}{192} &      & \textbf{0.328} & \textbf{0.367} &      & 0.330  & \textbf{0.367} &      & 0.343  & 0.380  &      &      & \multicolumn{1}{r}{192} &      & \textbf{0.328} & \textbf{0.367} &      & 0.331  & 0.370  &      & 0.330  & \textbf{0.367} \\
     & \multicolumn{1}{r}{336} &      & \textbf{0.345} & 0.381  &      & \textbf{0.345} & \textbf{0.380} &      & 0.373  & 0.401  &      &      & \multicolumn{1}{r}{336} &      & \textbf{0.345} & \textbf{0.381} &      & 0.347  & 0.386  &      & \textbf{0.345} & \textbf{0.381} \\
     & \multicolumn{1}{r}{720} &      & 0.388  & 0.422  &      & \textbf{0.386} & \textbf{0.419} &      & 0.404  & 0.431  &      &      & \multicolumn{1}{r}{720} &      & 0.388  & 0.422  &      & \textbf{0.376} & \textbf{0.416} &      & 0.388  & 0.422  \\
     & \multicolumn{1}{r}{avg} &      & \textbf{0.333} & 0.375  &      & 0.334  & \textbf{0.374} &      & 0.354  & 0.390  &      &      & \multicolumn{1}{r}{avg} &      & 0.333  & \textbf{0.375} &      & \textbf{0.332} & \textbf{0.375} &      & 0.334  & \textbf{0.375} \\
\cmidrule{1-11}\cmidrule{13-23}\multirow{5}[2]{*}{\rotatebox{90}{$ETTm1$}} & \multicolumn{1}{r}{96} &      & \textbf{0.341} & \textbf{0.347} &      & 0.366  & 0.354  &      & 0.399  & 0.374  &      & \multirow{5}[2]{*}{\rotatebox{90}{$ETTm1$}} & \multicolumn{1}{r}{96} &      & \textbf{0.341} & \textbf{0.347} &      & 0.345  & 0.348  &      & 0.342  & \textbf{0.347} \\
     & \multicolumn{1}{r}{192} &      & \textbf{0.360} & \textbf{0.360} &      & 0.383  & 0.367  &      & 0.397  & 0.376  &      &      & \multicolumn{1}{r}{192} &      & \textbf{0.360} & \textbf{0.360} &      & 0.364  & 0.362  &      & \textbf{0.360} & \textbf{0.360} \\
     & \multicolumn{1}{r}{336} &      & \textbf{0.377} & \textbf{0.374} &      & 0.396  & 0.381  &      & 0.386  & 0.380  &      &      & \multicolumn{1}{r}{336} &      & \textbf{0.377} & \textbf{0.374} &      & 0.378  & 0.375  &      & \textbf{0.377} & \textbf{0.374} \\
     & \multicolumn{1}{r}{720} &      & \textbf{0.416} & \textbf{0.405} &      & 0.429  & 0.409  &      & 0.417  & 0.409  &      &      & \multicolumn{1}{r}{720} &      & \textbf{0.416} & \textbf{0.405} &      & 0.419  & 0.408  &      & 0.417  & \textbf{0.405} \\
     & \multicolumn{1}{r}{avg} &      & \textbf{0.374} & \textbf{0.372} &      & 0.393  & 0.378  &      & 0.400  & 0.384  &      &      & \multicolumn{1}{r}{avg} &      & \textbf{0.374} & \textbf{0.372} &      & 0.376  & 0.373  &      & \textbf{0.374} & \textbf{0.372} \\
\cmidrule{1-11}\cmidrule{13-23}\multirow{5}[2]{*}{\rotatebox{90}{$ETTm2$}} & \multicolumn{1}{r}{96} &      & \textbf{0.228} & \textbf{0.282} &      & 0.246  & 0.296  &      & 0.264  & 0.326  &      & \multirow{5}[2]{*}{\rotatebox{90}{$ETTm2$}} & \multicolumn{1}{r}{96} &      & \textbf{0.228} & \textbf{0.282} &      & 0.230  & 0.286  &      & \textbf{0.228} & 0.283  \\
     & \multicolumn{1}{r}{192} &      & \textbf{0.262} & \textbf{0.305} &      & 0.273  & 0.313  &      & 0.273  & 0.328  &      &      & \multicolumn{1}{r}{192} &      & \textbf{0.262} & \textbf{0.305} &      & 0.264  & 0.308  &      & \textbf{0.262} & \textbf{0.305} \\
     & \multicolumn{1}{r}{336} &      & \textbf{0.293} & \textbf{0.328} &      & 0.303  & 0.334  &      & 0.297  & 0.343  &      &      & \multicolumn{1}{r}{336} &      & \textbf{0.293} & \textbf{0.328} &      & 0.298  & 0.332  &      & \textbf{0.293} & \textbf{0.328} \\
     & \multicolumn{1}{r}{720} &      & 0.343  & 0.370  &      & 0.343  & 0.370  &      & \textbf{0.334} & \textbf{0.369} &      &      & \multicolumn{1}{r}{720} &      & \textbf{0.343} & 0.370  &      & 0.350  & 0.373  &      & \textbf{0.343} & \textbf{0.369} \\
     & \multicolumn{1}{r}{avg} &      & \textbf{0.282} & \textbf{0.321} &      & 0.291  & 0.328  &      & 0.292  & 0.341  &      &      & \multicolumn{1}{r}{avg} &      & \textbf{0.282} & \textbf{0.321} &      & 0.285  & 0.325  &      & \textbf{0.282} & \textbf{0.321} \\
\cmidrule{1-11}\cmidrule{13-23}\rowc\multicolumn{2}{c}{\textbf{Average}} &      & \textbf{0.344} & \textbf{0.370} &      & 0.352  & 0.373  &      & 0.373  & 0.391  &      & \multicolumn{2}{c}{\textbf{Average}} &      & \textbf{0.344} & \textbf{0.370} &      & \textbf{0.344} & \textbf{0.370} &      & 0.345  & 0.371  \\
\rowc\multicolumn{2}{c}{\boldmath{}\textbf{$\bf 1^{st}$ count}\unboldmath{}} &      & \multicolumn{2}{c}{\textbf{30}} &      & \multicolumn{2}{c}{18} &      & \multicolumn{2}{c}{2} &      & \multicolumn{2}{c}{\boldmath{}\textbf{$\bf 1^{st}$ count}\unboldmath{}} &      & \multicolumn{2}{c}{\textbf{28}} &      & \multicolumn{2}{c}{16} &      & \multicolumn{2}{c}{21} \\
\bottomrule
\end{tabular}%

}
  \label{tab:zero_shot_ablation}%
\end{table}%

%% file: figure/ratio/exp_ratio.tex
\begin{figure*}[t]
  \centering
  \subfigure[ETTh1]{
    \includegraphics[width=\hyperparameterwidth]{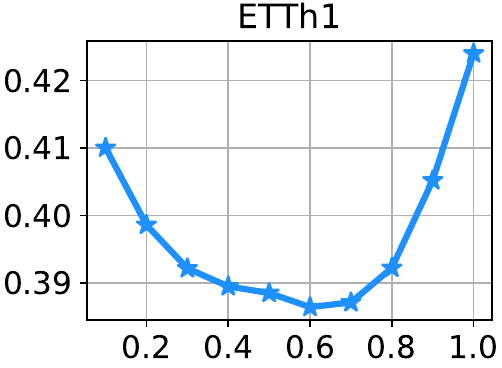}
  }
  \subfigure[ETTh2]{
    \includegraphics[width=\hyperparameterwidth]{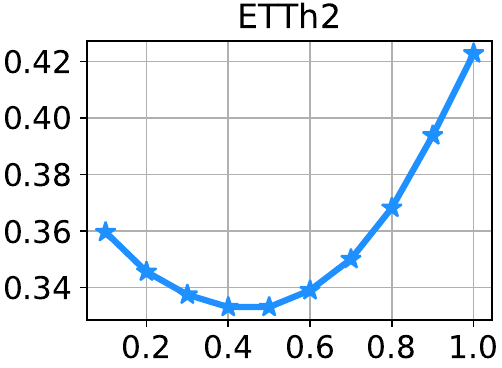}
  }
  \subfigure[ETTm1]{
    \includegraphics[width=\hyperparameterwidth]{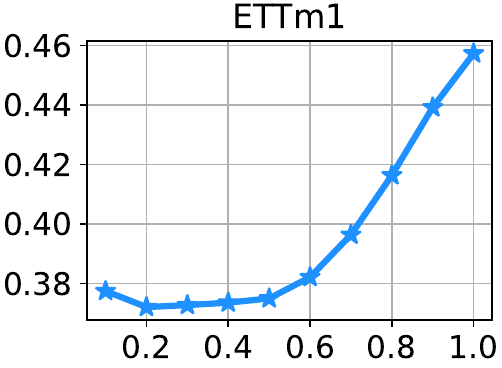}
  }\\
  \subfigure[ETTm2]{
    \includegraphics[width=\hyperparameterwidth]{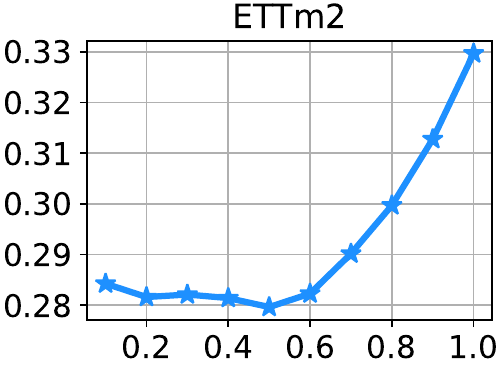}
  }
  \subfigure[Electricity]{
    \includegraphics[width=\hyperparameterwidth]{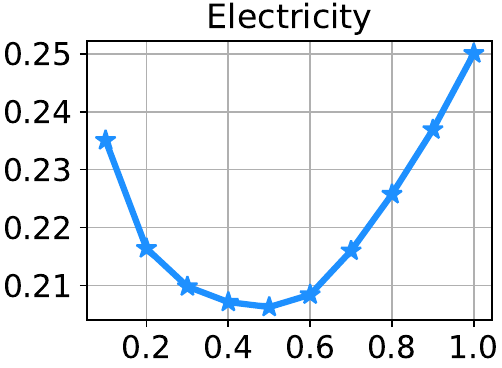}
  }
  \subfigure[Weather]{
    \includegraphics[width=\hyperparameterwidth]{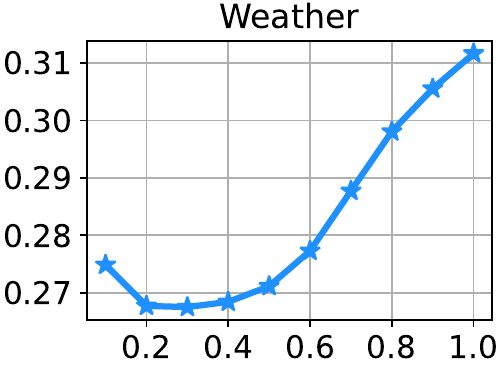}
  }
  \caption{MSE (Y-axis) performance of different normalization constants $r$ (X-axis).}
  \label{fig:ratio}
\end{figure*}

%% file: figure/scalex/exp_scalex.tex
\begin{figure*}[t]
  \centering
  \subfigure[ETTh1]{
    \includegraphics[width=\hyperparameterwidth]{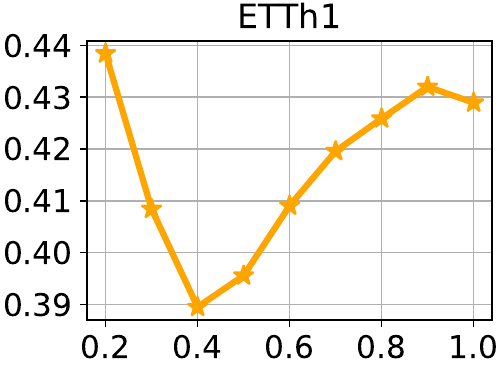}
  }
  \subfigure[ETTh2]{
    \includegraphics[width=\hyperparameterwidth]{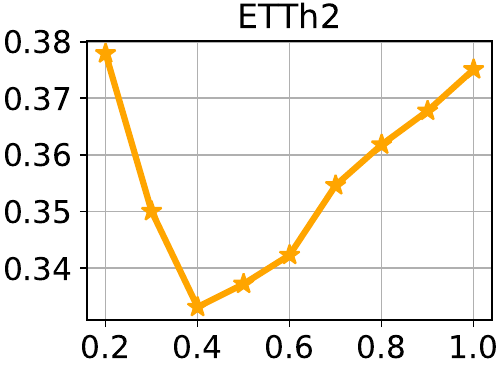}
  }
  \subfigure[ETTm1]{
    \includegraphics[width=\hyperparameterwidth]{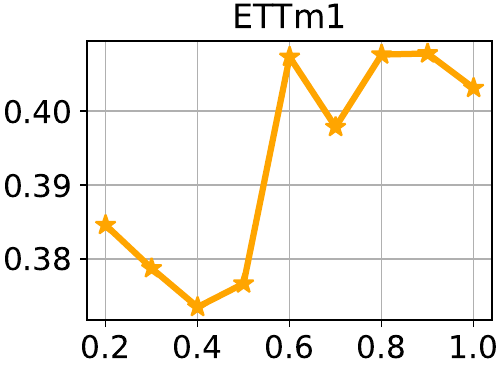}
  }\\
  \subfigure[ETTm2]{
    \includegraphics[width=\hyperparameterwidth]{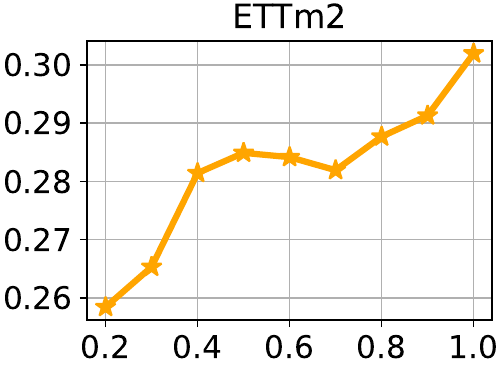}
  }
  \subfigure[Electricity]{
    \includegraphics[width=\hyperparameterwidth]{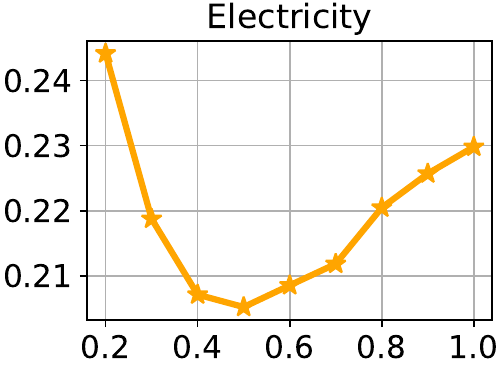}
  }
  \subfigure[Weather]{
    \includegraphics[width=\hyperparameterwidth]{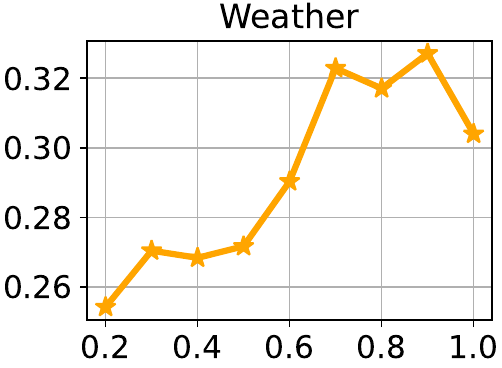}
  }
  \caption{MSE (Y-axis) performance of different alignment constants $c$ (X-axis).}
  \label{fig:scalex}
\end{figure*}

%% file: figure/sl/exp_sl.tex
\begin{figure*}[t]
  \centering
  \subfigure[ETTh1]{
    \includegraphics[width=\hyperparameterwidth]{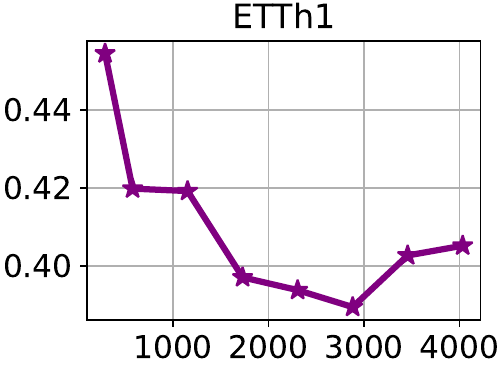}
  }
  \subfigure[ETTh2]{
    \includegraphics[width=\hyperparameterwidth]{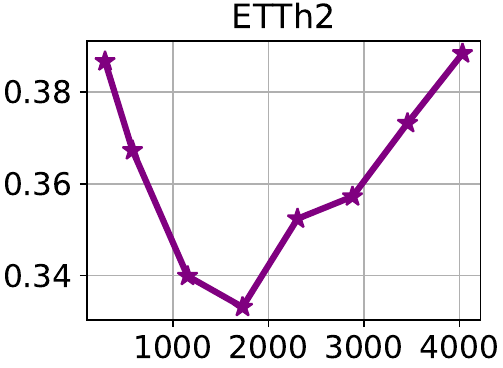}
  }
  \subfigure[ETTm1]{
    \includegraphics[width=\hyperparameterwidth]{figure/sl/exp_sl_zs_mse_ETTm1.pdf}
  }\\
  \subfigure[ETTm2]{
    \includegraphics[width=\hyperparameterwidth]{figure/sl/exp_sl_zs_mse_ETTm2.pdf}
  }
  \subfigure[Electricity]{
    \includegraphics[width=\hyperparameterwidth]{figure/sl/exp_sl_zs_mse_Electricity.pdf}
  }
  \subfigure[Weather]{
    \includegraphics[width=\hyperparameterwidth]{figure/sl/exp_sl_zs_mse_Weather.pdf}
  }
  \caption{MSE (Y-axis) performance of different context lengths $L$ (X-axis).}
  \label{fig:sl}
\end{figure*}

%% file: table/tab_full_shot_param.tex
\begin{table*}[h]
  \centering
  \caption{Final hyperparameters for \ours used in our full-shot forecasting.}
\vskip 0.15in
  \small
  
    \begin{tabular}{ccccccccc}
    \toprule
         & \textbf{ETTh1} & \textbf{ETTh2} & \textbf{ETTm1} & \textbf{ETTm2} & \textbf{Illness} & \textbf{Weather} & \textbf{Traffic} & \textbf{Electricity} \\
    \midrule
    Normalization constant $r$ & 0.4  & 0.4  & 0.4  & 0.4  & 1.0  & 1.0  & 0.4  & 0.4  \\
    Alignment constant $c$ & 0.4  & 0.4  & 0.4  & 0.4  & 0.4  & 0.7  & 0.4  & 0.4  \\
    Context length $L$ & 1152 & 1152 & 2304 & 1152 & 104 & 576  & 1152 & 1152 \\
    \bottomrule
    \end{tabular}%
  
  \label{tab:full_shot_tune}%
\end{table*}%

%% file: table/tab_full_shot_std.tex
\begin{table}[h]
  \centering
  \caption{Standard deviations of full-shot experiments.}
\vskip 0.15in
    \resizebox{0.6\linewidth}{!}{

\begin{tabular}{ccccccccccc}
\toprule
\multicolumn{2}{c}{\textbf{Method}} &      & \multicolumn{2}{c}{\textbf{\ours}} &      & \multicolumn{2}{c}{\textbf{Time-LLM}} &      & \multicolumn{2}{c}{\textbf{GPT4TS}} \\
\cmidrule{4-5}\cmidrule{7-8}\cmidrule{10-11}\multicolumn{2}{c}{\textbf{Metric}} &      & MSE  & MAE  &      & MSE  & MAE  &      & MSE  & MAE \\
\midrule
\multirow{4}[2]{*}{\rotatebox{90}{ETTh1}} & \multicolumn{1}{r}{96} &      & \boldmath{}\textbf{0.347 $\pm$ 0.002}\unboldmath{} & \boldmath{}\textbf{0.376 $\pm$ 0.000}\unboldmath{} &      & 0.376 $\pm$ 0.003 & 0.402 $\pm$ 0.002 &      & 0.370 $\pm$ 0.003 & 0.389 $\pm$ 0.001 \\
     & \multicolumn{1}{r}{192} &      & \boldmath{}\textbf{0.385 $\pm$ 0.001}\unboldmath{} & \boldmath{}\textbf{0.400 $\pm$ 0.000}\unboldmath{} &      & 0.407 $\pm$ 0.003 & 0.421 $\pm$ 0.002 &      & 0.412 $\pm$ 0.003 & 0.413 $\pm$ 0.001 \\
     & \multicolumn{1}{r}{336} &      & \boldmath{}\textbf{0.407 $\pm$ 0.001}\unboldmath{} & \boldmath{}\textbf{0.415 $\pm$ 0.001}\unboldmath{} &      & 0.430 $\pm$ 0.004 & 0.438 $\pm$ 0.001 &      & 0.448 $\pm$ 0.003 & 0.431 $\pm$ 0.001 \\
     & \multicolumn{1}{r}{720} &      & \boldmath{}\textbf{0.439 $\pm$ 0.001}\unboldmath{} & \boldmath{}\textbf{0.443 $\pm$ 0.000}\unboldmath{} &      & 0.457 $\pm$ 0.003 & 0.468 $\pm$ 0.001 &      & 0.441 $\pm$ 0.003 & 0.449 $\pm$ 0.001 \\
\midrule
\multirow{4}[2]{*}{\rotatebox{90}{ETTh2}} & \multicolumn{1}{r}{96} &      & \boldmath{}\textbf{0.269 $\pm$ 0.003}\unboldmath{} & \boldmath{}\textbf{0.328 $\pm$ 0.002}\unboldmath{} &      & 0.286 $\pm$ 0.003 & 0.346 $\pm$ 0.002 &      & 0.280 $\pm$ 0.001 & 0.335 $\pm$ 0.001 \\
     & \multicolumn{1}{r}{192} &      & \boldmath{}\textbf{0.332 $\pm$ 0.001}\unboldmath{} & \boldmath{}\textbf{0.374 $\pm$ 0.001}\unboldmath{} &      & 0.361 $\pm$ 0.003 & 0.391 $\pm$ 0.002 &      & 0.348 $\pm$ 0.002 & 0.380 $\pm$ 0.001 \\
     & \multicolumn{1}{r}{336} &      & \boldmath{}\textbf{0.351 $\pm$ 0.002}\unboldmath{} & \boldmath{}\textbf{0.395 $\pm$ 0.002}\unboldmath{} &      & 0.390 $\pm$ 0.003 & 0.414 $\pm$ 0.002 &      & 0.380 $\pm$ 0.002 & 0.405 $\pm$ 0.001 \\
     & \multicolumn{1}{r}{720} &      & \boldmath{}\textbf{0.390 $\pm$ 0.003}\unboldmath{} & \boldmath{}\textbf{0.430 $\pm$ 0.002}\unboldmath{} &      & 0.405 $\pm$ 0.003 & 0.434 $\pm$ 0.002 &      & 0.406 $\pm$ 0.002 & 0.436 $\pm$ 0.001 \\
\midrule
\multirow{4}[2]{*}{\rotatebox{90}{ETTm1}} & \multicolumn{1}{r}{96} &      & \boldmath{}\textbf{0.281 $\pm$ 0.001}\unboldmath{} & \boldmath{}\textbf{0.322 $\pm$ 0.001}\unboldmath{} &      & 0.291 $\pm$ 0.001 & 0.341 $\pm$ 0.001 &      & 0.300 $\pm$ 0.001 & 0.340 $\pm$ 0.000 \\
     & \multicolumn{1}{r}{192} &      & \boldmath{}\textbf{0.322 $\pm$ 0.006}\unboldmath{} & \boldmath{}\textbf{0.353 $\pm$ 0.002}\unboldmath{} &      & 0.341 $\pm$ 0.001 & 0.369 $\pm$ 0.001 &      & 0.343 $\pm$ 0.001 & 0.368 $\pm$ 0.000 \\
     & \multicolumn{1}{r}{336} &      & \boldmath{}\textbf{0.356 $\pm$ 0.003}\unboldmath{} & \boldmath{}\textbf{0.379 $\pm$ 0.002}\unboldmath{} &      & 0.359 $\pm$ 0.002 & 0.379 $\pm$ 0.001 &      & 0.376 $\pm$ 0.001 & 0.386 $\pm$ 0.000 \\
     & \multicolumn{1}{r}{720} &      & \boldmath{}\textbf{0.391 $\pm$ 0.001}\unboldmath{} & \boldmath{}\textbf{0.413 $\pm$ 0.001}\unboldmath{} &      & 0.433 $\pm$ 0.001 & 0.419 $\pm$ 0.001 &      & 0.431 $\pm$ 0.001 & 0.416 $\pm$ 0.000 \\
\midrule
\multirow{4}[2]{*}{\rotatebox{90}{ETTm2}} & \multicolumn{1}{r}{96} &      & 0.169 $\pm$ 0.003 & 0.256 $\pm$ 0.002 &      & \boldmath{}\textbf{0.162 $\pm$ 0.001}\unboldmath{} & \boldmath{}\textbf{0.248 $\pm$ 0.001}\unboldmath{} &      & 0.163 $\pm$ 0.001 & 0.249 $\pm$ 0.001 \\
     & \multicolumn{1}{r}{192} &      & 0.225 $\pm$ 0.003 & 0.294 $\pm$ 0.003 &      & 0.235 $\pm$ 0.002 & 0.304 $\pm$ 0.001 &      & \boldmath{}\textbf{0.222 $\pm$ 0.001}\unboldmath{} & \boldmath{}\textbf{0.291 $\pm$ 0.000}\unboldmath{} \\
     & \multicolumn{1}{r}{336} &      & 0.278 $\pm$ 0.002 & 0.334 $\pm$ 0.001 &      & 0.280 $\pm$ 0.002 & 0.329 $\pm$ 0.001 &      & \boldmath{}\textbf{0.273 $\pm$ 0.001}\unboldmath{} & \boldmath{}\textbf{0.327 $\pm$ 0.001}\unboldmath{} \\
     & \multicolumn{1}{r}{720} &      & 0.372 $\pm$ 0.002 & 0.392 $\pm$ 0.002 &      & 0.366 $\pm$ 0.002 & 0.382 $\pm$ 0.001 &      & \boldmath{}\textbf{0.357 $\pm$ 0.001}\unboldmath{} & \boldmath{}\textbf{0.376 $\pm$ 0.001}\unboldmath{} \\
\midrule
\multirow{4}[2]{*}{\rotatebox{90}{Weather}} & \multicolumn{1}{r}{96} &      & \boldmath{}\textbf{0.142 $\pm$ 0.000}\unboldmath{} & 0.192 $\pm$ 0.001 &      & 0.155 $\pm$ 0.001 & 0.199 $\pm$ 0.001 &      & 0.148 $\pm$ 0.001 & \boldmath{}\textbf{0.188 $\pm$ 0.000}\unboldmath{} \\
     & \multicolumn{1}{r}{192} &      & \boldmath{}\textbf{0.191 $\pm$ 0.000}\unboldmath{} & 0.238 $\pm$ 0.000 &      & 0.223 $\pm$ 0.001 & 0.261 $\pm$ 0.001 &      & 0.192 $\pm$ 0.001 & \boldmath{}\textbf{0.230 $\pm$ 0.000}\unboldmath{} \\
     & \multicolumn{1}{r}{336} &      & \boldmath{}\textbf{0.246 $\pm$ 0.003}\unboldmath{} & 0.282 $\pm$ 0.001 &      & 0.251 $\pm$ 0.001 & 0.279 $\pm$ 0.001 &      & \boldmath{}\textbf{0.246 $\pm$ 0.001}\unboldmath{} & \boldmath{}\textbf{0.273 $\pm$ 0.000}\unboldmath{} \\
     & \multicolumn{1}{r}{720} &      & 0.328 $\pm$ 0.004 & 0.337 $\pm$ 0.001 &      & 0.345 $\pm$ 0.001 & 0.342 $\pm$ 0.001 &      & \boldmath{}\textbf{0.320 $\pm$ 0.001}\unboldmath{} & \boldmath{}\textbf{0.328 $\pm$ 0.000}\unboldmath{} \\
\midrule
\multirow{4}[1]{*}{\rotatebox{90}{Traffic}} & \multicolumn{1}{r}{96} &      & \boldmath{}\textbf{0.344 $\pm$ 0.001}\unboldmath{} & \boldmath{}\textbf{0.236 $\pm$ 0.000}\unboldmath{} &      & 0.392 $\pm$ 0.001 & 0.267 $\pm$ 0.000 &      & 0.396 $\pm$ 0.001 & 0.264 $\pm$ 0.000 \\
     & \multicolumn{1}{r}{192} &      & \boldmath{}\textbf{0.372 $\pm$ 0.001}\unboldmath{} & \boldmath{}\textbf{0.249 $\pm$ 0.001}\unboldmath{} &      & 0.409 $\pm$ 0.001 & 0.271 $\pm$ 0.000 &      & 0.412 $\pm$ 0.001 & 0.268 $\pm$ 0.000 \\
     & \multicolumn{1}{r}{336} &      & \boldmath{}\textbf{0.383 $\pm$ 0.001}\unboldmath{} & \boldmath{}\textbf{0.257 $\pm$ 0.001}\unboldmath{} &      & 0.434 $\pm$ 0.001 & 0.296 $\pm$ 0.000 &      & 0.421 $\pm$ 0.001 & 0.273 $\pm$ 0.000 \\
     & \multicolumn{1}{r}{720} &      & \boldmath{}\textbf{0.422 $\pm$ 0.001}\unboldmath{} & \boldmath{}\textbf{0.280 $\pm$ 0.000}\unboldmath{} &      & 0.451 $\pm$ 0.001 & 0.291 $\pm$ 0.000 &      & 0.455 $\pm$ 0.001 & 0.291 $\pm$ 0.000 \\
\midrule
\multirow{4}[1]{*}{\rotatebox{90}{Electricity}} & \multicolumn{1}{r}{96} &      & \boldmath{}\textbf{0.126 $\pm$ 0.000}\unboldmath{} & \boldmath{}\textbf{0.218 $\pm$ 0.000}\unboldmath{} &      & 0.137 $\pm$ 0.000 & 0.233 $\pm$ 0.000 &      & 0.141 $\pm$ 0.000 & 0.239 $\pm$ 0.000 \\
     & \multicolumn{1}{r}{192} &      & \boldmath{}\textbf{0.146 $\pm$ 0.001}\unboldmath{} & \boldmath{}\textbf{0.239 $\pm$ 0.001}\unboldmath{} &      & 0.152 $\pm$ 0.000 & 0.247 $\pm$ 0.000 &      & 0.158 $\pm$ 0.000 & 0.253 $\pm$ 0.000 \\
     & \multicolumn{1}{r}{336} &      & \boldmath{}\textbf{0.161 $\pm$ 0.001}\unboldmath{} & \boldmath{}\textbf{0.255 $\pm$ 0.001}\unboldmath{} &      & 0.169 $\pm$ 0.000 & 0.267 $\pm$ 0.000 &      & 0.172 $\pm$ 0.000 & 0.266 $\pm$ 0.000 \\
     & \multicolumn{1}{r}{720} &      & \boldmath{}\textbf{0.193 $\pm$ 0.000}\unboldmath{} & \boldmath{}\textbf{0.286 $\pm$ 0.000}\unboldmath{} &      & 0.200 $\pm$ 0.000 & 0.290 $\pm$ 0.000 &      & 0.207 $\pm$ 0.000 & 0.293 $\pm$ 0.000 \\
\midrule
\rowc
\multicolumn{2}{c}{\boldmath{}\textbf{$\bf 1^{st}$ count}\unboldmath{}} &      & \multicolumn{2}{c}{\textbf{42}} &      & \multicolumn{2}{c}{2} &      & \multicolumn{2}{c}{12} \\
\bottomrule
\end{tabular}%

}
  \label{tab:full_shot_std}%
\end{table}%

%% file: table/tab_full_shot.tex
\begin{table*}[t!]
  \centering
  \caption{Full results of \cref{tab:full_shot_short}: Full-shot forecasting performance on the long-term TSF benchmark. \ours is fine-tuned only a single epoch on each dataset except for Illness.}
\vskip 0.15in
    \resizebox{\linewidth}{!}{

\begin{tabular}{ccccccccccccccccccccccccccccccccccc}
\toprule
\multicolumn{2}{c}{\textbf{Pretrain}} &      & \multicolumn{2}{c}{\textit{\textbf{\emoji{figure/em-frame_with_picture} Images}}} &      & \multicolumn{5}{c}{\textit{\textbf{\emoji{figure/em-memo} Text}}} &      & \multicolumn{23}{c}{\textit{\textbf{\emoji{figure/em-no_entry_sign} No Pretrain}}} \\
\cmidrule{4-5}\cmidrule{7-11}\cmidrule{13-35}\multicolumn{2}{c}{\textbf{Method}} &      & \multicolumn{2}{c}{\textbf{\ours}} &      & \multicolumn{2}{c}{\textbf{Time-LLM}} &      & \multicolumn{2}{c}{\textbf{GPT4TS}} &      & \multicolumn{2}{c}{\textbf{DLinear}} &      & \multicolumn{2}{c}{\textbf{PatchTST}} &      & \multicolumn{2}{c}{\textbf{TimesNet}} &      & \multicolumn{2}{c}{\textbf{FEDformer}} &      & \multicolumn{2}{c}{\textbf{Autoformer}} &      & \multicolumn{2}{c}{\textbf{Stationary}} &      & \multicolumn{2}{c}{\textbf{ETSformer}} &      & \multicolumn{2}{c}{\textbf{Informer}} \\
\cmidrule{4-5}\cmidrule{7-8}\cmidrule{10-11}\cmidrule{13-14}\cmidrule{16-17}\cmidrule{19-20}\cmidrule{22-23}\cmidrule{25-26}\cmidrule{28-29}\cmidrule{31-32}\cmidrule{34-35}\multicolumn{2}{c}{\textbf{Metric}} &      & MSE  & MAE  &      & MSE  & MAE  &      & MSE  & MAE  &      & MSE  & MAE  &      & MSE  & MAE  &      & MSE  & MAE  &      & MSE  & MAE  &      & MSE  & MAE  &      & MSE  & MAE  &      & MSE  & MAE  &      & MSE  & MAE \\
\midrule
\multirow{5}[2]{*}{\rotatebox{90}{$ETTh1$}} & \multicolumn{1}{r}{96} &      & \textbf{0.347} & \textbf{0.376} &      & 0.376  & 0.402  &      & 0.370  & 0.389  &      & 0.375  & 0.399  &      & 0.370  & 0.399  &      & 0.384  & 0.402  &      & 0.376  & 0.419  &      & 0.449  & 0.459  &      & 0.513  & 0.491  &      & 0.494  & 0.479  &      & 0.865  & 0.713  \\
     & \multicolumn{1}{r}{192} &      & \textbf{0.385} & \textbf{0.400} &      & 0.407  & 0.421  &      & 0.412  & 0.413  &      & 0.405  & 0.416  &      & 0.413  & 0.421  &      & 0.436  & 0.429  &      & 0.420  & 0.448  &      & 0.500  & 0.482  &      & 0.534  & 0.504  &      & 0.538  & 0.504  &      & 1.008  & 0.792  \\
     & \multicolumn{1}{r}{336} &      & \textbf{0.407} & \textbf{0.415} &      & 0.430  & 0.438  &      & 0.448  & 0.431  &      & 0.439  & 0.443  &      & 0.422  & 0.436  &      & 0.491  & 0.469  &      & 0.459  & 0.465  &      & 0.521  & 0.496  &      & 0.588  & 0.535  &      & 0.574  & 0.521  &      & 1.107  & 0.809  \\
     & \multicolumn{1}{r}{720} &      & \textbf{0.439} & \textbf{0.443} &      & 0.457  & 0.468  &      & 0.441  & 0.449  &      & 0.472  & 0.490  &      & 0.447  & 0.466  &      & 0.521  & 0.500  &      & 0.506  & 0.507  &      & 0.514  & 0.512  &      & 0.643  & 0.616  &      & 0.562  & 0.535  &      & 1.181  & 0.865  \\
     & \multicolumn{1}{r}{avg} &      & \textbf{0.395} & \textbf{0.409} &      & 0.418  & 0.432  &      & 0.418  & 0.421  &      & 0.423  & 0.437  &      & 0.413  & 0.431  &      & 0.458  & 0.450  &      & 0.440  & 0.460  &      & 0.496  & 0.487  &      & 0.570  & 0.537  &      & 0.542  & 0.510  &      & 1.040  & 0.795  \\
\midrule
\multirow{5}[2]{*}{\rotatebox{90}{$ETTh2$}} & \multicolumn{1}{r}{96} &      & \textbf{0.269} & \textbf{0.328} &      & 0.286  & 0.346  &      & 0.280  & 0.335  &      & 0.289  & 0.353  &      & 0.274  & 0.336  &      & 0.340  & 0.374  &      & 0.358  & 0.397  &      & 0.346  & 0.388  &      & 0.476  & 0.458  &      & 0.340  & 0.391  &      & 3.755  & 1.525  \\
     & \multicolumn{1}{r}{192} &      & \textbf{0.332} & \textbf{0.374} &      & 0.361  & 0.391  &      & 0.348  & 0.380  &      & 0.383  & 0.418  &      & 0.339  & 0.379  &      & 0.402  & 0.414  &      & 0.429  & 0.439  &      & 0.456  & 0.452  &      & 0.512  & 0.493  &      & 0.430  & 0.439  &      & 5.602  & 1.931  \\
     & \multicolumn{1}{r}{336} &      & 0.351  & 0.395  &      & 0.390  & 0.414  &      & 0.380  & 0.405  &      & 0.448  & 0.465  &      & \textbf{0.329} & \textbf{0.380} &      & 0.452  & 0.452  &      & 0.496  & 0.487  &      & 0.482  & 0.486  &      & 0.552  & 0.551  &      & 0.485  & 0.479  &      & 4.721  & 1.835  \\
     & \multicolumn{1}{r}{720} &      & 0.390  & 0.430  &      & 0.405  & 0.434  &      & 0.406  & 0.436  &      & 0.605  & 0.551  &      & \textbf{0.379} & \textbf{0.422} &      & 0.462  & 0.468  &      & 0.463  & 0.474  &      & 0.515  & 0.511  &      & 0.562  & 0.560  &      & 0.500  & 0.497  &      & 3.647  & 1.625  \\
     & \multicolumn{1}{r}{avg} &      & 0.336  & 0.382  &      & 0.361  & 0.396  &      & 0.354  & 0.389  &      & 0.431  & 0.447  &      & \textbf{0.330} & \textbf{0.379} &      & 0.414  & 0.427  &      & 0.437  & 0.449  &      & 0.450  & 0.459  &      & 0.526  & 0.516  &      & 0.439  & 0.452  &      & 4.431  & 1.729  \\
\midrule
\multirow{5}[2]{*}{\rotatebox{90}{$ETTm1$}} & \multicolumn{1}{r}{96} &      & \textbf{0.281} & \textbf{0.322} &      & 0.291  & 0.341  &      & 0.300  & 0.340  &      & 0.299  & 0.343  &      & 0.290  & 0.342  &      & 0.338  & 0.375  &      & 0.379  & 0.419  &      & 0.505  & 0.475  &      & 0.386  & 0.398  &      & 0.375  & 0.398  &      & 0.672  & 0.571  \\
     & \multicolumn{1}{r}{192} &      & \textbf{0.322} & \textbf{0.353} &      & 0.341  & 0.369  &      & 0.343  & 0.368  &      & 0.335  & 0.365  &      & 0.332  & 0.369  &      & 0.374  & 0.387  &      & 0.426  & 0.441  &      & 0.553  & 0.496  &      & 0.459  & 0.444  &      & 0.408  & 0.410  &      & 0.795  & 0.669  \\
     & \multicolumn{1}{r}{336} &      & \textbf{0.356} & \textbf{0.379} &      & 0.359  & \textbf{0.379} &      & 0.376  & 0.386  &      & 0.369  & 0.386  &      & 0.366  & 0.392  &      & 0.410  & 0.411  &      & 0.445  & 0.459  &      & 0.621  & 0.537  &      & 0.495  & 0.464  &      & 0.435  & 0.428  &      & 1.212  & 0.871  \\
     & \multicolumn{1}{r}{720} &      & \textbf{0.391} & \textbf{0.413} &      & 0.433  & 0.419  &      & 0.431  & 0.416  &      & 0.425  & 0.421  &      & 0.416  & 0.420  &      & 0.478  & 0.450  &      & 0.543  & 0.490  &      & 0.671  & 0.561  &      & 0.585  & 0.516  &      & 0.499  & 0.462  &      & 1.166  & 0.823  \\
     & \multicolumn{1}{r}{avg} &      & \textbf{0.338} & \textbf{0.367} &      & 0.356  & 0.377  &      & 0.363  & 0.378  &      & 0.357  & 0.379  &      & 0.351  & 0.381  &      & 0.400  & 0.406  &      & 0.448  & 0.452  &      & 0.588  & 0.517  &      & 0.481  & 0.456  &      & 0.429  & 0.425  &      & 0.961  & 0.734  \\
\midrule
\multirow{5}[2]{*}{\rotatebox{90}{$ETTm2$}} & \multicolumn{1}{r}{96} &      & 0.169  & 0.256  &      & \textbf{0.162} & \textbf{0.248} &      & 0.163  & 0.249  &      & 0.167  & 0.269  &      & 0.165  & 0.255  &      & 0.187  & 0.267  &      & 0.203  & 0.287  &      & 0.255  & 0.339  &      & 0.192  & 0.274  &      & 0.189  & 0.280  &      & 0.365  & 0.453  \\
     & \multicolumn{1}{r}{192} &      & 0.225  & 0.294  &      & 0.235  & 0.304  &      & 0.222  & \textbf{0.291} &      & 0.224  & 0.303  &      & \textbf{0.220} & 0.292  &      & 0.249  & 0.309  &      & 0.269  & 0.328  &      & 0.281  & 0.340  &      & 0.280  & 0.339  &      & 0.253  & 0.319  &      & 0.533  & 0.563  \\
     & \multicolumn{1}{r}{336} &      & 0.278  & 0.334  &      & 0.280  & 0.329  &      & \textbf{0.273} & \textbf{0.327} &      & 0.281  & 0.342  &      & 0.274  & 0.329  &      & 0.321  & 0.351  &      & 0.325  & 0.366  &      & 0.339  & 0.372  &      & 0.334  & 0.361  &      & 0.314  & 0.357  &      & 1.363  & 0.887  \\
     & \multicolumn{1}{r}{720} &      & 0.372  & 0.392  &      & 0.366  & 0.382  &      & \textbf{0.357} & \textbf{0.376} &      & 0.397  & 0.421  &      & 0.362  & 0.385  &      & 0.408  & 0.403  &      & 0.421  & 0.415  &      & 0.433  & 0.432  &      & 0.417  & 0.413  &      & 0.414  & 0.413  &      & 3.379  & 1.338  \\
     & \multicolumn{1}{r}{avg} &      & 0.261  & 0.319  &      & 0.261  & 0.316  &      & \textbf{0.254} & \textbf{0.311} &      & 0.267  & 0.334  &      & 0.255  & 0.315  &      & 0.291  & 0.333  &      & 0.305  & 0.349  &      & 0.327  & 0.371  &      & 0.306  & 0.347  &      & 0.293  & 0.342  &      & 1.410  & 0.810  \\
\midrule
\multirow{5}[2]{*}{\rotatebox{90}{$Illness$}} & \multicolumn{1}{r}{24} &      & 2.034  & 0.937  &      & 1.792  & 0.807  &      & 1.869  & 0.823  &      & 2.215  & 1.081  &      & \textbf{1.319} & \textbf{0.754} &      & 2.317  & 0.934  &      & 3.228  & 1.260  &      & 3.483  & 1.287  &      & 2.294  & 0.945  &      & 2.527  & 1.020  &      & 5.764  & 1.677  \\
     & \multicolumn{1}{r}{36} &      & 1.866  & 0.888  &      & 1.833  & \textbf{0.833} &      & 1.853  & 0.854  &      & 1.963  & 0.963  &      & \textbf{1.430} & 0.834  &      & 1.972  & 0.920  &      & 2.679  & 1.080  &      & 3.103  & 1.148  &      & 1.825  & 0.848  &      & 2.615  & 1.007  &      & 4.755  & 1.467  \\
     & \multicolumn{1}{r}{48} &      & 1.784  & 0.870  &      & 2.269  & 1.012  &      & 1.886  & 0.855  &      & 2.130  & 1.024  &      & \textbf{1.553} & \textbf{0.815} &      & 2.238  & 0.940  &      & 2.622  & 1.078  &      & 2.669  & 1.085  &      & 2.010  & 0.900  &      & 2.359  & 0.972  &      & 4.763  & 1.469  \\
     & \multicolumn{1}{r}{60} &      & 1.910  & 0.912  &      & 2.177  & 0.925  &      & 1.877  & 0.877  &      & 2.368  & 1.096  &      & \textbf{1.470} & \textbf{0.788} &      & 2.027  & 0.928  &      & 2.857  & 1.157  &      & 2.770  & 1.125  &      & 2.178  & 0.963  &      & 2.487  & 1.016  &      & 5.264  & 1.564  \\
     & \multicolumn{1}{r}{avg} &      & 1.899  & 0.902  &      & 2.018  & 0.894  &      & 1.871  & 0.852  &      & 2.169  & 1.041  &      & \textbf{1.443} & \textbf{0.798} &      & 2.139  & 0.931  &      & 2.847  & 1.144  &      & 3.006  & 1.161  &      & 2.077  & 0.914  &      & 2.497  & 1.004  &      & 5.137  & 1.544  \\
\midrule
\multirow{5}[2]{*}{\rotatebox{90}{$Weather$}} & \multicolumn{1}{r}{96} &      & \textbf{0.142} & 0.192 &      & 0.155  & 0.199  &      & 0.148  & \textbf{0.188} &      & 0.176  & 0.237  &      & 0.149  & 0.198  &      & 0.172  & 0.220  &      & 0.217  & 0.296  &      & 0.266  & 0.336  &      & 0.173  & 0.223  &      & 0.197  & 0.281  &      & 0.300  & 0.384  \\
     & \multicolumn{1}{r}{192} &      & \textbf{0.191} & 0.238 &      & 0.223  & 0.261  &      & 0.192  & \textbf{0.230} &      & 0.220  & 0.282  &      & 0.194  & 0.241  &      & 0.219  & 0.261  &      & 0.276  & 0.336  &      & 0.307  & 0.367  &      & 0.245  & 0.285  &      & 0.237  & 0.312  &      & 0.598  & 0.544  \\
     & \multicolumn{1}{r}{336} &      & 0.246 & 0.282 &      & 0.251  & 0.279  &      & 0.246  & \textbf{0.273} &      & 0.265  & 0.319  &      & \textbf{0.245} & 0.282  &      & 0.280  & 0.306  &      & 0.339  & 0.380  &      & 0.359  & 0.395  &      & 0.321  & 0.338  &      & 0.298  & 0.353  &      & 0.578  & 0.523  \\
     & \multicolumn{1}{r}{720} &      & 0.328 & 0.337 &      & 0.345  & 0.342  &      & 0.320  & \textbf{0.328} &      & 0.333  & 0.362  &      & \textbf{0.314} & 0.334  &      & 0.365  & 0.359  &      & 0.403  & 0.428  &      & 0.419  & 0.428  &      & 0.414  & 0.410  &      & 0.352  & 0.388  &      & 1.059  & 0.741  \\
     & \multicolumn{1}{r}{avg} &      & 0.227  & 0.262  &      & 0.244  & 0.270  &      & 0.227  & \textbf{0.255} &      & 0.249  & 0.300  &      & \textbf{0.226} & 0.264  &      & 0.259  & 0.287  &      & 0.309  & 0.360  &      & 0.338  & 0.382  &      & 0.288  & 0.314  &      & 0.271  & 0.334  &      & 0.634  & 0.548  \\
\midrule
\multirow{5}[2]{*}{\rotatebox{90}{$Traffic$}} & \multicolumn{1}{r}{96} &      & \textbf{0.344} & \textbf{0.236} &      & 0.392  & 0.267  &      & 0.396  & 0.264  &      & 0.410  & 0.282  &      & 0.360  & 0.249  &      & 0.593  & 0.321  &      & 0.587  & 0.366  &      & 0.613  & 0.388  &      & 0.612  & 0.338  &      & 0.607  & 0.392  &      & 0.719  & 0.391  \\
     & \multicolumn{1}{r}{192} &      & \textbf{0.372} & \textbf{0.249} &      & 0.409  & 0.271  &      & 0.412  & 0.268  &      & 0.423  & 0.287  &      & 0.379  & 0.256  &      & 0.617  & 0.336  &      & 0.604  & 0.373  &      & 0.616  & 0.382  &      & 0.613  & 0.340  &      & 0.621  & 0.399  &      & 0.696  & 0.379  \\
     & \multicolumn{1}{r}{336} &      & \textbf{0.383} & \textbf{0.257} &      & 0.434  & 0.296  &      & 0.421  & 0.273  &      & 0.436  & 0.296  &      & 0.392  & 0.264  &      & 0.629  & 0.336  &      & 0.621  & 0.383  &      & 0.622  & 0.337  &      & 0.618  & 0.328  &      & 0.622  & 0.396  &      & 0.777  & 0.420  \\
     & \multicolumn{1}{r}{720} &      & \textbf{0.422} & \textbf{0.280} &      & 0.451  & 0.291  &      & 0.455  & 0.291  &      & 0.466  & 0.315  &      & 0.432  & 0.286  &      & 0.640  & 0.350  &      & 0.626  & 0.382  &      & 0.660  & 0.408  &      & 0.653  & 0.355  &      & 0.632  & 0.396  &      & 0.864  & 0.472  \\
     & \multicolumn{1}{r}{avg} &      & \textbf{0.380} & \textbf{0.256} &      & 0.422  & 0.281  &      & 0.421  & 0.274  &      & 0.434  & 0.295  &      & 0.391  & 0.264  &      & 0.620  & 0.336  &      & 0.610  & 0.376  &      & 0.628  & 0.379  &      & 0.624  & 0.340  &      & 0.621  & 0.396  &      & 0.764  & 0.416  \\
\midrule
\multirow{5}[2]{*}{\rotatebox{90}{$Electricity$}} & \multicolumn{1}{r}{96} &      & \textbf{0.126} & \textbf{0.218} &      & 0.137  & 0.233  &      & 0.141  & 0.239  &      & 0.140  & 0.237  &      & 0.129  & 0.222  &      & 0.168  & 0.272  &      & 0.193  & 0.308  &      & 0.201  & 0.317  &      & 0.169  & 0.273  &      & 0.187  & 0.304  &      & 0.274  & 0.368  \\
     & \multicolumn{1}{r}{192} &      & \textbf{0.144} & \textbf{0.237} &      & 0.152  & 0.247  &      & 0.158  & 0.253  &      & 0.153  & 0.249  &      & 0.157  & 0.240  &      & 0.184  & 0.289  &      & 0.201  & 0.315  &      & 0.222  & 0.334  &      & 0.182  & 0.286  &      & 0.199  & 0.315  &      & 0.296  & 0.386  \\
     & \multicolumn{1}{r}{336} &      & \textbf{0.162} & \textbf{0.256} &      & 0.169  & 0.267  &      & 0.172  & 0.266  &      & 0.169  & 0.267  &      & 0.163  & 0.259  &      & 0.198  & 0.300  &      & 0.214  & 0.329  &      & 0.231  & 0.338  &      & 0.200  & 0.304  &      & 0.212  & 0.329  &      & 0.300  & 0.394  \\
     & \multicolumn{1}{r}{720} &      & \textbf{0.192} & \textbf{0.286} &      & 0.200  & 0.290  &      & 0.207  & 0.293  &      & 0.203  & 0.301  &      & 0.197  & 0.290  &      & 0.220  & 0.320  &      & 0.246  & 0.355  &      & 0.254  & 0.361  &      & 0.222  & 0.321  &      & 0.233  & 0.345  &      & 0.373  & 0.439  \\
     & \multicolumn{1}{r}{avg} &      & \textbf{0.156} & \textbf{0.249} &      & 0.165  & 0.259  &      & 0.170  & 0.263  &      & 0.166  & 0.264  &      & 0.162  & 0.253  &      & 0.193  & 0.295  &      & 0.214  & 0.327  &      & 0.227  & 0.338  &      & 0.193  & 0.296  &      & 0.208  & 0.323  &      & 0.311  & 0.397  \\
\midrule
\rowc \multicolumn{2}{c}{\textbf{$\bf 1^{st}$ count}} &      & \multicolumn{2}{c}{\textbf{46}} &      & \multicolumn{2}{c}{4} &      & \multicolumn{2}{c}{12} &      & \multicolumn{2}{c}{0} &      & \multicolumn{2}{c}{19} &      & \multicolumn{2}{c}{0} &      & \multicolumn{2}{c}{0} &      & \multicolumn{2}{c}{0} &      & \multicolumn{2}{c}{0} &      & \multicolumn{2}{c}{0} &      & \multicolumn{2}{c}{0} \\
\bottomrule
\end{tabular}%

}
  \label{tab:full_shot}%
\end{table*}%

%% file: table/tab_ablation.tex
\begin{table}[h]
  \centering
  \caption{Ablation studies (left) and fine-tuning strategies (right). Results are averaged on four prediction lengths: \{96, 192, 336, 720\}.}
\vskip 0.15in
    \tiny
    \resizebox{\linewidth}{!}{

\begin{tabular}{cccccccccccccccccc}
\toprule
     &      &      & \multicolumn{5}{c}{\textit{Ablation on Visual \mae (VM)}} & \multicolumn{1}{l}{\quad\quad} &      &      &      & \multicolumn{6}{c}{\textit{Ablation on trained parameters}} \\
     &      &      & \textbf{-} & \textbf{w/o VM} & \textbf{VM2Attn} & \textbf{VM2Trsf} & \textbf{Rand-VM} &      &      &      &      & \textbf{All} & \textbf{LN} & \textbf{Bias} & \textbf{MLP} & \textbf{Attn} & \textbf{Freeze} \\
\cmidrule{1-8}\cmidrule{10-18}\multicolumn{2}{c}{\multirow{2}[2]{*}{ETTh1}} & MSE  & \textbf{0.395} & 0.785  & 0.448  & 0.459  & 0.534  &      & \multicolumn{2}{c}{\multirow{2}[2]{*}{ETTh1}} & MSE  & 0.534  & \textbf{0.395} & 0.401  & 0.534  & 0.554  & 0.419  \\
\multicolumn{2}{c}{} & MAE  & \textbf{0.409} & 0.649  & 0.458  & 0.462  & 0.470  &      & \multicolumn{2}{c}{} & MAE  & 0.470  & \textbf{0.409} & 0.414  & 0.471  & 0.479  & 0.418  \\
\cmidrule{1-8}\cmidrule{10-18}\multicolumn{2}{c}{\multirow{2}[2]{*}{ETTh2}} & MSE  & \textbf{0.336} & 0.420  & 0.418  & 0.448  & 0.411  &      & \multicolumn{2}{c}{\multirow{2}[2]{*}{ETTh2}} & MSE  & 0.411  & \textbf{0.336} & 0.347  & 0.401  & 0.392  & 0.340  \\
\multicolumn{2}{c}{} & MAE  & \textbf{0.382} & 0.453  & 0.445  & 0.457  & 0.432  &      & \multicolumn{2}{c}{} & MAE  & 0.432  & 0.382  & 0.392  & 0.419  & 0.414  & \textbf{0.376} \\
\cmidrule{1-8}\cmidrule{10-18}\multicolumn{2}{c}{\multirow{2}[2]{*}{ETTm1}} & MSE  & \textbf{0.338} & 0.676  & 0.397  & 0.398  & 0.433  &      & \multicolumn{2}{c}{\multirow{2}[2]{*}{ETTm1}} & MSE  & 0.433  & \textbf{0.338} & 0.343  & 0.441  & 0.444  & 0.374  \\
\multicolumn{2}{c}{} & MAE  & \textbf{0.367} & 0.562  & 0.415  & 0.410  & 0.413  &      & \multicolumn{2}{c}{} & MAE  & 0.413  & \textbf{0.367} & 0.368  & 0.415  & 0.415  & 0.372  \\
\cmidrule{1-8}\cmidrule{10-18}\multicolumn{2}{c}{\multirow{2}[2]{*}{ETTm2}} & MSE  & \textbf{0.261} & 0.379  & 0.274  & 0.292  & 0.288  &      & \multicolumn{2}{c}{\multirow{2}[2]{*}{ETTm2}} & MSE  & 0.288  & 0.261  & \textbf{0.256} & 0.292  & 0.289  & 0.305  \\
\multicolumn{2}{c}{} & MAE  & \textbf{0.319} & 0.415  & 0.334  & 0.344  & 0.341  &      & \multicolumn{2}{c}{} & MAE  & 0.341  & 0.319  & \textbf{0.318} & 0.342  & 0.339  & 0.334  \\
\cmidrule{1-8}\cmidrule{10-18}\multicolumn{2}{c}{\multirow{2}[1]{*}{\textbf{Average}}} & MSE  & \textbf{0.333} & 0.565  & 0.384  & 0.399  & 0.417  &      & \multicolumn{2}{c}{\multirow{2}[1]{*}{\textbf{Average}}} & MSE  & 0.417  & \textbf{0.333} & 0.337  & 0.417  & 0.420  & 0.360  \\
\multicolumn{2}{c}{} & MAE  & \textbf{0.369} & 0.520  & 0.413  & 0.418  & 0.414  &      & \multicolumn{2}{c}{} & MAE  & 0.414  & \textbf{0.369} & 0.373  & 0.412  & 0.412  & 0.375  \\
\rowc
\multicolumn{3}{c}{\boldmath{}\textbf{$\bf 1^{st}$ count}\unboldmath{}} & \textbf{10} & 0    & 0    & 0    & 0    &      & \multicolumn{3}{c}{\boldmath{}\textbf{$\bf 1^{st}$ count}\unboldmath{}} & 0    & \textbf{7} & 2    & 0    & 0    & 1 \\
\bottomrule
\end{tabular}%

}
  \label{tab:full_shot_ablation}%
\end{table}%

%% file: figure/etth1_case1/etth1_case1.tex
\begin{figure*}[t]
  \centering
  \subfigure[Input Image]{
    \includegraphics[width=0.2\textwidth]{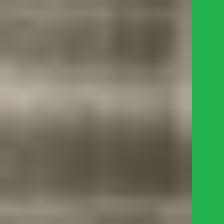}
  }\quad\quad\quad\quad
  \subfigure[Reconstructed Image]{
    \includegraphics[width=0.2\textwidth]{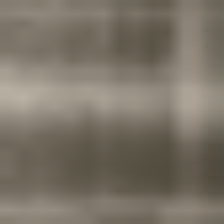}
  }
  \subfigure[\ours (MAE = 0.312)]{
    \includegraphics[width=0.8\textwidth]{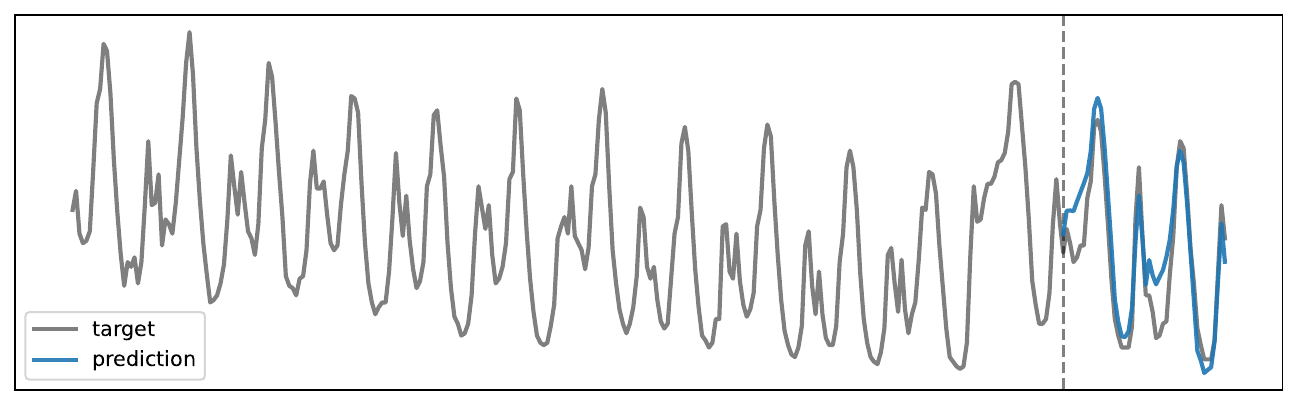}
  }
  \subfigure[\method{Moirai\textsubscript{Large}}~(MAE = 0.503)]{
    \includegraphics[width=0.8\textwidth]{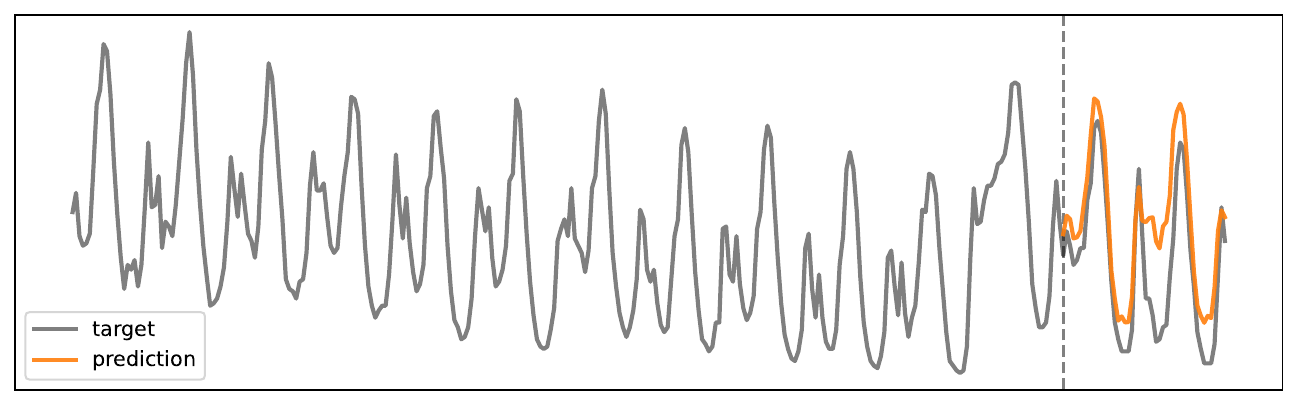}
  }
  \subfigure[Seasonal Naïve (MAE = 0.774)]{
    \includegraphics[width=0.8\textwidth]{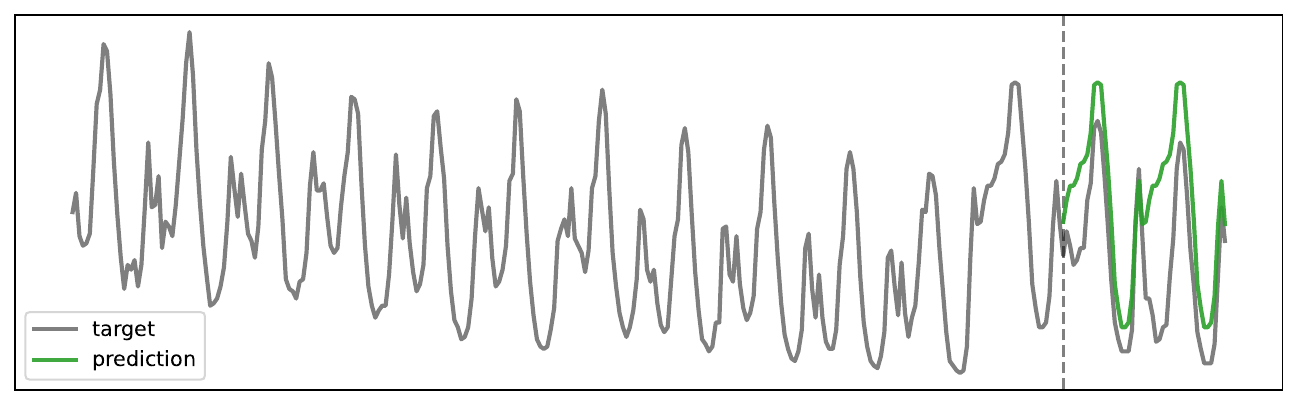}
  }
  \caption{Forecasting visualization on a sample from ETTh1. (a-b) Input/output images of \ours. (c-e) Forecasting visualization.}
  \label{fig:etth1_case1}
\end{figure*}

%% file: figure/etth2_case1/etth2_case1.tex
\begin{figure*}[t]
  \centering
  \subfigure[Input Image]{
    \includegraphics[width=0.2\textwidth]{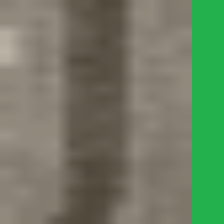}
  }\quad\quad\quad\quad
  \subfigure[Reconstructed Image]{
    \includegraphics[width=0.2\textwidth]{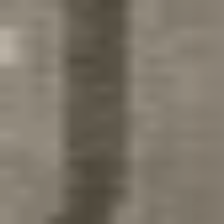}
  }
  \subfigure[\ours (MAE = 0.157)]{
    \includegraphics[width=0.8\textwidth]{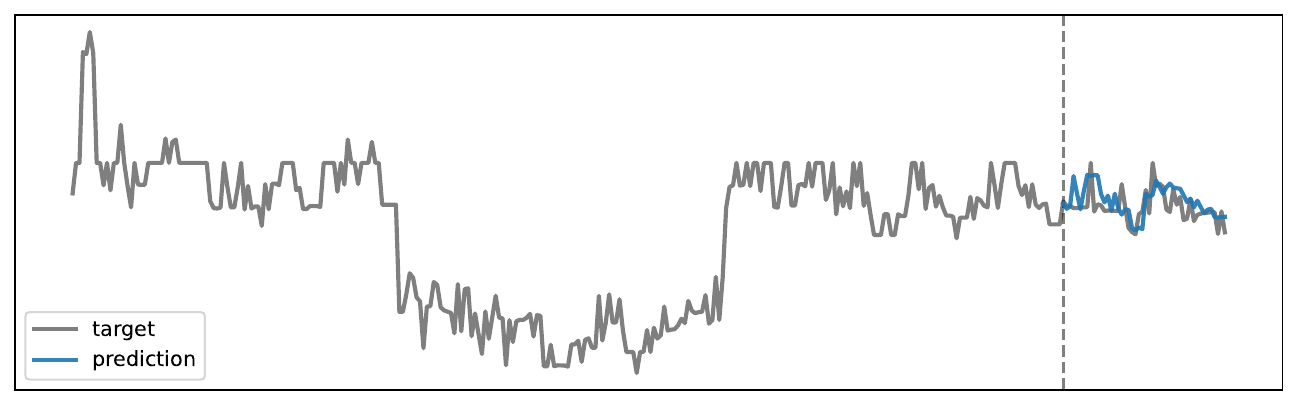}
  }
  \subfigure[\method{Moirai\textsubscript{Large}}~(MAE = 0.251)]{
    \includegraphics[width=0.8\textwidth]{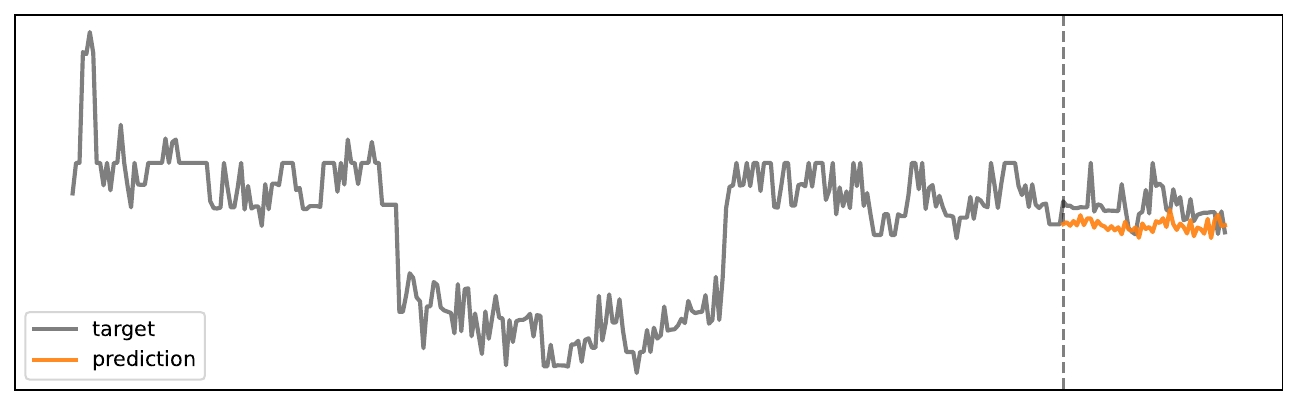}
  }
  \subfigure[Seasonal Naïve (MAE = 0.235)]{
    \includegraphics[width=0.8\textwidth]{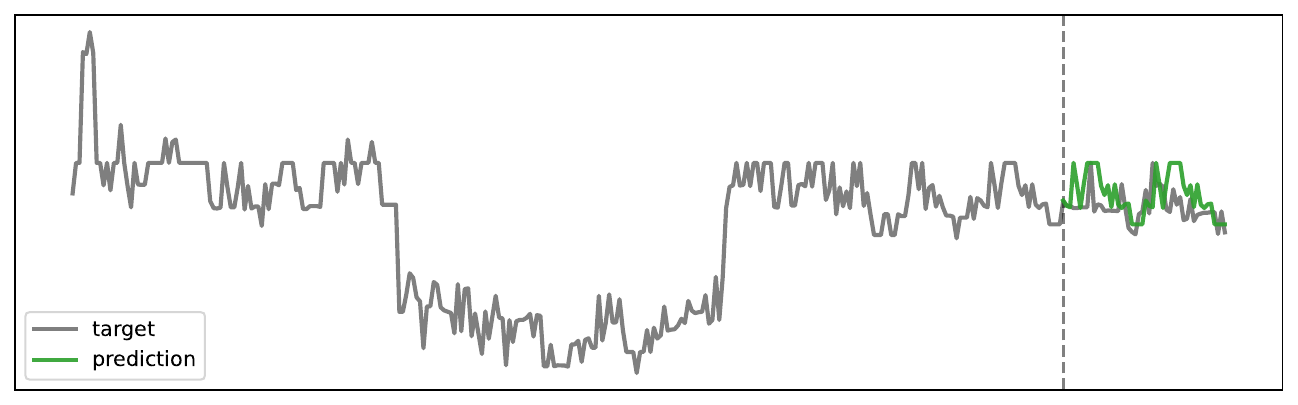}
  }
  \caption{Forecasting visualization on a sample from ETTh2. (a-b) Input/output images of \ours. (c-e) Forecasting visualization.}
  \label{fig:etth2_case1}
\end{figure*}

%% file: figure/etth2_case2/etth2_case2.tex
\begin{figure*}[t]
  \centering
  \subfigure[Input Image]{
    \includegraphics[width=0.2\textwidth]{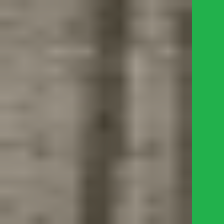}
  }\quad\quad\quad\quad
  \subfigure[Reconstructed Image]{
    \includegraphics[width=0.2\textwidth]{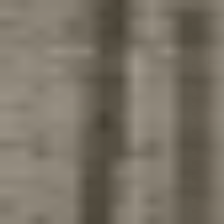}
  }
  \subfigure[\ours (MAE = 0.821)]{
    \includegraphics[width=0.8\textwidth]{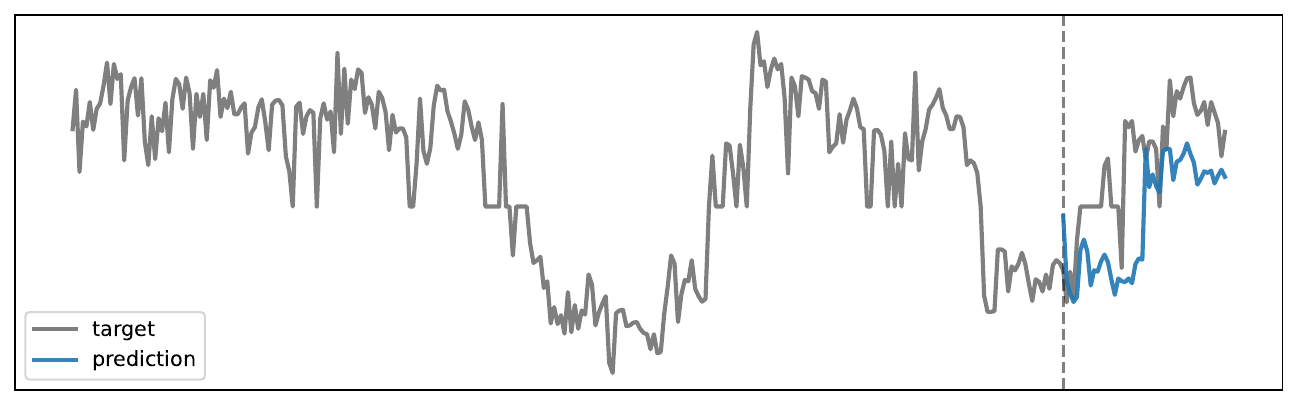}
  }
  \subfigure[\method{Moirai\textsubscript{Large}}~(MAE = 1.285)]{
    \includegraphics[width=0.8\textwidth]{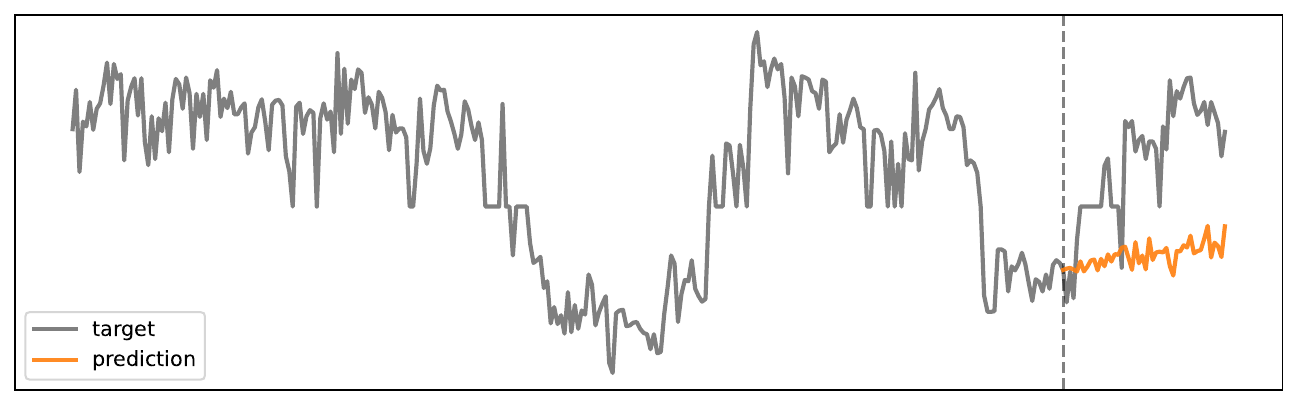}
  }
  \subfigure[Seasonal Naïve (MAE = 1.523)]{
    \includegraphics[width=0.8\textwidth]{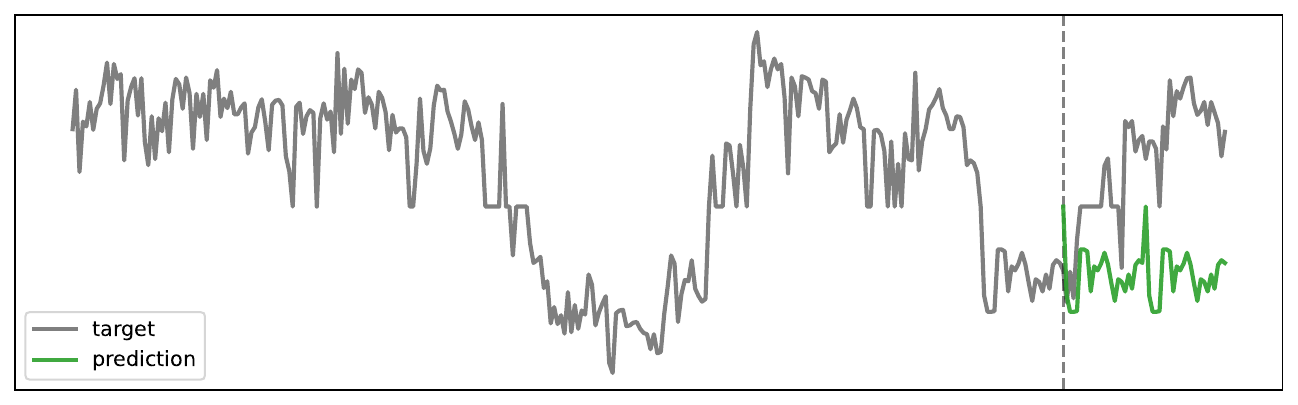}
  }
  \caption{Forecasting visualization on a sample from ETTh2. (a-b) Input/output images of \ours. (c-e) Forecasting visualization.}
  \label{fig:etth2_case2}
\end{figure*}

%% file: figure/etth1_case_bad/etth1_case_bad.tex
\begin{figure*}[t]
  \centering
  \subfigure[Input Image]{
    \includegraphics[width=0.2\textwidth]{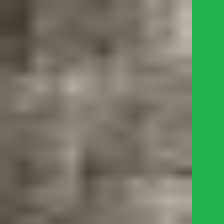}
  }\quad\quad\quad\quad
  \subfigure[Reconstructed Image]{
    \includegraphics[width=0.2\textwidth]{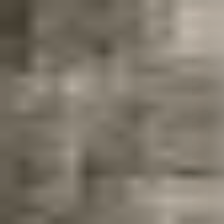}
  }
  \subfigure[\ours (MAE = 0.327)]{
    \includegraphics[width=0.8\textwidth]{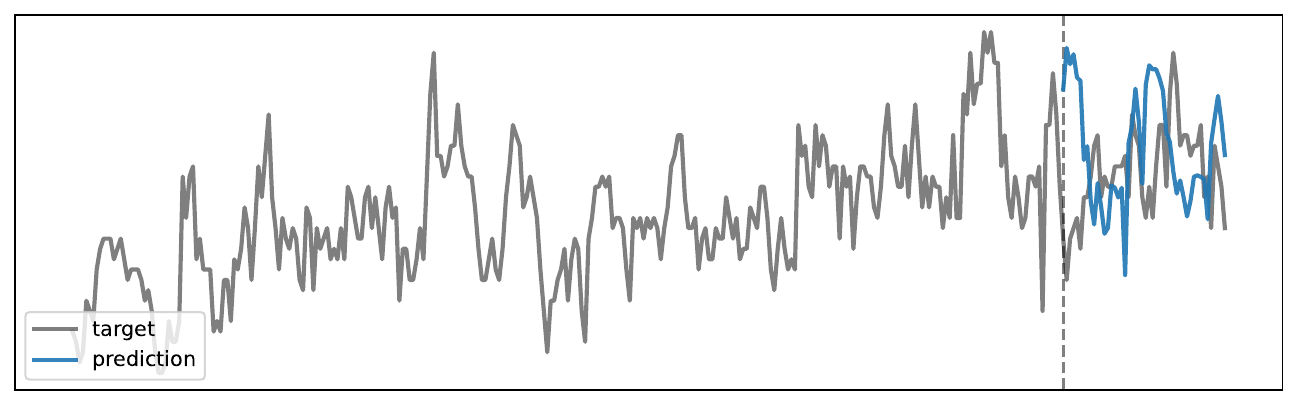}
  }
  \subfigure[\method{Moirai\textsubscript{Large}}~(MAE = 0.172)]{
    \includegraphics[width=0.8\textwidth]{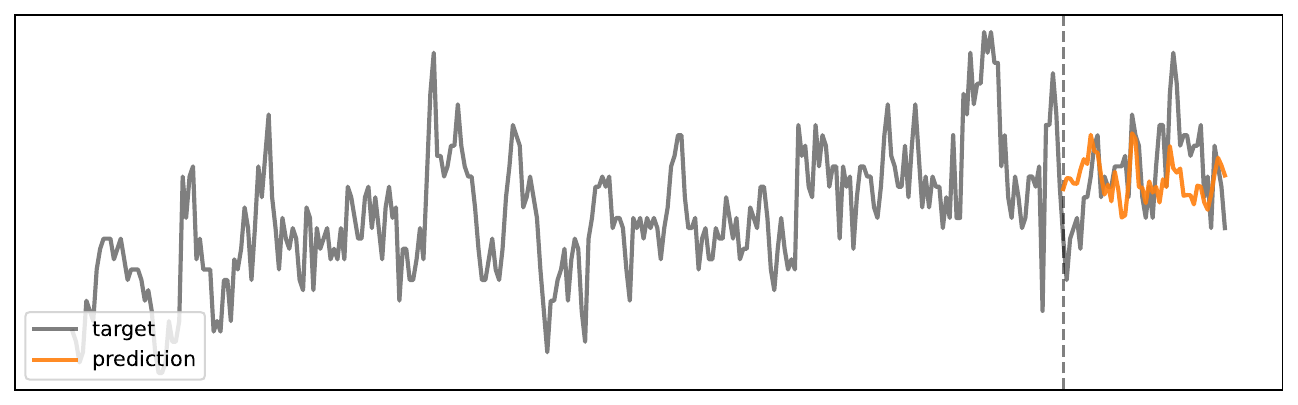
    }
  }
  \subfigure[Seasonal Naïve (MAE = 0.364)]{
    \includegraphics[width=0.8\textwidth]{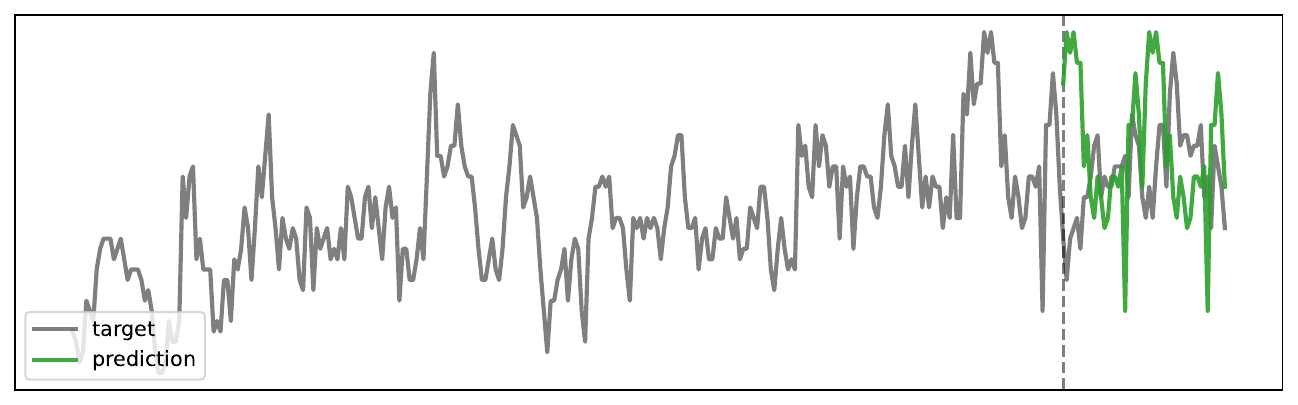}
  }
  \caption{Forecasting visualization on a sample from ETTh1, where \moirai{} outperforms \ours in terms of MAE. (a-b) Input/output images of \ours. (c-e) Forecasting visualization.}
  \label{fig:etth1_case_bad}
\end{figure*}

%% file: main.bbl
\begin{thebibliography}{65}
\providecommand{\natexlab}[1]{#1}
\providecommand{\url}[1]{\texttt{#1}}
\expandafter\ifx\csname urlstyle\endcsname\relax
  \providecommand{\doi}[1]{doi: #1}\else
  \providecommand{\doi}{doi: \begingroup \urlstyle{rm}\Url}\fi

\bibitem[Aksu et~al.(2024)Aksu, Woo, Liu, Liu, Liu, Savarese, Xiong, and Sahoo]{GIFT-Eval}
Aksu, T., Woo, G., Liu, J., Liu, X., Liu, C., Savarese, S., Xiong, C., and Sahoo, D.
\newblock Gift-eval: A benchmark for general time series forecasting model evaluation, 2024.
\newblock URL \url{https://arxiv.org/abs/2410.10393}.

\bibitem[Alexandrov et~al.(2020)Alexandrov, Benidis, Bohlke-Schneider, Flunkert, Gasthaus, Januschowski, Maddix, Rangapuram, Salinas, Schulz, Stella, Türkmen, and Wang]{GluonTS}
Alexandrov, A., Benidis, K., Bohlke-Schneider, M., Flunkert, V., Gasthaus, J., Januschowski, T., Maddix, D.~C., Rangapuram, S., Salinas, D., Schulz, J., Stella, L., Türkmen, A.~C., and Wang, Y.
\newblock {GluonTS: Probabilistic and Neural Time Series Modeling in Python}.
\newblock \emph{Journal of Machine Learning Research}, 21\penalty0 (116):\penalty0 1--6, 2020.
\newblock URL \url{http://jmlr.org/papers/v21/19-820.html}.

\bibitem[Ansari et~al.(2024)Ansari, Stella, Turkmen, Zhang, Mercado, Shen, Shchur, Rangapuram, Arango, Kapoor, et~al.]{Chronos}
Ansari, A.~F., Stella, L., Turkmen, C., Zhang, X., Mercado, P., Shen, H., Shchur, O., Rangapuram, S.~S., Arango, S.~P., Kapoor, S., et~al.
\newblock Chronos: Learning the language of time series.
\newblock \emph{arXiv preprint arXiv:2403.07815}, 2024.

\bibitem[Bao et~al.(2022)Bao, Dong, Piao, and Wei]{BEiT}
Bao, H., Dong, L., Piao, S., and Wei, F.
\newblock {BE}it: {BERT} pre-training of image transformers.
\newblock In \emph{International Conference on Learning Representations}, 2022.
\newblock URL \url{https://openreview.net/forum?id=p-BhZSz59o4}.

\bibitem[Bian et~al.(2024)Bian, Ju, Li, Xu, Cheng, and Xu]{aLLM4TS}
Bian, Y., Ju, X., Li, J., Xu, Z., Cheng, D., and Xu, Q.
\newblock Multi-patch prediction: Adapting llms for time series representation learning.
\newblock \emph{arXiv preprint arXiv:2402.04852}, 2024.

\bibitem[Bommasani et~al.(2021)Bommasani, Hudson, Adeli, Altman, Arora, von Arx, Bernstein, Bohg, Bosselut, Brunskill, et~al.]{FoundationModel}
Bommasani, R., Hudson, D.~A., Adeli, E., Altman, R., Arora, S., von Arx, S., Bernstein, M.~S., Bohg, J., Bosselut, A., Brunskill, E., et~al.
\newblock On the opportunities and risks of foundation models.
\newblock \emph{arXiv preprint arXiv:2108.07258}, 2021.

\bibitem[Brown et~al.(2020)Brown, Mann, Ryder, Subbiah, Kaplan, Dhariwal, Neelakantan, Shyam, Sastry, Askell, Agarwal, Herbert-Voss, Krueger, Henighan, Child, Ramesh, Ziegler, Wu, Winter, Hesse, Chen, Sigler, Litwin, Gray, Chess, Clark, Berner, McCandlish, Radford, Sutskever, and Amodei]{GPT}
Brown, T.~B., Mann, B., Ryder, N., Subbiah, M., Kaplan, J., Dhariwal, P., Neelakantan, A., Shyam, P., Sastry, G., Askell, A., Agarwal, S., Herbert-Voss, A., Krueger, G., Henighan, T., Child, R., Ramesh, A., Ziegler, D.~M., Wu, J., Winter, C., Hesse, C., Chen, M., Sigler, E., Litwin, M., Gray, S., Chess, B., Clark, J., Berner, C., McCandlish, S., Radford, A., Sutskever, I., and Amodei, D.
\newblock Language models are few-shot learners, 2020.
\newblock URL \url{https://arxiv.org/abs/2005.14165}.

\bibitem[Chen et~al.(2024)Chen, Shen, Fu, Li, Sun, and Liu]{CalibrationCDS}
Chen, M., Shen, L., Fu, H., Li, Z., Sun, J., and Liu, C.
\newblock Calibration of time-series forecasting: Detecting and adapting context-driven distribution shift.
\newblock In \emph{Proceedings of the 30th ACM SIGKDD Conference on Knowledge Discovery and Data Mining}, KDD '24, pp.\  341–352, New York, NY, USA, 2024. Association for Computing Machinery.
\newblock ISBN 9798400704901.
\newblock \doi{10.1145/3637528.3671926}.
\newblock URL \url{https://doi.org/10.1145/3637528.3671926}.

\bibitem[Das et~al.(2024)Das, Kong, Sen, and Zhou]{TimesFM}
Das, A., Kong, W., Sen, R., and Zhou, Y.
\newblock A decoder-only foundation model for time-series forecasting.
\newblock In \emph{Forty-first International Conference on Machine Learning}, 2024.

\bibitem[Deng et~al.(2009)Deng, Dong, Socher, Li, Li, and Fei-Fei]{ImageNet}
Deng, J., Dong, W., Socher, R., Li, L.-J., Li, K., and Fei-Fei, L.
\newblock Imagenet: A large-scale hierarchical image database.
\newblock In \emph{2009 IEEE Conference on Computer Vision and Pattern Recognition}, pp.\  248--255, 2009.
\newblock \doi{10.1109/CVPR.2009.5206848}.

\bibitem[Devlin et~al.(2019)Devlin, Chang, Lee, and Toutanova]{BERT}
Devlin, J., Chang, M.-W., Lee, K., and Toutanova, K.
\newblock {BERT}: Pre-training of deep bidirectional transformers for language understanding.
\newblock In \emph{Proceedings of the 2019 Conference of the North {A}merican Chapter of the Association for Computational Linguistics: Human Language Technologies, Volume 1 (Long and Short Papers)}, pp.\  4171--4186, Minneapolis, Minnesota, June 2019. Association for Computational Linguistics.
\newblock \doi{10.18653/v1/N19-1423}.
\newblock URL \url{https://aclanthology.org/N19-1423}.

\bibitem[Dong et~al.(2024)Dong, Wu, Wang, Qiu, Zhang, Wang, and Long]{TimeSiam}
Dong, J., Wu, H., Wang, Y., Qiu, Y.-Z., Zhang, L., Wang, J., and Long, M.
\newblock Timesiam: A pre-training framework for siamese time-series modeling.
\newblock In \emph{Forty-first International Conference on Machine Learning}, 2024.

\bibitem[Dosovitskiy et~al.(2021)Dosovitskiy, Beyer, Kolesnikov, Weissenborn, Zhai, Unterthiner, Dehghani, Minderer, Heigold, Gelly, Uszkoreit, and Houlsby]{ViT}
Dosovitskiy, A., Beyer, L., Kolesnikov, A., Weissenborn, D., Zhai, X., Unterthiner, T., Dehghani, M., Minderer, M., Heigold, G., Gelly, S., Uszkoreit, J., and Houlsby, N.
\newblock An image is worth 16x16 words: Transformers for image recognition at scale.
\newblock In \emph{International Conference on Learning Representations}, 2021.
\newblock URL \url{https://openreview.net/forum?id=YicbFdNTTy}.

\bibitem[Ekambaram et~al.(2024)Ekambaram, Jati, Dayama, Mukherjee, Nguyen, Gifford, Reddy, and Kalagnanam]{TTM}
Ekambaram, V., Jati, A., Dayama, P., Mukherjee, S., Nguyen, N., Gifford, W.~M., Reddy, C., and Kalagnanam, J.
\newblock Tiny time mixers (ttms): Fast pre-trained models for enhanced zero/few-shot forecasting of multivariate time series.
\newblock \emph{Advances in Neural Information Processing Systems}, 37:\penalty0 74147--74181, 2024.

\bibitem[Feng et~al.(2024)Feng, Huang, and Krompass]{GTT}
Feng, C., Huang, L., and Krompass, D.
\newblock Only the curve shape matters: Training foundation models for zero-shot multivariate time series forecasting through next curve shape prediction.
\newblock \emph{arXiv preprint arXiv:2402.07570}, 2024.

\bibitem[Fu et~al.(2024)Fu, Chen, Zhang, Yang, Ma, and Yang]{SyntheticTS}
Fu, F., Chen, J., Zhang, J., Yang, C., Ma, L., and Yang, Y.
\newblock Are synthetic time-series data really not as good as real data?, 2024.
\newblock URL \url{https://arxiv.org/abs/2402.00607}.

\bibitem[Godahewa et~al.(2021)Godahewa, Bergmeir, Webb, Hyndman, and Montero-Manso]{Monash}
Godahewa, R.~W., Bergmeir, C., Webb, G.~I., Hyndman, R., and Montero-Manso, P.
\newblock Monash time series forecasting archive.
\newblock In \emph{Thirty-fifth Conference on Neural Information Processing Systems Datasets and Benchmarks Track (Round 2)}, 2021.
\newblock URL \url{https://openreview.net/forum?id=wEc1mgAjU-}.

\bibitem[Goswami et~al.(2024)Goswami, Szafer, Choudhry, Cai, Li, and Dubrawski]{Moment}
Goswami, M., Szafer, K., Choudhry, A., Cai, Y., Li, S., and Dubrawski, A.
\newblock Moment: A family of open time-series foundation models.
\newblock In \emph{Forty-first International Conference on Machine Learning}, 2024.

\bibitem[Gruver et~al.(2023)Gruver, Finzi, Qiu, and Wilson]{LLMTime}
Gruver, N., Finzi, M., Qiu, S., and Wilson, A.~G.
\newblock Large language models are zero-shot time series forecasters.
\newblock \emph{Advances in Neural Information Processing Systems}, 36, 2023.

\bibitem[Han et~al.(2024)Han, Ye, and Zhan]{ChannelIndependence}
Han, L., Ye, H.-J., and Zhan, D.-C.
\newblock The capacity and robustness trade-off: Revisiting the channel independent strategy for multivariate time series forecasting.
\newblock \emph{IEEE Transactions on Knowledge \& Data Engineering}, \penalty0 (01):\penalty0 1--14, 2024.

\bibitem[Hatami et~al.(2018)Hatami, Gavet, and Debayle]{hatami2018classification}
Hatami, N., Gavet, Y., and Debayle, J.
\newblock Classification of time-series images using deep convolutional neural networks.
\newblock In \emph{Tenth international conference on machine vision (ICMV 2017)}, volume 10696, pp.\  242--249. SPIE, 2018.

\bibitem[He et~al.(2022)He, Chen, Xie, Li, Doll{\'a}r, and Girshick]{mae}
He, K., Chen, X., Xie, S., Li, Y., Doll{\'a}r, P., and Girshick, R.
\newblock Masked autoencoders are scalable vision learners.
\newblock In \emph{Proceedings of the IEEE/CVF conference on computer vision and pattern recognition}, pp.\  16000--16009, 2022.

\bibitem[Hsu et~al.(2021)Hsu, Bolte, Tsai, Lakhotia, Salakhutdinov, and Mohamed]{HuBert}
Hsu, W.-N., Bolte, B., Tsai, Y.-H.~H., Lakhotia, K., Salakhutdinov, R., and Mohamed, A.
\newblock {HuBERT}: Self-supervised speech representation learning by masked prediction of hidden units.
\newblock \emph{IEEE/ACM Transactions on Audio, Speech, and Language Processing}, 29:\penalty0 3451--3460, 2021.

\bibitem[Jin et~al.(2024)Jin, Wang, Ma, Chu, Zhang, Shi, Chen, Liang, Li, Pan, et~al.]{timellm}
Jin, M., Wang, S., Ma, L., Chu, Z., Zhang, J.~Y., Shi, X., Chen, P.-Y., Liang, Y., Li, Y.-F., Pan, S., et~al.
\newblock Time-llm: Time series forecasting by reprogramming large language models.
\newblock In \emph{The Twelfth International Conference on Learning Representations}, 2024.

\bibitem[Kim et~al.(2022)Kim, Kim, Tae, Park, Choi, and Choo]{RevIn}
Kim, T., Kim, J., Tae, Y., Park, C., Choi, J.-H., and Choo, J.
\newblock Reversible instance normalization for accurate time-series forecasting against distribution shift.
\newblock In \emph{International Conference on Learning Representations}, 2022.
\newblock URL \url{https://openreview.net/forum?id=cGDAkQo1C0p}.

\bibitem[Krizhevsky et~al.(2012)Krizhevsky, Sutskever, and Hinton]{CNN}
Krizhevsky, A., Sutskever, I., and Hinton, G.~E.
\newblock Imagenet classification with deep convolutional neural networks.
\newblock \emph{Advances in neural information processing systems}, 25, 2012.

\bibitem[Li et~al.(2020)Li, Kang, and Li]{li2020forecasting}
Li, X., Kang, Y., and Li, F.
\newblock Forecasting with time series imaging.
\newblock \emph{Expert Systems with Applications}, 160:\penalty0 113680, 2020.

\bibitem[Li et~al.(2024)Li, Li, and Yan]{li2024time}
Li, Z., Li, S., and Yan, X.
\newblock Time series as images: Vision transformer for irregularly sampled time series.
\newblock \emph{Advances in Neural Information Processing Systems}, 36, 2024.

\bibitem[Lin et~al.(2024)Lin, Lin, Wu, Chen, and Yang]{SparseTSF}
Lin, S., Lin, W., Wu, W., Chen, H., and Yang, J.
\newblock Sparsetsf: Modeling long-term time series forecasting with 1k parameters.
\newblock In \emph{Forty-first International Conference on Machine Learning}, 2024.

\bibitem[Liu et~al.(2025)Liu, Guo, Dai, Li, Bao, Ren, Jiang, and Xia]{CALF}
Liu, P., Guo, H., Dai, T., Li, N., Bao, J., Ren, X., Jiang, Y., and Xia, S.-T.
\newblock Calf: Aligning llms for time series forecasting via cross-modal fine-tuning.
\newblock In \emph{Proceedings of the AAAI Conference on Artificial Intelligence}, volume~39, pp.\  18915--18923, 2025.

\bibitem[Liu et~al.(2022)Liu, Wu, Wang, and Long]{Non_stationary_Transformer}
Liu, Y., Wu, H., Wang, J., and Long, M.
\newblock Non-stationary transformers: Exploring the stationarity in time series forecasting, 2022.

\bibitem[Liu et~al.(2024)Liu, Zhang, Li, Huang, Wang, and Long]{Timer}
Liu, Y., Zhang, H., Li, C., Huang, X., Wang, J., and Long, M.
\newblock Timer: Generative pre-trained transformers are large time series models.
\newblock In \emph{Forty-first International Conference on Machine Learning}, 2024.

\bibitem[Ma et~al.(2023)Ma, Liu, Zheng, Huang, Zhu, Yu, and Kwok]{Pretraining-Survey}
Ma, Q., Liu, Z., Zheng, Z., Huang, Z., Zhu, S., Yu, Z., and Kwok, J.~T.
\newblock A survey on time-series pre-trained models.
\newblock \emph{arXiv preprint arXiv:2305.10716}, 2023.

\bibitem[Nie et~al.(2022)Nie, Nguyen, Sinthong, and Kalagnanam]{PatchTST}
Nie, Y., Nguyen, N.~H., Sinthong, P., and Kalagnanam, J.
\newblock A time series is worth 64 words: Long-term forecasting with transformers.
\newblock In \emph{The Eleventh International Conference on Learning Representations}, 2022.

\bibitem[Peebles \& Xie(2023)Peebles and Xie]{DiT}
Peebles, W. and Xie, S.
\newblock Scalable diffusion models with transformers.
\newblock In \emph{Proceedings of the IEEE/CVF International Conference on Computer Vision}, pp.\  4195--4205, 2023.

\bibitem[Qiu et~al.(2024)Qiu, Hu, Zhou, Wu, Du, Zhang, Guo, Zhou, Jensen, Sheng, and Yang]{qiu2024tfb}
Qiu, X., Hu, J., Zhou, L., Wu, X., Du, J., Zhang, B., Guo, C., Zhou, A., Jensen, C.~S., Sheng, Z., and Yang, B.
\newblock {TFB:} towards comprehensive and fair benchmarking of time series forecasting methods.
\newblock \emph{Proc. {VLDB} Endow.}, 17\penalty0 (9):\penalty0 2363--2377, 2024.

\bibitem[Radford et~al.(2019)Radford, Wu, Child, Luan, Amodei, Sutskever, et~al.]{GPT-2}
Radford, A., Wu, J., Child, R., Luan, D., Amodei, D., Sutskever, I., et~al.
\newblock Language models are unsupervised multitask learners.
\newblock \emph{OpenAI blog}, 1\penalty0 (8):\penalty0 9, 2019.

\bibitem[Rombach et~al.(2022)Rombach, Blattmann, Lorenz, Esser, and Ommer]{Latent-Diffusion}
Rombach, R., Blattmann, A., Lorenz, D., Esser, P., and Ommer, B.
\newblock High-resolution image synthesis with latent diffusion models.
\newblock In \emph{Proceedings of the IEEE/CVF conference on computer vision and pattern recognition}, pp.\  10684--10695, 2022.

\bibitem[Schick \& Sch{\"u}tze(2021)Schick and Sch{\"u}tze]{PromptTuning}
Schick, T. and Sch{\"u}tze, H.
\newblock Exploiting cloze-questions for few-shot text classification and natural language inference.
\newblock In \emph{Proceedings of the 16th Conference of the European Chapter of the Association for Computational Linguistics: Main Volume}, pp.\  255--269, 2021.

\bibitem[Semenoglou et~al.(2023)Semenoglou, Spiliotis, and Assimakopoulos]{semenoglou2023image}
Semenoglou, A.-A., Spiliotis, E., and Assimakopoulos, V.
\newblock Image-based time series forecasting: A deep convolutional neural network approach.
\newblock \emph{Neural Networks}, 157:\penalty0 39--53, 2023.

\bibitem[Shi et~al.(2024)Shi, Wang, Nie, Li, Ye, Wen, and Jin]{Time-MoE}
Shi, X., Wang, S., Nie, Y., Li, D., Ye, Z., Wen, Q., and Jin, M.
\newblock Time-moe: Billion-scale time series foundation models with mixture of experts.
\newblock \emph{arXiv preprint arXiv:2409.16040}, 2024.

\bibitem[Sood et~al.(2021)Sood, Zeng, Cohen, Balch, and Veloso]{sood2021visual}
Sood, S., Zeng, Z., Cohen, N., Balch, T., and Veloso, M.
\newblock Visual time series forecasting: an image-driven approach.
\newblock In \emph{Proceedings of the Second ACM International Conference on AI in Finance}, pp.\  1--9, 2021.

\bibitem[Suvorov et~al.(2022)Suvorov, Logacheva, Mashikhin, Remizova, Ashukha, Silvestrov, Kong, Goka, Park, and Lempitsky]{LaMa}
Suvorov, R., Logacheva, E., Mashikhin, A., Remizova, A., Ashukha, A., Silvestrov, A., Kong, N., Goka, H., Park, K., and Lempitsky, V.
\newblock Resolution-robust large mask inpainting with fourier convolutions.
\newblock In \emph{Proceedings of the IEEE/CVF winter conference on applications of computer vision}, pp.\  2149--2159, 2022.

\bibitem[Tan et~al.(2024)Tan, Merrill, Gupta, Althoff, and Hartvigsen]{AreLLMUseful}
Tan, M., Merrill, M.~A., Gupta, V., Althoff, T., and Hartvigsen, T.
\newblock Are language models actually useful for time series forecasting?
\newblock \emph{arXiv preprint arXiv:2406.16964}, 2024.

\bibitem[Touvron et~al.(2023)Touvron, Lavril, Izacard, Martinet, Lachaux, Lacroix, Rozi{\`e}re, Goyal, Hambro, Azhar, et~al.]{llama}
Touvron, H., Lavril, T., Izacard, G., Martinet, X., Lachaux, M.-A., Lacroix, T., Rozi{\`e}re, B., Goyal, N., Hambro, E., Azhar, F., et~al.
\newblock Llama: Open and efficient foundation language models.
\newblock \emph{arXiv preprint arXiv:2302.13971}, 2023.

\bibitem[Wang et~al.(2025)Wang, Zhao, Luo, Zhou, Jiang, Li, Li, and Pan]{CBraMod}
Wang, J., Zhao, S., Luo, Z., Zhou, Y., Jiang, H., Li, S., Li, T., and Pan, G.
\newblock {CB}ramod: A criss-cross brain foundation model for {EEG} decoding.
\newblock In \emph{The Thirteenth International Conference on Learning Representations}, 2025.
\newblock URL \url{https://openreview.net/forum?id=NPNUHgHF2w}.

\bibitem[Wang \& Oates(2015{\natexlab{a}})Wang and Oates]{wang2015imaging}
Wang, Z. and Oates, T.
\newblock Imaging time-series to improve classification and imputation.
\newblock In \emph{Proceedings of the 24th International Conference on Artificial Intelligence}, pp.\  3939--3945, 2015{\natexlab{a}}.

\bibitem[Wang \& Oates(2015{\natexlab{b}})Wang and Oates]{wang2015spatially}
Wang, Z. and Oates, T.
\newblock Spatially encoding temporal correlations to classify temporal data using convolutional neural networks.
\newblock \emph{arXiv preprint arXiv:1509.07481}, 2015{\natexlab{b}}.

\bibitem[Wimmer \& Rekabsaz(2023)Wimmer and Rekabsaz]{wimmer2023leveraging}
Wimmer, C. and Rekabsaz, N.
\newblock Leveraging vision-language models for granular market change prediction.
\newblock \emph{arXiv preprint arXiv:2301.10166}, 2023.

\bibitem[Woo et~al.(2022{\natexlab{a}})Woo, Liu, Sahoo, Kumar, and Hoi]{CoST}
Woo, G., Liu, C., Sahoo, D., Kumar, A., and Hoi, S.
\newblock Co{ST}: Contrastive learning of disentangled seasonal-trend representations for time series forecasting.
\newblock In \emph{International Conference on Learning Representations}, 2022{\natexlab{a}}.
\newblock URL \url{https://openreview.net/forum?id=PilZY3omXV2}.

\bibitem[Woo et~al.(2022{\natexlab{b}})Woo, Liu, Sahoo, Kumar, and Hoi]{ETSformer}
Woo, G., Liu, C., Sahoo, D., Kumar, A., and Hoi, S. C.~H.
\newblock Etsformer: Exponential smoothing transformers for time-series forecasting.
\newblock \emph{CoRR}, abs/2202.01381, 2022{\natexlab{b}}.
\newblock URL \url{https://arxiv.org/abs/2202.01381}.

\bibitem[Woo et~al.(2024)Woo, Liu, Kumar, Xiong, Savarese, and Sahoo]{Moirai}
Woo, G., Liu, C., Kumar, A., Xiong, C., Savarese, S., and Sahoo, D.
\newblock Unified training of universal time series forecasting transformers.
\newblock In \emph{Forty-first International Conference on Machine Learning}, 2024.

\bibitem[Wu et~al.(2021)Wu, Xu, Wang, and Long]{Autoformer}
Wu, H., Xu, J., Wang, J., and Long, M.
\newblock Autoformer: Decomposition transformers with auto-correlation for long-term series forecasting.
\newblock \emph{Advances in Neural Information Processing Systems}, 34:\penalty0 22419--22430, 2021.

\bibitem[Wu et~al.(2023)Wu, Hu, Liu, Zhou, Wang, and Long]{TimesNet}
Wu, H., Hu, T., Liu, Y., Zhou, H., Wang, J., and Long, M.
\newblock Timesnet: Temporal 2d-variation modeling for general time series analysis.
\newblock In \emph{The Eleventh International Conference on Learning Representations}, 2023.
\newblock URL \url{https://openreview.net/forum?id=ju_Uqw384Oq}.

\bibitem[Xue \& Salim(2023)Xue and Salim]{PromptCast}
Xue, H. and Salim, F.~D.
\newblock Promptcast: A new prompt-based learning paradigm for time series forecasting.
\newblock \emph{IEEE Transactions on Knowledge and Data Engineering}, 2023.

\bibitem[Yang et~al.(2024)Yang, Wang, Fan, Cohen, Zhao, and Zhang]{ViTime}
Yang, L., Wang, Y., Fan, X., Cohen, I., Zhao, Y., and Zhang, Z.
\newblock Vitime: A visual intelligence-based foundation model for time series forecasting.
\newblock \emph{arXiv preprint arXiv:2407.07311}, 2024.

\bibitem[Yue et~al.(2022)Yue, Wang, Duan, Yang, Huang, Tong, and Xu]{TS2Vec}
Yue, Z., Wang, Y., Duan, J., Yang, T., Huang, C., Tong, Y., and Xu, B.
\newblock Ts2vec: Towards universal representation of time series.
\newblock In \emph{Proceedings of the AAAI Conference on Artificial Intelligence}, volume~36, pp.\  8980--8987, 2022.

\bibitem[Zaken et~al.(2022)Zaken, Goldberg, and Ravfogel]{BitFit}
Zaken, E.~B., Goldberg, Y., and Ravfogel, S.
\newblock Bitfit: Simple parameter-efficient fine-tuning for transformer-based masked language-models.
\newblock In \emph{Proceedings of the 60th Annual Meeting of the Association for Computational Linguistics (Volume 2: Short Papers)}, pp.\  1--9, 2022.

\bibitem[Zeng et~al.(2023)Zeng, Chen, Zhang, and Xu]{LTSF-Linear}
Zeng, A., Chen, M., Zhang, L., and Xu, Q.
\newblock Are transformers effective for time series forecasting?
\newblock In \emph{Proceedings of the AAAI conference on artificial intelligence}, volume~37, pp.\  11121--11128, 2023.

\bibitem[Zerveas et~al.(2021)Zerveas, Jayaraman, Patel, Bhamidipaty, and Eickhoff]{Denoising-Autoencoders}
Zerveas, G., Jayaraman, S., Patel, D., Bhamidipaty, A., and Eickhoff, C.
\newblock A transformer-based framework for multivariate time series representation learning.
\newblock In \emph{Proceedings of the 27th ACM SIGKDD conference on knowledge discovery \& data mining}, pp.\  2114--2124, 2021.

\bibitem[Zhang et~al.(2024)Zhang, Wen, Zhang, Cai, Jin, Liu, Zhang, Liang, Pang, Song, et~al.]{Self-Supervised-Survey}
Zhang, K., Wen, Q., Zhang, C., Cai, R., Jin, M., Liu, Y., Zhang, J.~Y., Liang, Y., Pang, G., Song, D., et~al.
\newblock Self-supervised learning for time series analysis: Taxonomy, progress, and prospects.
\newblock \emph{IEEE Transactions on Pattern Analysis and Machine Intelligence}, 2024.

\bibitem[Zhang et~al.(2023)Zhang, Zhang, Zheng, Chen, Gao, Ge, Teng, Jelloul, Rao, Guo, et~al.]{zhang2023insight}
Zhang, Y., Zhang, Y., Zheng, M., Chen, K., Gao, C., Ge, R., Teng, S., Jelloul, A., Rao, J., Guo, X., et~al.
\newblock Insight miner: A time series analysis dataset for cross-domain alignment with natural language.
\newblock In \emph{NeurIPS 2023 AI for Science Workshop}, 2023.

\bibitem[Zhou et~al.(2021)Zhou, Zhang, Peng, Zhang, Li, Xiong, and Zhang]{Informer}
Zhou, H., Zhang, S., Peng, J., Zhang, S., Li, J., Xiong, H., and Zhang, W.
\newblock Informer: Beyond efficient transformer for long sequence time-series forecasting.
\newblock In \emph{Proceedings of the AAAI conference on artificial intelligence}, volume~35, pp.\  11106--11115, 2021.

\bibitem[Zhou et~al.(2022)Zhou, Ma, Wen, Wang, Sun, and Jin]{FEDformer}
Zhou, T., Ma, Z., Wen, Q., Wang, X., Sun, L., and Jin, R.
\newblock Fedformer: Frequency enhanced decomposed transformer for long-term series forecasting.
\newblock In \emph{International Conference on Machine Learning}, pp.\  27268--27286. PMLR, 2022.

\bibitem[Zhou et~al.(2023)Zhou, Niu, Sun, Jin, et~al.]{GPT4TS}
Zhou, T., Niu, P., Sun, L., Jin, R., et~al.
\newblock One fits all: Power general time series analysis by pretrained lm.
\newblock In \emph{Advances in neural information processing systems}, volume~36, pp.\  43322--43355, 2023.

\end{thebibliography}
